%% file: main.tex
\documentclass[examplefnt,biber]{nowfnt} 

\usepackage{graphicx}
\usepackage{subcaption}
\usepackage[utf8]{inputenc}
\usepackage{amsmath}
\usepackage{bm}
\usepackage{cancel}
\usepackage[normalem]{ulem}

\usepackage{amsfonts}
\usepackage{mathtools}

\DeclarePairedDelimiter\floor{\lfloor}{\rfloor}

\newcommand{\Tr}{\text{Tr}}
\newcommand{\MF}{\text{MF}}
\newcommand{\GP}{\text{GP}}
\newcommand{\inn}{\text{in}}
\newcommand{\GPR}{\text{GPR}}
\newcommand{\MSE}{\text{MSE}}
\newcommand{\EK}{\text{EK}}
\newcommand{\RG}{\text{RG}}
\newcommand{\eff}{\text{eff}}
\newcommand{\con}{\text{con}}
\newcommand{\Erf}{\text{Erf}}
\newcommand{\const}{\text{const}}
\newcommand{\dif}{\text{d}}
\newcommand{\data}{\text{data}}
\newcommand{\cL}{\mathcal{L}}
\newcommand{\cD}{\mathcal{D}}
\newcommand{\cN}{\mathcal{N}}
\newcommand{\R}{\mathbb{R}}

\title{Applications of Statistical Field Theory in Deep Learning}

\subtitle{A gentle introduction}

\issuesetup
{%
 copyrightowner={Z.~Ringel, I.~Seroussi, M.~Helias},
 volume        = xx,
 issue         = xx,
 pubyear       = 2025,
 isbn          = xxx-x-xxxxx-xxx-x,
 eisbn         = xxx-x-xxxxx-xxx-x,
 doi           = 10.1561/XXXXXXXXX,
 firstpage     = 1, 
 lastpage      = 18
 }

\usepackage{mwe}

\author[1]{Ringel,Zohar}
\author[1]{Rubin, Noa}
\author[1]{Mor, Edo}
\author[2,3]{Helias, Moritz}
\author[4]{Seroussi, Inbar}

\affil[1]{The Hebrew University of Jerusalem, Jerusalem, Israel.}
\affil[2]{Jülich Research Centre, Jülich, Germany.}
\affil[3]{RWTH Aachen University, Aachen, Germany.}
\affil[4]{Tel Aviv University, Tel Aviv, Israel.}

\articledatabox{\nowfntstandardcitation}

\begin{document}

\makeabstracttitle

\begin{abstract}
Deep learning algorithms have made incredible strides in the past decade, yet due to their complexity, the science of deep learning remains in its early stages. Being an experimentally driven field, it is natural to seek a theory of deep learning within the physics paradigm. As deep learning is largely about learning functions and distributions over functions, statistical field theory, a rich and versatile toolbox for tackling complex distributions over functions (fields) is an obvious choice of formalism. Research efforts carried out in the past few years have demonstrated the ability of field theory to provide useful insights on generalization, implicit bias, and feature learning effects. Here we provide a pedagogical review of this emerging line of research.    
\end{abstract}

\input{Chapters/Preliminaries}
\input{Chapters/Infinite_networks_in_standard_scaling}

\input{Chapters/Bayesian_Neural_Networks_in_the_Feature_Learning_Regime}

\input{Chapters/Field_theory_approach_to_dynamics}
\input{Chapters/Summary_and_Outlook}

\backmatter  


\end{document}

%% file: Chapters/Preliminaries.tex
\chapter{Preliminaries}
\label{c-Preliminaries} In this chapter we establish notation, manage expectations, and give a basic introduction to the analytical tools we require. The tools introduced are quite diverse, spanning math and physics topics. Rather than aiming to make the reader confident in them, the purpose here is to give the minimal user interface required for the analysis of trained neural networks carried out in later chapters. Readers looking to establish their knowledge of Replicas, Path-Integrals, and Gaussian Processes can consider the following introductory books   \cite{MezardBook,schulman1996,Rasmussen2005}.

\section{Motivation}
The goal of this review is to demonstrate how field theory and statistical physics provide a rich framework that may, in the future, accommodate a unified theory of deep learning. This then brings into mind questions concerning the need for such a theory and its general contours. 

In traditional scientific disciplines, simple universal results are celebrated. Snell's law describing light scattering, Carnot's optimal engine efficiency, and the universal scaling of magnetic clusters in Ising-like transitions lead to concrete and universal predictions. Unlike naturally occurring systems, virtual ones, such as deep neural networks, have been post-selected by software engineers to adapt to various complex learning settings. They are therefore more akin to biological systems than to physical systems, where complexity is a fact of life rather than an obfuscation of a simple underlying phenomenon.  

To further illustrate this point, let us imagine one finds something similar to Schrödinger's equation (a linear partial different equation) as a universal description of a trained transformer network. This would be very convenient to theorists, as we have various analytical and numerical tools to analyze linear differential equations. However, it would also mean that a transformer is equivalent to a linear model. However, linear models were studied extensively in machine learning and did not lead to transformer-grade performance. It is therefore difficult to believe that such a linear theory can exist. This fleshes out a tension between analytical solutions which, by virtue of being called ``solutions'', should be computationally simple or ``linear'' and a theory capable of tracking deep learning.

The above explainability paradox is not new to physicists studying complex systems and various workarounds exist based on averaging, universality, dimensional reduction, modularity, and asymptotic limits. For instance, while SAT-3 is an NP-hard problem, various computational techniques borrowed from spin-glass theory can accurately predict thresholds, in terms of the ratio of variables and clauses/constraints, at which large-scale SAT-3 instances become easy to solve. These theories escape the paradox of finding an analytical solution to an NP-hard problem by averaging over ensembles of problems instead of solving a particular SAT-3 instance. Still, they provide useful statistical rules for determining whether a typical instance is solvable. This is also an example of dimensional reduction since out of the many parameters defining a SAT-3 instance, only a single quantity, the above mentioned variable to clauses ratio, controls hardness in large-scale SAT-3 problems \citep{MezardBook}. Similarly, Random Matrix Theory \citep{Potters_Bouchaud_2020} provides accurate statistical information on the spectrum large matrices (e.g. Wigner's semi-circle law) while circumventing the $O(n^3)$ computational difficulty of diagonalizing $n$ dimensional matrices. This example also showcases universality, as various changes to the statistics of random matrix elements still lead to the same distribution law for eigenvalues. 

In building a theory of deep learning, one needs to negotiate these difficulties and tensions. One should also be wary of a common pitiful in which the analytical ``solutions" offered by theories involve equations which require expensive numerical solutions, potentially more complex than training the networks themselves. 

Optimism can however be drawn from various successful recent outcomes, some of which are described in detail in this review. In particular {\bf (i)} A linear description of highly overparametrized DNNs (the {\it GP limit}) which, despite describing a less powerful regime of deep learning, provides insights on fundamental questions about overparametrization and allows rationalizing architecture choices. {\bf (ii)} Analytically inspired recipes for {\it hyper-parameter transfer} between small and large networks, which are based on solutions of the very short time dynamics yet appear to apply throughout training. {\bf (iii)} The empirical observation of {\it scaling laws} where, based on only two constants (and faith in the scaling law), one can predict the performance of ChatGPT and other real-world networks aiding the allocation of computation resources. We shall further explain how these predictions can be derived within our field-theory framework.

\section{Notation}
The following notation will be used throughout this review. 
\begin{center}
\begin{tabular}{ |c|c||c|c| } 
 \hline
 Width & $N$ & Number of channels & $C$ \\ 
 Input dimension & $d$ & Training set size & $P$ \\ 
 Data indices & $\mu,\nu,\chi,\xi$ & Neuron indices & $i,j$ \\ 
 Feature indices & $k,q$ & Replica indices & $a,b$ \\ 
 Learnable parameters & $\theta$ & Readout layer weights & $a_i$ \\ 
 Hidden layer weights & $w^{(l)}_{ij}$ & Learning rate & $\eta$ \\ 
 Ridge parameter & $\kappa^2$ & Weight variances & $\sigma_a^2,\sigma_{w^{(l)}}^2$ \\
 Training points & $x_{\mu}$ & Test point & $x_*$ or $x_0$  \\
 NNGP Kernel & $K_{\mu \nu}$ & NTK Kernel & $\Theta_{\mu \nu}$ \\Regression Target & $y(x)$,$y_{\mu}$ & Data measure & $\dif\mu_x = \dif x \; p(x)$ \\
 Gaussian Dis. in $x$ & ${\cal N}(\mu,\Sigma;x)$ & Pre-activations & $h_{i,\mu}$, \; $h_i(x)$ \\
 Network outputs & $f_{\mu}$,$f(x)$ & Determinant & \text{det}(..) \\
 Activation function & $\sigma(...)$ & $\kappa_{\mu_1...\mu_n}$ & $n$'th cumulant \\
 \hline
\end{tabular}
\end{center}

\section{Gaussian Integrals}
Multivariate Gaussian integrals are central to field theory. Here we recall the definition of those and several useful identities, in particular Wick's theorem and square completion. 

We denote a multivariate Gaussian distribution of $w \in R^d$ with mean $\mu$ and covariance matrix $\Sigma$, by $\mathcal{N}[\mu,\Sigma;w]$ where 
\begin{align}
\mathcal{N}[\mu,\Sigma;w] &\equiv \frac{1}{\sqrt{(2\pi)^d \det(\Sigma)}} e^{-\frac{1}{2} [w-\mu]^T \Sigma^{-1} [w-\mu]}
\end{align}

{\bf Marginalization.} Given a Gaussian $\mathcal{N}[\mu,\Sigma;(w_1..w_d)]$, the marginalzied probability distribution for $w'=(w_1..w_{d'<d})$ is easily written in terms of the $d'\times d'$ matrix $\Sigma'_{ij}\equiv\Sigma_{ij}$ ($i,j \in [1..d']$) 
\begin{align}
\int dw_{d'+1}..dw_d \mathcal{N}[\mu,\Sigma;w] &= \mathcal{N}[\mu',\Sigma';w']
\end{align}
where $\mu'=(\mu_1...\mu_{d'})$. 

{\bf Expectation values.} Any finite moment of a Gaussian random variable can be calculated using Wick's-Isserlis' theorem. Denoting the moment by $w_{i_1}...w_{i_n}$, we consider all partitions of the indices $i_1...i_n$ as follows: We first split the indices into one set $A$, then the remaining indices into all possible pairs. Each of these possible joint splits is denoted by $A,P$. The n-th moment can be expressed as the sum of all such splits, 
\begin{align}
\label{Eq:WickWithAverage}
\int \prod_{j=1}^n dw_{i_j} \mathcal{N}[\mu,\Sigma;w] w_{i_1}...w_{i_n} &= \sum_{A,P \in Splits} [\prod_{a \in A} \mu_a][\prod_{p \in P} \Sigma_{p_1 p_2}] 
\end{align}
where $p$ is the set of pairs in $P$. For instance, say we have the moment $w_1 w_2 w_3 w_4$ we find 
\begin{align}
&\int \prod_{j=1}^4dw_j \mathcal{N}[\mu,\Sigma;w] w_{1}w_{2}w_{3}w_4 = \mu_1 \mu_2 \mu_3 \mu_4 \\ \nonumber 
&+ \mu_1 \mu_2 \Sigma_{34} + \mu_1 \mu_3 \Sigma_{24} + \mu_1 \mu_4 \Sigma_{23} + \mu_2 \mu_3 \Sigma_{14} + \mu_2 \mu_4 \Sigma_{13} + \mu_3 \mu_4 \Sigma_{12} \\ \nonumber
&+ \Sigma_{12}\Sigma_{34}+\Sigma_{13}\Sigma_{24}+\Sigma_{14}\Sigma_{23}
\end{align}
The common use case for this formula is when $\mu=0$, where it amounts to splitting the indices into all possible pairs. 

{\bf Square completion.} We often encounter integrals of the following form, say when using Fourier transforms (moment generating function)
\begin{align}
\int dw \mathcal{N}[0,\Sigma;w] e^{-k^T w} &= \int dw \frac{1}{\sqrt{(2\pi)^d ||\Sigma||}} e^{-\frac{1}{2} w^T \Sigma^{-1} w-k^T w} = e^{\frac{1}{2} k^T \Sigma k}
\end{align}
where we completed the square (absorbed $k^T w$ into the quadratic piece) and used the freedom to shift the integration counter. Used with complex $k$ ($k \rightarrow ik$, one recognizes that the above l.h.s. is the Fourier transform of a Gaussian, thereby yielding the familiar result that the Fourier transform of a multivariate Gaussian is an (unnormalized) multivariate Gaussian with the inverse covariance matrix.

\section{Network Definition and Training Protocol}
\label{Sec:TrainingProtocols}
Here we define, very briefly, the mechanical (or algorithm) aspects of deep learning. A more thorough introduction appears in Ref. \cite{nielsenneural}.  

For simplicity, we focus our exposition on fully connected networks (FCNs) with $L$ layers and $N_l$ neurons in each layer. This is the simplest deep neural networks (DNNs) architecture, we will later comment on generalizations to Convolutional Neural Networks (CNNs) and Transformers. We denote by $x \in R^d$ the input of a network, $z_i^{(l)}(x)$ denotes output of the $l$'th layer on the $i$'th neuron, and take $z_i^{l=0}(x)=x_i$. The network is defined recursively by 
\begin{align}
z^{(l)}_{i}(x) &= \sigma(h_{i}^{(l)}(x)+b^{(l)}_i) \\ \nonumber 
h^{(l)}_{i}(x) &= \sum_{j} W^{(l)}_{ij} z^{(l-1)}_j(x) 
\end{align}
where $W^{(l)}_{ij}$ and $b^{(l)}_i$ are respectively the weights and biases (which we often refer to just as weights) and $\sigma(...)$ is the activation function. We encapsulate all weights and biases into a vector $\theta_{\alpha}$ where $\alpha$ is an abstract index spanning all the variables defining the network. The network's output is $f(x)=h^{(L)}(x)$

Considering a supervised learning problem where given a dataset $\mathcal{D} =\{(x_\mu, y_\mu)\}_{\mu=1}^P$ of input and output pairs where the task is either regression or classifications of new examples not in the data set. To achieve this task, most DNNs are trained to minimize a scalar quantity known as the loss function ($\cL$) given by a sum over all training points. 

{\bf MSE loss.} This loss is a common choice for regression problems 
\begin{align}
\mathcal{L} &= P^{-1}\sum_{\mu=1}^P |f(x_{\mu})-y_{\mu}|^2
\end{align}
where $P$ is the number of training points, $y_{\mu} \in R^{d_o}$ (where $d_o$ is the output dimension) is the regression target function (or vector function for $d_o>1$). Alternatively, one may treat $y(x)$ as a one-hot encoding of categorical labels taking $d_o$ distinct values via $[y(x)]_i = \delta_{i c(x)}$ where $c(x) \in [1..d_o]$ assigns a serial number for the category of $x$. Last, in binary classification problems, one can take $d_o=1$ and $y(x)=c(x)$ where $c(x) \in 
\{+1,-1\}$. 

{\bf Cross entropy loss.} This loss is another common choice relevant to classification problems. Here we consider again $f(x)$ as a vector of dimension $d_o$ and write the loss as  
\begin{align}
\cL &= -\sum_{\mu=1}^P \left[\log(e^{-f_{c(x_{\mu})}(x_{\mu})})-\log(\sum_{l=1}^{d_o} e^{-f_o(x_{\mu})})\right]
\end{align}

{\bf Weight decay.} To both these losses, one may add weight decay terms of the form $\gamma_{\alpha}\theta^2_{\alpha}$, to regulate the overall weight norm. 

Many DNNs perform better in the over-parametrized regime where the number of network parameters exceeds the number of training points. For networks with scalar outputs ($d_o=1$), fitting perfectly a single data-point can be seen as placing a single constraint in weight space. Hence in the over-parametrized setting, finding $\theta$'s which fit the training set perfectly is an unconstrained optimization problem. As appreciated even before the recent rise of deep learning \citep{Breiman2018ReflectionsAR} and later supported by various theoretical works  \citep{Dauphin2014,Choromanska2014,Jacot2018} such optimization problems tend to be simple. While different optimization algorithms (e.g. Gradient descent, stochastic Gradient descent (SGD), Adam \cite{kingma2017adammethodstochasticoptimization}) may excel at bringing us faster to the minima or provide some implicit bias towards wider/better minima, the success of deep learning seems to be more generic than the peculiarities of each algorithm. Hence, we find it reasonable to focus on two relatively tractable algorithms.

{\bf Gradient flow.} Is a continuum idealization of the more standard discrete Gradient Descent algorithm. In deep learning jargon, it corresponds to full batch Gradient Descent at vanishing step-size/learning-rate. It is described by the partial differential equation 
\begin{align}
\frac{\dif}{\dif t} \theta_{\alpha}&= -\eta\partial_{\theta_{\alpha}} \cL
\end{align}
where $\eta$ is the learning rate. While in discrete gradient descent, $\eta$ may induce qualitative effects in standard/discrete gradient descent \citep{lewkowycz2020large}, in the above, continuum idealization, it only sets the time scale and may be absorbed into a re-definition of $t$. We note by passing that some effects of finite learning rate can be represented with the continuum description in the form of augmenting the loss with additional gradient terms \citep{Barrett2020,Smith2021}. 

{\bf Langevin dynamics} \citep{Welling2011,williams1996computing}. Another tractable form of dynamics is the following stochastic differential equation 
\begin{align}
\frac{\dif}{\dif t} \theta_{\alpha}&= -\eta\partial_{\theta_{\alpha}} \cL + \sqrt{2 T \eta} \xi_{\alpha}(t) 
\end{align}
where the gradient noise, $\xi_{\alpha}(t)$, is an an uncorrelated Gaussian process obeying $\langle \xi_{\alpha}(t) \rangle = 0, \langle \xi_{\alpha}(t) \xi_{\beta}(t') \rangle = \delta(t-t') \delta_{\alpha \beta}$ where $\delta(t)$ is the Dirac delta function. Unlike gradient flow, the above dynamics loosely mimic the stochastic nature of the dynamics. Using standard results from the theory of Langevin equations, the distribution induced on weight space by this dynamics as $t \rightarrow \infty$ (i.e. equilibrium distribution) is given by \cite{gardiner2010stochastic}
\begin{align}
\label{Eq:BoltzmannDistWeights}
P_{\infty}(\theta) &\propto e^{-\cL/T}
\end{align}

Several doubts could be raised regarding the ability of this process to reach equilibrium at reasonable times. As argued earlier, the ruggedness of the loss landscape is less of a concern in the over-parametrized regime, however diffusion in the high-dimensional near zero loss manifold may potentially result in slow equilibration for some observables, as well as possible entropy barriers. In practice, one finds good support for ergodicity, at least for relevant observables such as network outputs and pre-activations. In particular relaxation times for the loss are of a similar scale as relaxation times for various macroscopic network observable (i.e. involving a summation of many statistically equivalent parts) see for example \cite{naveh2021predicting}. In addition, \cite{LiSompolinsky2021, ariosto2022statistical,naveh2021predicting} show that theoretical descriptions offered by $P_{\infty}(\theta)$ provides accurate predictions on DNNs trained for a reasonable amount of epochs when taking into account the use of full-batch and low learning rates. 

\section{Infinite Random Neural Networks}
\label{Sec:InfiniteRandomDNNs}
Neural networks are, at face value, a very complex ansatz for functions. Grasping the statistical relations between weights and outputs is generally a complex task. Here we discuss some simplifications which occur in the limit of highly over-parametrized neural networks, which would be useful in the next chapters.  

As DNNs, per given weights, define a function, random DNNs define a distribution over functions. Interestingly, for all relevant DNN architectures, there is a suitable infinite-parameter limit where the latter distribution becomes Gaussian (more formally a Gaussian Process or a free field theory). For FCNs, this entails taking the width of all layers to infinity \citep{neal1996priors}, for CNNs the number of channels \citep{novak2018bayesian}, and for transformers the number of heads \citep{hron2020infinite}. Though it would turn out that this limit is rarely reached in practice, it does turn out to be an insightful viewpoint which can be extended to more realistic scenarios.

Consider a DNN with uncorrelated random weights with sub-Gaussian tail, e.g. each weight and bias is i.i.d Gaussian random variable $W_{ij}^{(l)}\sim \mathcal{N}(0, \sigma^2_w)$ and biases $b_i^{(l)}\sim \mathcal{N}(0, \sigma^2_b)$. Starting from the most upstream layer, the input layer, $h^{(0)}_i(x) = w_i \cdot x$, one finds that $h^{(0)}_i(x)$, per given $x$, are uncorrelated random variables. Consequently, by induction $h_i^{(l)}(x)=\sum_{j=1}^N W^{(l)}_{ij} \sigma(h^{(l-1)}_{j}(x))$ appear as a sum of $N$ uncorrelated random variables. Thus, for reasonable choice of $\sigma(..)$ and properly normalized $W^{(1)}$, the preactivation $h_i^{(1)}(x)$, per $x$, would be uncorrelated Gaussian variables. The proper normalization is to take a variance of $\sigma_w^2 = 1/N$ for $W^{(1)}_{ij}$ (more generally one over the fan-in or number of signals coming in), which is a common way to initialize DNNs. 

Following the above argument, we see that the output of a neural network as $N\rightarrow \infty$, per $x$, is a Gaussian random variable. Let us obtain the variance of this random variable, and at the same time, the correlations between the outputs of the network on two different data points. Repeating the previous line of argument, the two dimensional vector random variable $[h^{(0)}_i(x_1),h^{(0)}_i(x_2)]$ is a two-dimensional multivariate Gaussian since it is linear in $w_i \in R^d$ which we assume is Gaussian. The correlations $\langle h^{(0)}_i(x_{\alpha}) h^{(0)}_{j}(x_{\beta})\rangle_{w_i,w_j \sim {\cal N}(0,d^{-1}I_{d\times d}]}=\delta_{ij} d^{-1} (x_{\alpha} \cdot x_{\beta})$, with $\alpha,\beta \in \{1,2\}$. We denote the resulting distribution for $[h^{(0)}_i(x_{1}),h^{(0)}_i(x_{2})]$, per any neuron index $i$, as ${\cal N}(0,K^{(0)})$, namely a two-dimensional centered Gaussian distribution with a covariance matrix $K^{(0)}=d^{-1} \begin{pmatrix}
|x_1|^2 & x_1 \cdot x_2 \\
x_2 \cdot x_1 & |x_2|^2
\end{pmatrix}$

Continuing downstream, the vector of random variables $[h^{(1)}_i(x_1),h^{(1)}_i(x_2)]$ can be written as 
\begin{align}
[h^{(1)}_i(x_1),h^{(1)}_i(x_2)] &= \sum_{j=1}^N W^{(1)}_{ij} [\sigma(h_j^{0}(x_1)),\sigma(h_j^{0}(x_2))]
\end{align}
is given as a summation of $N$ uncorrelated 2d vector random variables. Hence, it is again Gaussian by the vector version of the central limit theorem. Furthermore, by definition, there is no linear correlation between different $i$'s (due to $W_{ij}^{(1)}$), which for Gaussian variables means no correlation at all. Thus its distribution is fully determined by its average and correlation specifically on any specific $i$. The average $\langle [h^{(1)}_i(x_1),h^{(1)}_i(x_2)] \rangle_{W^{(1)}_{ij} \sim {\cal N}(0,N^{-1}I_{N \times N}),w_{1..N} \sim {\cal N}[0,d^{-1} I_{d\times d}]}$ is easily seen to be zero, and the $2 \times 2$ correlations matrix $K^{(1)}_{\mu \nu}$ given by 
\begin{align}
\label{Eq:NNGP_K_1}
K^{(1)}_{\mu \nu} &= \left\langle h^{(1)}_i(x_{\mu})h^{(1)}_i(x_{\nu}) \right\rangle_{W^{(1)}_{ij} \sim {\cal N}(0,N^{-1}I_{N \times N}),w_{1..N} \sim {\cal N}[0,d^{-1} I_{d\times d}] } \\ \nonumber &= \left\langle h^{(1)}_i(x_{\mu})h^{(1)}_i(x_{\nu}) \right\rangle_{W^{(1)}_{ij} \sim {\cal N}(0,N^{-1}I_{N \times N}),h^{(0)} \sim {\cal N}(0,K^{(0)})} \\ \nonumber 
&= \frac{1}{N}\sum_{j=1}^N \int d h^{(0)}_{j}(x_\mu) d h^{(0)}_{j}(x_\nu) \; {\cal N}(0,K^{(1)};h_{j}^{(0)}(x_.)) \; \sigma(h^{(0)}_j(x_{\mu}))\sigma(h^{(0)}_j(x_{\nu})) \\ \nonumber 
&= \int d h^{(0)}_1(x_\mu) d h^{(0)}_1(x_\nu) \; {\cal N}(0,K^{(0)};h^{(0)}) \; \sigma(h^{(0)}_{1}(x_{\mu})) \sigma(h^{(0)}_{1}(x_{\nu}))
\end{align} 
where in passing between the first and second line, we used the fact that $h^{(1)}$ depends on the $w_{1..N}$ only through $h^{(0)}$, to replace the average under the inputs weights with an average over $h^{(0)}$. The resulting two-dimensional integral turns out to be solvable for various activation functions in particular ReLU \citep{Saul2009,lee2017deep} and Erf \citep{williams1996computing}. For instance, in the latter case one obtains 
\begin{align}
\label{Eq:ErfKernel}
K^{(1)}_{\mu \nu}  \equiv K^{(1)}(x_{\mu},x_{\nu})  &= \frac{2}{\pi} \sin^{-1}\left[\frac{2 K^{(0)}_{\mu \nu}}{\sqrt{1+2 K^{(0)}_{\mu \mu}}\sqrt{1+2K^{(0)}_{\nu \nu}}}\right]
\end{align}
More generally, the above reasoning now extends naturally to the downstream layer, with $(1)\rightarrow(2)$ and $(0) \rightarrow (1)$. Thus we have a concrete formula for the correlations between the outputs on $x_1,x_2$ for an arbitrarily deep ReLU or Erf network. 

Next, we wish to extend this to any finite number of points ($x_1...x_{P'}$). At large enough $N$ (in particular $N \gg P'$) each pre-activation as well as the output remains a multivariate $P'$-dimensional Gaussian and hence solely defined by the correlation between each two data-points--- the same correlation described by the above formula. Thus as $N\rightarrow \infty$, we find that the distribution induced by the DNN on function space is such that when examined on any finite set of points--- it behaves as a multivariate Gaussian with a kernel $K^{(L)}_{\mu \nu}$. In physics, we call such distributions over functions free field theories. The advantage of the latter, a less rigorous but practically exact viewpoint, is that it transcends the mathematical description based on the awkward notion of all possible finite sets of points, to the simple physical notion of a continuum. It also offers a convenient starting point for analyzing the effect of non-linearities via perturbation theory, saddle point, and mean-field approaches, as discussed below. 

\section{Operator algebra on spaces with measures}
\label{Sec:Intro_Measure}
As is common with Gaussian distributions, the eigenvectors and eigenvalues of the covariance matrix play a useful role. As alluded to in the above section, our real object of interest is an extreme case of a multivariate Gaussian distribution, wherein instead of considering a Gaussian distribution over vectors, we consider a Gaussian distribution over an infinite-dimensional vector or equivalently a smooth function. 

To substantiate this viewpoint and explain what are the coefficients of this infinite vector, consider the following eigenvalue problem 
\begin{align}
\label{Eq:EigenvalueEquation}
\int \dif x' p(x') K(x,x') \phi_{k}(x') &= \lambda_k \phi_k(x)  
\end{align}
where $p(x')$ is some distribution over inputs and we shall soon denote $\dif x \; p(x)=\dif \mu_x$. Given that (i) $p(x')$ decays sufficiently fast at infinity, (ii) $K(x,x')=K(x',x)$, and (iii) $\int \dif \mu_{x'} \dif\mu_x K(x,x') g(x) g(x') \geq 0$ for all real functions $g(x)$ with finite norm ($\int \dif \mu_x g^2(x) < \infty$) we have that (a) the above eigenvalue problem yields a discrete set of solutions $(\lambda_k,\phi_k(x))$ with $\lambda_k$ being non-negative real numbers (b) $\phi_k(x)$ may be chosen normalized and orthogonal namely $\int \dif\mu_x \phi_{k}(x) \phi_{k'}(x)=\delta_{kk'}$ (c) $K(x,x')=\sum_{k=1}^{\infty} \lambda_k \phi_k(x) \phi_k(x')$ (Mercer's theorem). (d) $\phi_k(x)$ define a complete basis for function space namely we may express any $g(x)$ via $\sum_k g_k \phi_k(x)$ where $g_k = \int \dif\mu_x \phi_k(x) g(x)$. 

Notably, kernels of DNNs obey requirements $(ii),(iii)$. Concerning point $(i)$ above, input datasets, typically have a natural scale and decay to zero beyond some value (e.g., the brightness of a pixel cannot be larger than 255, or the input amplitude of a microphone cannot exceed 150db). 

Next, we define several additional linear algebra notions on kernels. In particular, when all $\lambda_k > 0$, the inverse kernel $K^{-1}(x,x')$ can be defined and is given by 
\begin{align}
K^{-1}(x,x') &= \sum_k \lambda_k^{-1} \phi_k(x) \phi_{k}(x')  
\end{align}
Noting further that $\int \dif x' p(x') \sum_k \phi_k(x) \phi_k(x') g(x')= g(x)$ we find that 
\begin{align}
\sum_k \phi_k(x) \phi_{k}(x') &= \delta(x-x')/p(x') 
\end{align}
using the above we find an equivalent definition of $K^{-1}(x,x')$
\begin{align}
\label{Eq:Alt_Inv}
\int \dif\mu_{x'} K(x,x') K^{-1}(x',x'') &= \delta(x-x'')/p(x) 
\end{align}

Notably while $K(x,x')$ is given and independent of the measure $p(x)$, $K^{-1}(x,x')$ does depend on it, yet in a predictable way: Given $K^{-1}(x,x')$ defined w.r.t. the measure $p(x)$ and $\tilde{K}^{-1}(x,x')$ defined w.r.t. the measure $\tilde{p}(x)$ we find
\begin{align}
\frac{p(x)}{\tilde{p}(x)}K^{-1}(x,x') \frac{p(x')}{\tilde{p}(x')}&= \tilde{K}^{-1}(x,x')
\end{align}
which can be verified via 
\begin{align}\int &\dif\tilde{\mu}_{x'}K(x,x') \tilde{K}^{-1}(x',x'')=\int \dif\tilde{\mu}_{x'}K(x,x')\frac{p(x')}{\tilde{p}(x')} K^{-1}(x',x'')\frac{p(x'')}{\tilde{p}(x'')}\\ \nonumber &=\int \dif\mu_{x'}K(x,x') K^{-1}(x',x'')\frac{p(x'')}{\tilde{p}(x'')}=\frac{\delta(x-x'')}{\tilde{p}(x'')}\end{align} where $\dif\tilde{\mu}_x=\dif x \; \tilde{p}(x)$. We thus find that $\tilde{K}^{-1}$ obeys Eq. \ref{Eq:Alt_Inv}.  

An interesting corollary of this is that the so-called Reproducing Kernel Hilbert Space (RKHS) norm of a function $g(x)$ 
\begin{align}
\label{Eq:RKHS}
|g|_K^2 \equiv \int \dif\mu_x \dif\mu_{x'} g(x) g(x') K^{-1}(x,x')
\end{align}
is in fact the same for any two measures with the same support, loosely speaking we mean that $\infty>p(x)/\tilde{p}(x) > 0$ for any $x$. For more details, see \cite{Rasmussen2005} section 6.1 and references therein.

\section{ Symmetries: Equivariance and Invariance} Kernels of DNNs are typically highly symmetric objects. For instance, as can be inferred from our previous computation for the kernel-- any DNN with a fully-connected input layer would have a kernel which depends only on $x \cdot x',|x|,|x'|$. As a result, it would be symmetric to rotations of the input. More formally let $O$ be some symmetry action on $x$ ($d \times d$ rotation matrix in the above example), we say that a kernel is equivariant under $O$ when $K(x,x')=K(Ox,Ox')$. We say that the kernel is invariant under $O$ when $K(x,Ox')=K(Ox,x')=K(x,x')$. Intuitively, equivariance means that the RKHS norm of a function ($f(x)$) is equal to that of $f(Ox)$ where invariance means non-symmetric functions are outside-RKHS/inexpressible using that kernel. Note also that invariance implies equivariance. 

We comment that these notions of equivariance and invariance coincide with those used in DNN nomenclature (e.g. \cite{novak2018bayesian}). Indeed, equivariance means that the action of $O \in O(d)$ on $x$ has some non-trivial action ($\tilde{O} \in O(N)$) on the $c-$index vector $h^{(l)}_{.}(x)$ namely that $h^{(l)}_{.}(Ox)=\tilde{O}h^{(l)}_{.}(x)$ whereas invariance means $\tilde{O}$ is the identity. Using the definition of the kernel of the $l$'th layer as the average of $[h^{(l)}_{.}(x)]^T h^{(l)}_{.}(x')$ (over weights) one recovers the above definitions for symmetries which are a subset of the orthogonal group (e.g. permutations, reflections, rotations on convolutional patches). \footnote{Scale invariance, present in ReLU networks without biases, has a different manifestation which does not fall strictly within the above two categories, specifically $K(x,cx')=cK(x,x')$).} 

We comment that one can map this notion of symmetry to that used in quantum mechanics. Specifically, define the operator associated with $O$ ($\tilde{O}$) via the action $\tilde{O} f(x)=f(Ox)$. This then implies the right action $OK(x,x')=K(Ox,x')$. Next using the action of $K$ on function ($\int \dif\mu_{x'} K(x,x') f(x')$) we find the left action via $\int \dif\mu_{x'}K(x,x') \tilde{O}f(x')=\int \dif\mu_{x'}K(x,x')f(Ox')=\int \dif\mu_{O^T x'}K(x,O^T x')f(x')$. Conditioned on the measure being symmetric ($\dif\mu_{O^T x'}=\dif\mu_{x'}$) we find $K(x,x')\tilde{O}=K(x,O^T x')$ \footnote{For non-symmetric measure the ratio $\dif\mu_{x'}/\dif\mu_{O^T x'}=p(x')/p(O^T x')$ appears as a factor} and therefore $\tilde{O} K(x,x') \tilde{O}^T=K(Ox,Ox')=K(x,x')$ as the definition of a symmetry which coincides with that used in quantum mechanics, when viewing $K$ as the Hamiltonian. 

Much like in quantum physics, symmetries have direct implications on the eigenfunctions and eigenvalues of a kernel. However, at least in non-relativistic physics, we often assume that space itself possesses all symmetries. Here the situation is different since the measure, $\dif\mu_x=\dif x \; p(x)$ need not be symmetric. In case it is, namely when $p(x)=p(Ox)$, we find the following simplifications: Equivariance implies that eigenfunctions can be grouped into irreducible representations (irrep) of the symmetry group (e.g. the rotation group for an FCN). All eigenfunctions which are part of the irrep would have the same eigenvalue. Invariance implies equivariance, hence the above conclusions hold but in addition, one finds that all non-trivial irreps of the symmetry have a zero eigenvalue. Given that irreps for standard symmetry groups are well known (e.g. spherical-harmonics for $O(3)$ and hyper-spherical harmonics for $O(d)$), these considerations simplify the diagonalization of kernels when both kernel and measure are symmetric. For instance, any DNN having a fully-connected first layer together with a measure that is uniform on a hyper-sphere in input space, yields an exactly diagonalizable kernel with hyper-spherical harmonics as eigenfunctions (e.g. \cite{cohen2021learning}). 

\section{Essentials of path integrals}
\label{Sec:PathIntegrals}
Path integrals are an extreme version of multivariable integrals and an underlying formalism for field theory \cite{schulman1996}. Multivariable integrals are needed when we integrate over functions having a multi-variable argument. Path integrals are required when we are integrated over functionals-- functions which take a function as their argument.  

In physics, we often encounter functionals-- say the elastic energy associated with the height configuration/field ($f(x,y)$) of a 2d membrane which we denote as $E[f]$ (from now on, we use square brackets (e.g. $A[..]$) to denote $A$ as functional and angular brackets if $A$ is a single or multi-variable function). At equilibrium, statistical physics tells us that the probability of finding the membrane at height configuration $f(x,y)$ is {\it proportional to} $e^{-E[f]/(k_B T)}$ where $T$ is the temperature and $k_B$ is Boltzmann's constant. However, to discuss the probability density of $f$ ($P[f]$) or calculate averages such as the mean value of $f$ at $(x_0,y_0)$, we need to integrate over all possible $f$ configurations. Developing a formal way of doing so, especially in cases where $E[f]$ is not a simple quadratic form in $f$, is the main purpose of path integrals. Alongside this, path integrals provide a unified formalism for statistical field theory, stochastic processes, quantum mechanics, and quantum field theory and aid the transfer of knowledge between these diverse fields. 

Turning to deep learning, it is natural to consider the output of the network $f({x})$, where ${x\in \mathbb{R}^d}$ is the network input, in analogy with the height/field configuration of the membrane. Analogous to the Boltzmann distribution $e^{-E[f]/(k_bT)}$, we may now consider several natural distributions on the output function, $f$, of the network: {\bf (i)} The prior distribution i.e. the distribution of $f$ induced by choosing suitably normalized random Gaussian weights, {\bf (ii)} The posterior distribution i.e. the distribution of $f$ induced by the Bayesian posterior (i.e. the distribution obtained multiplying the prior distribution by the probability of seeing the training set as the output of the model and normalizing the product), or {\bf (iii)} The distribution induced on $f$ by training an ensemble of neural networks, with different weights drawn from the initialization ensemble and different algorithmic noise. 

Path integrals can be formally defined as an infinite-variable limit of multi-variable integrals. The most common way is to discretize the space (${x}$) over some fine grid. Concretely, let us assume ${x}$ does not exceed some $[-L,L]^d$ hypercube and allow each real coordinate $[{x}]_i$ to take on only the values $-L + n/\Lambda$ with $n$ spanning the integers in the interval $[0,2 \floor*{L \Lambda}]$. Limiting space in this manner, $f$ is equivalent to a vector, $f_{\Lambda}$ of dimension $d_{\Lambda}=(2 \floor*{\Lambda L})^d$ with each coordinate of $f_{\Lambda}$ marking its value on one point on the grid. Next, we need to map our functional of interest (say $E[f]$), into a multivariate function $E(f_{\Lambda})$. The standard way of doing so is to trade derivatives with discrete derivatives and integrals with summations. \footnote{In cases where the typical configurations of $f$ are not smooth beyond some small length scale, as is the case in stochastic processes or quantum field theories with Gauge fields, the ambiguities of how to define discrete derivatives matter. However, in our case, they do not, since as we shall soon see our $E[f]$ involves integrals and not derivatives.} Last we consider some quantity of interest, say $f({x}_0)$, quantize ${x}_0$ to the nearby grid point (${x}_{g0}$), and calculate its average via $\int \dif^{d_{\Lambda}} e^{-E(f)} f({x}_{g0})/\int \dif^{d_{\Lambda}} e^{-E(f)}$. Finally, we take the limit of $\Lambda,L \rightarrow \infty$ and treat that as the path integral. In practice, this underlying machinery is kept under the hood, and, following some general identities of path integrals we may derive in this manner, we work directly in the continuum and make no reference to a grid. 

Let us demonstrate this procedure for the case where $f$ is the output of a network with random weights in the infinite-width limit. Let us further enumerate the grid points with a single index (${x}_p$, $p \in 1..d_{\Lambda}$). Four relevant objects here are the network kernel matrix, or correlations matrix between outputs, $K_{pp'}=K({x}_p,{x}_{p'})$ and its inverse $K^{-1}_{pp'}$ as well as the operator versions of these namely $K({x},{x}')$ and $K^{-1}({x},{x}')$, which we assume are all smooth function.

We turn to obtain the path integral formulation of a Gaussian Process as a limit of an infinitely fine discretization grid. Based on the smooth assumptions discussed above we have that 
\begin{align}
\lambda \phi_{\lambda}(x_b) &= \int \dif\mu_z K(x_b,z) \phi_{
\lambda}(z) \approx d_{\Lambda}^{-1} \sum_{a=1}^{d_{\Lambda}} K(x_b,z_a) \phi(z_a) \\ \nonumber
\phi_{\lambda}(x_b) &= \int \dif\mu_z \dif\mu_w K^{-1}(x_b,z)K(z,w) \phi_{
\lambda}(w) \\ \nonumber 
&\approx d_{\Lambda}^{-2} \sum_{a,c=1}^{d_{\Lambda}} K^{-1}(x_b,z_a) K(z_a,w_c) \phi(w_c) 
\end{align}
where $z_a, a \in 1..d_{\Lambda}$ goes over all grid points as does $w_c$, $x_b$ is some grid point, and factors of $d_{\Lambda}$ come from turning an integral to a sum. The first equation shows that the eigenfunctions of the operator $K(x,z)$, when sampled on the grid, are the eigenvectors of the $d_{\Lambda} \times d_{\Lambda}$ matrix $K_{ba}=K(x_b,z_a)$ however with eigenvalues multiplied by $d_{\Lambda}$. This is what we mean by  asymptotic association $K(x,z) \rightarrow d_{\Lambda}^{-1}K(x_a,z_b)$. The second equation then implies that the matrix obtained by the discretization of the operator $K^{-1}(x,z)$ acts as the matrix inverse of $K_{ab}$ up to a factor of $d_{\Lambda}^2$, namely $K^{-1}(x_a,z_b) \rightarrow d_{\Lambda}^{2}[K^{-1}]_{ab}$. Given these associations we may write the log probability of the GP prior as  
\begin{align}
&\sum_{ab} f(x_a) [K^{-1}]_{ab} f(x_b) \rightarrow \sum_{ab} d_{\Lambda}^{-2}  f(x_a) K^{-1}(x_a,x_b) f(x_b) \\ \nonumber  &\approx \int \dif\mu_x \dif\mu_{x'} f(x) K^{-1}(x,x') f(x') + Const\\ \nonumber 
\end{align}
We thus find the following path integral formulation of the probability induced on function space by an infinitely over-parameterized DNN 
\begin{align}
P[f] &= Z^{-1} e^{-\frac{1}{2} \int \dif\mu_x \dif\mu_{x'} \,f(x) K^{-1}(x,x') f(x')}
\end{align}
where the normalization factor, also known as a partition function, is given by the following path integral over the function space $f$ ($\int Df$)
\begin{align}
Z &= \int Df  e^{-\frac{1}{2} \int \dif\mu_x \dif\mu_{x'} \, f(x) K^{-1}(x,x') f(x')}
\end{align}
where $\int Df$, the path integral over $f$, is defined via the limit $Df = \lim_{L,\Lambda\to \infty}\prod_{i=1}^{d_{\Lambda}} \dif f_i$. 

Borrowing physics computation style, we would, from now on, focus mainly on $\log(Z)$ together with so-called source terms ($\alpha(x)$ dependent terms), as our main object of interest. Specifically, we will define 
\begin{align}
Z[\alpha] &= \int Df \; e^{-S[f,\alpha]}\\ \nonumber 
S[f,\alpha] &= \frac{1}{2} \int \dif\mu_x \dif\mu_{x'}\, f(x) K^{-1}(x,x') f(x')+\int \dif x \; \alpha(x) f(x)
\end{align}
where we also introduced, borrowing jargon from field-theory, the ``action" (or negative log probability) $S[f,\alpha]$. Next we introduce the functional derivative $\delta_{\alpha(x)}$ which obeys 
\begin{align}
\delta_{\alpha(x)} \alpha(y) &= \delta(x-y)
\end{align}
to calculate the statistics of $f(x)$ under $P[f]$ we use the chain rule as such 
\begin{align}
&\delta_{\alpha(x_0)} \log(Z[\alpha])|_{\alpha=0} = Z^{-1} \delta_{\alpha(x_0)} Z[\alpha]\mid_{\alpha=0} = \\ \nonumber  &Z^{-1} \int Df  e^{-\frac{1}{2} \int \dif\mu_x \dif\mu_{x'} \,f(x) K^{-1}(x,x') f(x')} \delta_{\alpha(x_0)}\int \dif x \alpha(x) f(x) \mid_{\alpha=0} \\ \nonumber 
&= Z^{-1} \int Df  e^{-\frac{1}{2} \int \dif\mu_x \dif\mu_{x'} \, f(x) K^{-1}(x,x') f(x')}\int \dif x \delta(x_0-x) f(x) \\ \nonumber 
&= Z^{-1} \int Df  e^{-\frac{1}{2} \int \dif\mu_x \dif\mu_{x'} \, f(x) K^{-1}(x,x') f(x')}f(x_0) \\ \nonumber
&= \int Df P[f] f(x_0)
\end{align}
where the second term in the exponential vanishes at $\alpha=0$.

Similarly, one can show that second order functional derivatives provide the second cumulants (correlations) namely  
\begin{align}
\delta_{\alpha(x_0)}\delta_{\alpha(x_1)} \log(Z[\alpha])|_{\alpha=0} &= \int Df f(x_0) f(x_1) -  \int Df f(x_0) \int Df f(x_1)
\end{align}

\noindent we comment that $\log(Z[\alpha])$ can be understood as a path integral version of a multivariate cumulant generating function (which however omits an inconsequential constant, since $\log(Z[0])$ need not be $1$). In physics $-\log(Z)$ times $k_B T$ is known as the free energy. 

One useful identity by which one can evaluate averages as the one above is 
\begin{align}
Z[\alpha] &= \int Df  e^{-\frac{1}{2} \int \dif\mu_x \dif\mu_{x'} f(x) K^{-1}(x,x') f(x')+\int \dif x \alpha(x) f(x)} \\ \nonumber 
&= \left[\int Df' e^{-\frac{1}{2} \int \dif\mu_x \dif\mu_{x'} f'(x) K^{-1}(x,x') f'(x')}\right] e^{\frac{1}{2} \int \dif\mu_x \dif\mu_{x'} \alpha(x) K(x,x') \alpha(x')} \\ \nonumber 
&= e^{\frac{1}{2} \int \dif\mu_x \dif\mu_{x'} \alpha(x) K(x,x') \alpha(x')+Z[0]}
\end{align}
where we perform square-completion defining $f'(x) = f(x)-\int \dif\mu_{x'} K(x,x') \alpha(x')$ and in the penultimate line we used freedom in defining the integration contour to cancel $\alpha$ dependence of the functional integral on the 2nd line. A corollary of this identity is that the correlation between $f(x_0)$ and $f(x_1)$ under the path integral is 
\begin{align}
&\delta_{\alpha(x_0)} \delta_{\alpha(x_1)}\log(Z[\alpha]) = \delta_{\alpha(x_0)} \delta_{\alpha(x_1)}\log\left( 
e^{\frac{1}{2} \int \dif\mu_x \dif\mu_{x'} \alpha(x) K(x,x') \alpha(x')+Const} \right) \\ \nonumber 
&= \delta_{\alpha(x_0)} \delta_{\alpha(x_1)} \left( \frac{1}{2} \int \dif\mu_x \dif\mu_{x'} \alpha(x) K(x,x') \alpha(x')\right) = \frac{1}{2}[K(x_0,x_1)+K(x_1,x_0)] \\ \nonumber 
&=K(x_0,x_1)
\end{align}
which can also be viewed as a sanity check for this formalism. Notably, since we shall always be interested in evaluating such functional derivatives at $\alpha=0$, we kept this requirement implicit in the above and from now on.

Another useful identity is the spectral representation of the path integral. First note that due to the orthogonality of eigenfunction under the measure (see below Eq. (\ref{Eq:EigenvalueEquation})) we have that
\begin{align}
 \int \dif\mu_x \dif\mu_{x'} f(x) K^{-1}(x,x') f(x') &= \sum_{k=1}^{\infty} f_k \lambda_k^{-1} f_k \\ \nonumber 
f_k &= \int \dif\mu_x f(x) \phi_{\lambda_k}(x)
\end{align}
with $\lambda_k, \phi_{\lambda_k}(x)$ be the eigenvalues and eigenfunction of the kernel $K$. Further, since we do not need to pay attention to constant factors multiplying ($Z[\alpha]$), as those would anyway be lost in any functional derivative of $\log(Z[\alpha])$ we may ignore the Jacobian (which is anyways one for any finite discretization), associated with changing the integration variable from $Df$ to $\Pi_{k=1}^{\infty}d f_k$ and obtain 
\begin{align}
Z[\alpha] &\propto \int \Pi_k \dif f_k \; e^{-\frac{1}{2}\sum_{k=1}^{\infty} f_k \lambda_k^{-1} f_k + \int \dif\mu_x \alpha(x) f(x)} \\ \nonumber 
&= \int \Pi_k \dif f_k \; e^{-\frac{1}{2}\sum_{k=1}^{\infty} f_k \lambda_k^{-1} f_k + \sum_{k=1}^{\infty} \alpha_k f_k}
\end{align}
where, in the last line, we used again the orthogonality relations together with the completeness relation $f(x) = \sum_k f_k \phi_k(x)$ and $\alpha(x) = \sum_k \alpha_k \phi_k(x)$.

\section{Approximating path integrals}
Evaluating path integrals or multivariable integrals is generally impossible apart from the cases where the action (negative log probability) is at most quadratic in the integration variables (i.e. Gaussian). The typical action would, however, be non-Gaussian, requiring an extra set of tools. Here we will cover three basic tools: Perturbation theory, mean-field theory, and saddle point methods.  

\subsection{Perturbation theory}
\label{Sec:PT}
Consider an action of the form 
\begin{align}
S[f] &= S_0[f] + U[f]
\end{align}
where $S_0[f]$ is quadratic in $f$ (e.g. $\int \dif\mu_x \dif\mu_y f(x) K^{-1}(x,y) f(y)$) and $U[f]$, henceforth the {\it interaction}, is non-quadratic, and, for simplicity, has some finite power of $f$ (e.g. $\int \dif\mu_{x_1}..\dif\mu_{x_4}U(x_1..x_4) f(x_1)..f(x_4)$). We assume the Gaussian part of the action ($S_0[f]$) is solved so we know the mean and variances under that action 
\begin{align}
\langle f(x) \rangle_{S_0} &= \bar{f}(x) \\ \nonumber 
\langle [f(x)-\bar{f}(x)] [f(y)-\bar{f}(y)] \rangle_{S_0} &= K(x,y)
\end{align}
We wish to evaluate the average value of some observable (e.g. $f(x)f(y)$) under that action as a series expansion in $U(x_1..x_4)$. 
The answer, up to the second order, can be formally written as a Taylor expansion of the interaction term exponent
\begin{align}
\langle f(x) f(y) \rangle &= \frac{Z_{U=0}}{Z}\langle f(x) f(y) e^{-U[f]} \rangle_{S_0} = \langle f(x) f(y) \rangle_{S_0} - \langle f(x) f(y) U[f] \rangle_{S_0,\con} \\ \nonumber &+ \frac{1}{2!} \langle f(x) f(y) U[f] U[f]\rangle_{S_0,\con} - ...
\end{align}
where we introduced connected average notation ($\langle .. \rangle_{S_0,\con}$), which accounts for the normalization of the partition function, and is defined as follows (cf. \citep{K_hn_2018} App. A.): Noting that the averages are under Gaussian distribution, one first considers all terms contributing to this average according to Eq. (\ref{Eq:WickWithAverage}). Next, one omits the ``disconnected'' contributions to the sum in that equation. A disconnected contribution is one in which one or both of the following two conditions are met: {\bf (i)} The $f$ factor in the observable are not paired through $P$ with any $f$ factors in the interaction term. {\bf (ii)} At least one of the $U[f]$ factors are not paired with $f$ factors in the observable or in another $U[f]$. The sum of all remaining terms (ones in which all the different $U[f]$ terms as well as the observable are connected through pairs) constitute the connected average. For instance, in the above example, assuming for simplicity that $\bar{f}(x)=0$, we obtain 
\begin{align}
&\langle f(x) f(y) U[f] \rangle_{S_0,\con} = \int \dif\mu_{x_1}..d \mu_{x_4} U(x_1..x_4) \langle f(x) f(y) f(x_1)..f(x_4) \rangle_{S_0,\con} \\ \nonumber 
&\langle f(x) f(y) f(x_1)f(x_2)f(x_3)f(x_4) \rangle_{S_0,\con} = K(x,x_1)K(y,x_2) K(x_3,x_4)\\ \nonumber 
&+ K(x,x_1)K(y,x_3) K(x_2,x_4) + K(x,x_1)K(y,x_4) K(x_3,x_3) \\ \nonumber &+ K(x,x_2)K(y,x_1) K(x_3,x_4) + K(x,x_2)K(y,x_3) K(x_1,x_4) \\ \nonumber 
&+ K(x,x_2)K(y,x_4) K(x_1,x_3) +  K(x,x_3)K(y,x_1) K(x_2,x_4)\\ \nonumber 
&+ K(x,x_3)K(y,x_2) K(x_1,x_3) +  K(x,x_3)K(y,x_4) K(x_1,x_2) \\ \nonumber 
&+ K(x,x_4)K(y,x_1)K(x_2,x_3) +  K(x,x_4)K(y,x_2) K(x_1,x_3)\\ \nonumber 
&+  K(x,x_4)K(y,x_3) K(x_1,x_2)
\end{align}
which can be simplified into a single term times $12$, in the typical case where $U(x_1..x_4)$ is permutation symmetric in four arguments. 

Notwithstanding, going to higher orders order and/or having $\bar{f}\neq 0$ quickly generates many terms and becomes intractable. Perturbation theory is thus mainly useful in the following three circumstances: {\bf (i)} To perfect an already nearly-accurate result at zeroth order {\bf (ii)} As means of estimating when the zeroth order approximation starts breaking down {\bf (iii)} In scenarios which we will encounter, partial re-summations-- summing a certain simple subset of pairing choices to infinite order in $U[f]$, may allow tracking strong $U[f]$ effects accurately. For instance, it can be checked that keeping only pairings in which two $f$'s from each $U[f]$ get paired with themselves amounts to an effective change of $K(x,y)$. However, whether this subset of pairings dominates over all the exponentially many other pairings, would depend on the particularities of the problem. 

\subsection{Mean-field}
\label{ssec:mean_field}
A different technique that can capture strong $U[f]$ effects is taking a mean-field. The idea here is to identify a certain combination of $f$ within the interaction term which is likely to be weakly fluctuating under the path-integral. Typically, these would be coherent (equal sign) sums/integrals over many weakly correlated variables. Next, one replaces these with their expected mean value. This expected mean value is the value of this combination in the theory in which this replacement was made. Typically, the replacement should be such that it results in a tractable theory under which this mean value can be calculated. 

Let us demonstrate this seemingly circular logic in action through a concrete example. Consider an interaction of the form 
\begin{align}
U[f] &= u\left(\int \dif\mu_{x_1} f^2(x_1)\right)\left(\int \dif\mu_{x_2} f^2(x_2)\right) 
\end{align}
Next, one recognizes that both terms are an integration of a positive variable ($f^2(x)$) and hence coherent. Furthermore, within the free ($S_0$) theory, each such term is $\sum_{k=1}^{\infty}
f_k^2$. In cases where many $f_k$'s contribute significantly to the sum, we expect the mean to be much larger than the variance. Following this, we make the replacement 
\begin{align}
U[f] &= u\left[\left\langle \left(\int \dif\mu_{x_2} f^2(x_2)\right) \right\rangle_{S_{\MF}}+\Delta[f]\right]^2 \\ \nonumber 
\Delta[f] &= \int \dif\mu_{x_2} f^2(x_2)-\left\langle \left(\int \dif\mu_{x_2} f^2(x_2)\right) \right\rangle_{S_{\MF}}
\\ \nonumber 
S_{\MF} &\equiv S_0 + 2 u \Delta[f]\left\langle \left(\int \dif\mu_{x_2} f^2(x_2)\right) \right\rangle_{S_{\MF}} + \const  \\ \nonumber 
&= S_0 + 2 u \int d \mu_x f^2(x)\left\langle \left(\int \dif\mu_{x_2} f^2(x_2)\right) \right\rangle_{S_{\MF}} + \const 
\end{align}
where by $\langle ...\rangle_{S_{\MF}}$ is an expectation with respect to the measure generated by the above mean-field action $S_{\MF}$ (which neglects $O(\Delta^2)$ contributions), and therefore will soon be evaluated self consistently.  The mean-field approximation is to neglect the $\Delta[f]^2$ contribution and accordingly use the mean-field action ($S_{\MF}$) given by the original action without this term. Neglecting $\Delta[f]^2$ is justified as subtracting the average makes $\Delta[f]$ an incoherent sum (i.e. $\sum_k f_k^2 -\langle f^2_k \rangle_{S_{\MF}}$). As $f_k$'s are uncorrelated under $S_{\MF}$ and, as assumed, many $k$ modes contribute $\Delta[f]^2$ would be smaller by a factor of one over the effective number of contributing modes. Of course, for this to hold, we should subtract the correct average--- the average under the mean-field action. We turn to obtain this quantity. 

Since $S_0[f]$ was assumed Gaussian and the mean-field contribution of the interaction to the action is quadratic, $S_{\MF}[f]$ remains Gaussian and given by 
\begin{align}
S_{\MF} &= \frac{1}{2} \int \dif\mu_{x_1} \dif\mu_{x_2} f(x_1) K^{-1}(x_1,x_2) f(x_2) \\ \nonumber 
&- \int \dif\mu_{x_1} \dif\mu_{x_2} f(x_1) K^{-1}(x_1,x_2) \bar{f}(x_2) + 2u M \int \dif\mu_x f^2(x) 
\end{align}
where $K(x_1,x_2)$ and $\bar{f}(x_1)$ are, respectively, the covariance and mean which define $S_0$ and $M$, the mean-field, equals $\left\langle \left(\int \dif\mu_{x_2} f^2(x_2)\right) \right\rangle_{S_{\MF}}$. We may also write the last term as $2u M \int \dif\mu_{x_1} \dif\mu_{x_2} f(x_1) f(x_2) \delta(x_1-x_2)/p_{data}(x_1)$. By a standard splitting of the second moment, we have 
\begin{align}
\label{Eq:MF_1}
&\left\langle \left(\int \dif\mu_{x_2} f^2(x_2)\right) \right\rangle_{S_{\MF}} = \int \dif\mu_{x_2}  [K_{\MF}(x_2,x_2)+ \langle f(x_2) \rangle_{S_{\MF}}^2] 
\end{align}
where $K_{\MF}$, the covariance of the mean-field action, is the inverse operator to $K^{-1}(x_1,x_2)+4 u M \delta(x_1-x_2)/p(x_1)$. To obtain this operator, we first note that $\delta(x_1-x_2)/p(x_1)$ acts as the identity operator ($I$) w.r.t. the data measure since $\int \dif\mu_{x_2} \delta(x_1-x_2)/p(x_1) \phi(x_2)=\phi(x_1)$. Hence, we find that it simply shifts the eigenvalue of $K^{-1}$ (i.e. $\lambda_k^{-1}$) by $4uM$. We thus obtain 
\begin{align}
\label{Eq:KMF_Spectral}
K_{\MF}(x,y) &= \sum_{k=1}^{\infty} \phi_k(x) \phi_k(y) [\lambda_k^{-1}+4uM]^{-1}
\end{align}
alternatively, using the fact that the identity operator commutes with any other operator, or by algebraic manipulations of the above formula, we equivalently find 
\begin{align}
K_{\MF} &= K [I+4 uM K]^{-1} = [I+4 uM K]^{-1} K
\end{align}
Next using square completion on $S_{\MF}$ we further find 
\begin{align}
\langle f(x) \rangle_{S_{\MF}} &= \int \dif\mu_y \dif\mu_z K_{\MF}(x,y) K^{-1}(y,z) \bar{f}(z) 
&= \int \dif\mu_z [K_{\MF}K^{-1}](x,z) \bar{f}(z) \\ \nonumber 
&= \int \dif\mu_z [I+4uMK]^{-1}(x,z) \bar{f}(z)
\end{align}
one can verify that these two results reproduce averages under $S_0$ for $u=0$. 

Having obtained an approximate Gaussian theory for $f(x)$ and fleshed out its correlation functions and averages, what remains is to determine the parameter $M$. Plugging the last several results into Eq. (\ref{Eq:MF_1}) we obtain the following {\it self-consistency} equation for $M$ 
\begin{align}
M &= \int \dif\mu_{x} K_{\MF}(x,x)+\left(\int \dif\mu_{z} [I+4uMK]^{-1}(x,z) \bar{f}(z)\right)^2
\end{align}
To simplify matters, let us next assume $\bar{f}(z)$ is some eigenfunction ($\phi_q(z)$) of $K$. Turning to the spectral representation of $K_{\MF}$ (Eq. \ref{Eq:KMF_Spectral}) and using the fact that now $\int \dif\mu_{z} [I+4uMK]^{-1}(x,z) \bar{f}(z)=\phi_q(z) (1+4uM \lambda_q)^{-1}$ we obtain an  algebraic equation for $M$  
\begin{align}
M &= \sum_{k=1}^{\infty}[ \lambda_k^{-1} + 4u M]^{-1}+ [1+4uM \lambda_q]^{-2}
\end{align}
which can be easily solved on a computer. Seeking to do pure analytics, we note that typically the largest $\lambda_k$ ($\lambda_{k=1}$) is order $1$ ($O(1)$) and also $\sum_{k=1}^{\infty} \lambda_k=\int \dif\mu_x K(x,x)\equiv \Tr[K]$. Thus for $u \ll 1$, we may Taylor expand the above equation to leading order in $M$ which gives 
\begin{align}
M &= \Tr[K]+1 - 4uM \sum_k \lambda_k^2 -8 u M \lambda_q \Rightarrow M &= \frac{\Tr[K]+1}{1+4u(\sum_k \lambda_k^2+2\lambda_q)}
\end{align}
which, as aforementioned, holds for sufficiently small $u$. 

We briefly comment that, in general, when solving self-consistency equations, one may find several plausible (i.g. non-imaginary, non-exploding, having the allowed sign) solutions for $M$. One should then compare the free energy of these solutions ($\log(\int Df e^{-S_{\MF}[f]})$ \footnote{Here one should be careful to include all the constant contributions to the action, such $uM^2$, we so far discarded}) and choose the one with the higher free energy. If one obtains that just near $u=u_c$ several solutions co-exist with the same free energy-- this marks a {\it phase transition} (provided that our mean-field treatment is accurate namely, that $\Delta^2[f]$ is indeed negligible). If these different solutions for $M$ coincide at $u=u_c$ we call it a {\it second-order phase transition} or a {\it continuous phase transition}. If the solutions are distinct at $u=u_c$, we call it a {\it first-order phase transition}.  

We also note that the mean-field can be improved using perturbation theory in the $\Delta^2[f]$ term we neglected taking $S_0$, in the perturbation theory context, to be our $S_{\MF}$. Showing that such corrections are small, is also a way of testing the accuracy of the mean-field treatment or assessing which scales control it. 

\subsection{Saddle point}
Another useful approximation which contains mean-field as a particular case is the {\it saddle point approximation}. It applies in cases where a large parameter appears in the action and controls its scale. For instance 
\begin{align}
Z &= \int Df e^{-P \tilde{S}[f]}
\end{align}
where $\tilde{S}[f]$ is some potentially non-linear action. Let us assume that the global minima of $\tilde{S}[f]$ as a function of $f$ is unique and obtained by some real-valued function $f_0(x)$. The saddle-point approximation then amounts to expanding the action to quadratic order around $f_0(x)$ namely  
\begin{align}
Z &= e^{-P\tilde{S}[f_0]} \int Df e^{-P \int \dif x \dif  y \; \left(\delta_{f(x)} \delta_{f(y)} \tilde{S}[f_0]\right) (f(x)-f_0(x))(f(y)-f_0(y))}
\end{align}
the exactness of this approximation at large $P$ stems from (I) our assumption that $\tilde{S}[f]$ is some smooth functional which is independent of $P$ and hence analytic even as $P\rightarrow 0$. (II) As we found a global minimum, $P\delta_{f(x)} \delta_{f(y)} \tilde{S}[f_0]$ is a positive definite operator, with eigenvalues scaling as $O(P)$ hence the contributions of $f(x)$ such that $f(x)-f_0(x) \gg 1/\sqrt{P}$ to the path integral are exponentially suppressed making the Taylor expansion exact at $P \rightarrow \infty$. 

In more elaborate scenarios, the minimum value of $S[f]$ may come out imaginary. Typically, the above formula still holds as is however, verifying it becomes more subtle. In particular, it requires analytically continuing the path integral to imaginary functions (imaginary $f_k$'s) and making sure that an integration contour exists that goes through the minima along its convex direction and obeys the path-integral boundary conditions (negative real infinity to positive real infinity). In general, this is quite hard to be certain of, however, if this happens to hold, the above approximation is again exact. Thus one can view the above as generating a concrete hypothesis that should be tested experimentally. 

{\bf Example.} Let us revisit the previous mean-field setup but handle it through a saddle point. To this end, one first notes that the action has no such explicit $P$ factor outside. However, we flesh out this factor using the following Fourier identity trick, namely 
\begin{align}
Z &= \int_{-\infty}^{\infty} dM \int_{-\infty}^{\infty} d\tilde{M}\int Df e^{i \tilde{M}(M-\int \dif\mu_x f^2(x))} e^{-S[f_0]-uM^2} 
\end{align}
where we ignored constant factors that are independent of the integration variables. Next, we note that the $f$ appears quadratically in the action and hence we may integrate over it yielding 
\begin{align}
Z &= \int_{-\infty}^{\infty} dM \int_{-\infty}^{\infty} d\tilde{M} e^{i \tilde{M} M-uM^2 - S(\tilde{M})} \\ \nonumber 
S(\tilde{M}) &= -\log \left( \int Df e^{-S_0[f]-i\tilde{M} \int \dif\mu_x f^2(x)}\right)
\end{align}
Using again square completion identities on $S_0[f]$ to extract the $\bar{f}(x)$ related contribution as well as using $\int Df e^{-\frac{1}{2} \int \dif\mu_x \int \dif\mu_y f(x) K^{-1}(x,y) f(y)}=\Pi_{k} (2\pi \lambda_k)^{1/2}$ and using the identity $\log(\text{det}(A))=\Tr \log(A)$ we find 
\begin{align}
S(\tilde{M}) &= \frac{1}{2}\Tr\log(K^{-1}+2i \tilde{M} I)-\frac{1}{2}
\int \dif\mu_x 
\int \dif\mu_y \bar{f}(x) [K^{-1}+2i\tilde{M}I]^{-1}(x,y) \bar{f}(y)  \\ \nonumber 
&= \frac{1}{2}\sum_k \log(\lambda_k^{-1}+2i\tilde{M}) \\ \nonumber 
&-\frac{1}{2}\int \int[K^{-1}\bar{f}](x) [K^{-1}+2i\tilde{M}I]^{-1}(x,y) [K^{-1}\bar{f}](y)
\end{align}
up to a constant factor. Next requiring that the $M$ integration is extremized we obtain $i \tilde{M}=2uM$, similarly from the requirement for the $\tilde{M}$ integration to be at its saddle we obtain
\begin{align}
&\partial_{i\tilde{M}
}\left(i \tilde{M}M - S(\tilde{M}) \right) = 0 \\ \nonumber 
&\Rightarrow  M =  \sum_k (\lambda_k^{-1}+2i \tilde{M})^{-1}+\int \dif\mu_x \dif\mu_y [K^{-1}\bar{f}](x)[K^{-1}+2i\tilde{M}I]^{-2} [K^{-1}\bar{f}](y) \\ \nonumber 
&\stackrel{\text{saddle}}{=} \sum_k (\lambda_k^{-1}+4uM)^{-1}+\int \dif\mu_x \dif\mu_y \bar{f}(x)[I+4uM K]^{-2} \bar{f}(y)\end{align}
where we recognize our previous mean-field equations for $M$. We here used the identity $\partial_ {A_{ij}} \log \det(A) = [A^{-1}]_{ij}$.

Let us re-examine some of the subtleties we glanced over in the above computation. First is the lack of a clear large parameter controlling the scale of the action. To flesh it out, let us consider a case where the spectrum of $K(x,y)$ consists of $d$ degenerate modes with eigenvalue $1/d$ \footnote{This degeneracy and scaling would be the case for a linear fully connected with a uniform data measure on the $d$-dim hypersphere}. In which case $\sum_k \log(\lambda_k^{-1} + i 2\tilde{M})=d \log(1 + i 2\tilde{M}/d)+d\log(d)$ which at large $d$ (i.e. many contributing modes) contains a large pre-factor (and even becomes more and more smooth). Turning to $M$, we note that the action is already quadratic and hence Taylor expanding it in $M$ to second order is exact and not part of the saddle point approximation. 

The second issue is that the saddle point for $\tilde{M}$ is imaginary, while the integration contour is real. As aforementioned, this requires analytically continuing the $\tilde{M}$ integration to the complex plane, such that it goes through the saddle along the concave direction. This can be done and a more detailed treatment, using large deviation theory and in particular in this case, the G\"{a}rtner Ellis theorem [\cite{touchette2009large}]. 
 
\section{The Replica Method}
\label{Sec:Replicas}
Often the action or equivalently the partition function depends on some unknown random parameters.  In machine learning, these could be the model parameters and/or the specific choice of dataset. In such settings, we wish to compute averages/observables given a particular choice of random parameters but then average these over the distribution of parameters (also known as {\it quenched average}). We thus have two ensembles in mind, the first concerns the actual degrees of freedom of the system and the second is the ensemble of model parameters. The Replica Method [e.g.  \cite{MezardBook}] is a technique which helps us treat these two ensembles on equal footing, effectively treating parameters as additional degrees of freedom. This then allows us to use standard techniques from field theory to treat the effects of disorder/parameter-fluctuations.

To illustrate the method, let us take the simplest system possible with one degree of freedom ($f$) and one parameter ($m$) which we wish to average observables over. Specifically, consider a simple non-centred Gaussian in the scalar variable $f$ whose action is given by 
\begin{align}
S(f;m) &= \frac{1}{2} f^2 - m f
\end{align}
The parameter $m$ is not known; rather, its distribution is known to be $N[0,\sigma^2;m]$. Consider some observable, say the average $f$, we may then ask what is the expectation value of this second moment under the above distribution for $m$ namely 
\begin{align}
\int dm {\cal N}(\bar{m},\sigma^2;m) \langle f \rangle_{S(f;m)}
\end{align}
This type of average, an average over action parameters taken after computing expectation value, is known in physics as a {\it quenched average}. More generally, $f$ would represent the configuration space of a statistical mechanics model and $m$ some set of random parameters in the action.  

In our simple model, the above-quenched average can be computed straightforwardly. Specifically since $\langle f \rangle_{S(f;m)}=m$ we find 
\begin{align}
\int dm {\cal N}(\bar{m},\sigma^2;m) m = \bar{m}
\end{align}

Crucially, however, we typically do not know how to solve the action for a given $m$. Indeed, our ability to perform computations often hinges on symmetry and random parameters generally spoil these helpful symmetries. Consequently, it is useful to find a formulation where fluctuations in parameter and fields/integration-variables are treated on more equal footing.  

To set the stage for the replica method, let us first express the desired expectation value via the partition function for a given $m$, using a source field $\alpha$,
\begin{align}
Z(\alpha;m) &= \int \dif f e^{-\frac{1}{2}f^2 + mf + \alpha f} \\ \nonumber 
\bar{f}(m) &= \partial_{\alpha}\log(Z(\alpha;m))|_{\alpha=0}
\end{align}
We note that by taking higher derivatives w.r.t. to $\alpha$, we can obtain all the statistics of $f$. Consequently, our main focus turns to the free energy, since its dependence on $\alpha$ encodes the entire statistics. Moreover, since averaging w.r.t. $m$ and taking derivatives w.r.t. $\alpha$ commute, we can obtain the quenched average of all cumulants of $f$ from the quenched average of the free energy namely $\int dm {\cal N}(\bar{m},\sigma^2;m) \log(Z(\alpha;m))$. 

To compute the latter average, we note the identity  $\log(x)=\lim_{Q \rightarrow 0}(x^Q-1)/Q$. We may thus write 
\begin{align}
\int dm {\cal N}(\bar{m},\sigma^2;m)\log(Z(\alpha;m)) &\equiv \langle\log(Z(\alpha;m))\rangle_{m} \\ \nonumber 
&= \lim_{Q \rightarrow 0} \left \langle \frac{Z(\alpha;m)^Q - 1}{Q}
\right \rangle_{m} \end{align}
where, for quenched averages, we denote the variable being averaged by the subscript of the angular brackets. 

Next, we shall compute the r.h.s. for all positive integer $Q$ and analytically continue the result to the positive reals so that we can take the limit. The uniqueness of such an analytical continuation, is unfortunately unclear. For instance, for any function we find one may add the function $\sin(\pi Q)$ as it is zero on integer $Q$'s. Fortunately, the simplest analytical continuation, which is to treat $Q$ as a real variable, is typically sufficient. 

Let us perform this procedure on the above example. Splitting $m$ into its mean part ($\bar{m}$) and fluctuation part ($m-\bar{m}$) we obtain  
\begin{align}
\notag \left\langle Z(\alpha;m)^Q 
\right\rangle_{m} &= \frac{1}{\sqrt{2\pi \sigma^2}}\int dm \int \dif f_1..\dif f_Q \; e^{-\frac{(m-\bar{m})^2}{2\sigma^2}-\sum_{a=1}^Q \left[\frac{f_a^2}{2}-[(m-\bar{m})+(\alpha+\bar{m})] f_a\right]}\\  &= ...
\end{align}
where each of the resulting $Q$ copies of $f$ is referred to as a replica of $f$. As advertised, for a given positive integer $Q$, this can be viewed as the partition function of $Q+1$ degrees of freedom, one of them being the action parameter. 

In general the replicas are decoupled for given parameter $m$, however integrating over $m$ generates a coupling between them namely
\begin{align}
... &= \int \dif f_1..\dif f_Q e^{-\sum_{a=1}^Q \left[\frac{f_a^2}{2}-(\alpha+\bar{m}) f_a\right]+\sigma^2 \sum_{ab=1}^Q \frac{f_a f_b}{2}} = ...
\end{align}
Notably the quadratic terms in the action can be written as $\frac{1}{2}\bm{f}^T [I-\sigma^2\Gamma] \bm{f}$, where $\bm{f}$ is a vector in replica space and $\Gamma$ is the all-ones matrix ($\Gamma_{ab}=1$). The $\Gamma$ matrix is rank one, with a single non-zero eigenvalue $Q$. Thus, the above Gaussian integral is non-convergent when $\sigma^2 \geq 1/Q$. Since we wish to obtain the behavior as $Q \rightarrow 0$, we do not concern ourselves with this issue and proceed with calculating it assuming $Q$ or $\sigma^2$ are small enough. This yields 
\begin{align}
... &= (\sqrt{2\pi})^Q e^{-\frac{1}{2}\Tr\log([I-\sigma^2 \Gamma])+\frac{(\alpha+\bar{m})^2}{2} \bm{1}^T [I + \sigma^2\Gamma]^{-1} \bm{1}} \\ \nonumber 
&= (\sqrt{2\pi})^Q e^{-\frac{1}{2}\log(1-Q\sigma^2)+\frac{(\alpha+\bar{m})^2}{2} \frac{Q}{1+Q\sigma^2}}\end{align}
where $\bm{1}$ is the all-ones vector in replica space. 

Next, we perform analytical continuation simply by treating $Q$ as a real variable, taking $Q
\rightarrow 0$, and using a Taylor expansion to yield 
\begin{align}
\lim_{Q \rightarrow 0} \left \langle \frac{Z(\alpha;m)^Q - 1}{Q}
\right \rangle_{m}  &= \lim_{Q \rightarrow 0}  \frac{e^{\frac{1}{2} Q\sigma^2+\frac{(\alpha+\bar{m})^2}{2}Q}-1}{Q} = \frac{1}{2}[\sigma^2+(\alpha+\bar{m})^2]
\end{align}
Next, we recall that the r.h.s. is just the quenched averaged free energy, hence taking its derivatives w.r.t. $\alpha$ should recover that quenched averaged cumulants. Indeed taking one derivative yields $\bar{f}=\bar{m}$, twice yields $1$ (the variance of $f$ which is unaffected by $m$). Further derivatives yield zero since, regardless of $m$, the distribution is Gaussian and its higher cumulants vanish. 

%% file: Chapters/Infinite_networks_in_standard_scaling.tex
\chapter{Infinite networks in standard scaling}

One of the surprising empirical findings about neural networks is the fact that increasing the number of DNN parameters has, typically, a beneficial effect on performance. Historically, this was noted as soon as 1995 \citep{Breiman2018ReflectionsAR}, but made more notable following the recent rise of DNNs by \cite{Neyshabur2014,Zhang2016}. Namely, despite having orders of magnitude more parameters than training points to fit, DNNs generalize (fit/predict previously unseen data points) quite well. This stands in contrast to the intuition that such DNN would find one random weight configuration out of many to fit the training data and, as each such configuration is expected to give a different predictor, what would be the chance that this random choice ended up the right one? 

This discussion motivates us to consider infinitely over-parametrized DNNs, such as those discussed in Sec. \ref{Sec:InfiniteRandomDNNs}, as these embody the above issue most clearly. In addition, one may hope this limit would also describe a well-performing regime of DNNs. However, as we shall see, this description often overshoots the best-performing regime which occurs at finite overparametrization (see 
 \cite{novak2018bayesian} for an empirical comparison)\footnote{Or at infinite overparametrization in so-called mean-field scaling \cite{yang2022tensorprogramsvtuning,MeiMeanField2018}}. Nevertheless, the description obtained in this limit has three advantages: {\bf (i)} It is simple and insightful, {\bf (ii)} It correlates with actual network performance \citep{novak2018bayesian,lee2019wide} thereby allowing us to rationalize architecture choices, {\bf (iii)} It serves as an analytical starting point for studying more advanced, finitely over-parametrized regimes.

\section{The DNN-NNGP mapping for trained networks}
\label{Sec:NNGP_On_Data}
Consider a DNN trained for infinitely long using the Langevin dynamics discussed in Sec. \ref{Sec:TrainingProtocols}. Let us first focus on DNN outputs ($z(x)$) on $P$ training points ($x_1...x_P$) and one test point ($x_0$). Furthermore, for simplicity, we focus on scalar outputs. Consider the partition function associated with the equilibrium distribution of the weights found in Eq. (\ref{Eq:BoltzmannDistWeights}) (i.e. $P(\theta) \propto e^{-L(\theta)/T}$). As common in physics, instead of working with the probability (which implies carrying normalization factors and the dummy variable $\theta$), we work with its normalization factor also known as the partition function 
\begin{align}
Z &= \int d\theta e^{-L(\theta)/T}
\end{align}
where we use the following  unnormalized-MSE+weight-decay loss 
\begin{align}
L(\theta) &= \frac{1}{2}\sum_{\mu=1}^P (z_{\theta}(x_{\mu})-y(x_{\mu}))^2 + T \sum_{\alpha} \frac{\theta^2_{\alpha}}{2\sigma^2_{\alpha}}
\end{align}
and allowed different weight decay factors for each weight, to be determined soon. We often denote 
\begin{align}
\kappa^2 &= T
\end{align}
where $\kappa$ is referred to as the ridge parameter. 

We comment that the above distribution over network parameters has the following Bayesian interpretation. Consider the statistical model  $P((x,y)|\theta) \propto e^{-(z_{\theta}(x)-y)^2/(2\kappa^2)}$, with $\theta$ being the model parameters. This model can be interpreted as a neural network with an output corrupted by Gaussian noise of std $\kappa$. The distribution in Eq. \ref{Eq:BoltzmannDistWeights} coincides with the Bayesian posterior of the above model given a Gaussian iid prior given by $\theta_{\alpha} \sim {\cal N}(0,\sigma^2_{\alpha})$. This is a particular link between the physical notion of equilibrium and Bayesian inference. 

Next, we wish to simplify the above distribution. To this end, we note that the complex dependence on $\theta$ enters only through the network outputs one may write 
\begin{align}
\label{Eq:Derivation1}
Z &= \int df_0..df_P \int d\theta \Pi_{\mu=0}^P \; \delta(f_{\mu}-z_{\theta}(x_{\mu}))  e^{-L(\theta)/T} \\ \nonumber 
&= \int df_0..df_P \; e^{-\frac{1}{2T} \sum_{\mu=1}^P (f_{\mu}-y(x_{\mu}))^2}\int d\theta \; e^{-\sum_{\alpha} \frac{\theta^2_{\alpha}}{2\sigma^2_{\alpha}}} \Pi_{\mu=0}^P \; \delta(f_{\mu}-z_{\theta}(x_{\mu}))  
\end{align}
where we denote by $f_\mu = f(x_\mu)$. Viewing the last term as a function of $f_0..f_P$, one finds that apart from an inconsequential normalization factor, it is really the definition of the probability density induced on $f_0..f_P$ by a DNN with random Gaussian weights. Denoting this probability by $P_0(f_0..f_P)$ we thus find 
\begin{align}
Z &= \int df_0..df_P e^{-\frac{1}{2T} \sum_{\mu=1}^P (f_{\mu}-y(x_{\mu}))^2} P_0(f_0..f_P)
\end{align}
we comment that, reading off a probability density for $f_0..f_P$ by removing the integrals and normalizing by $Z$, one obtains the formula for the Bayesian posterior with additive Gaussian measurement noise and $P_0$ as a prior in function space  \citep{Welling2011,cohen2021learning}. Thus, our results carry through to Bayesian Inference and Bayesian neural networks.

So far our manipulations have been exact. Taking the limit of infinite overparametrization, as discussed in Sec. \ref{Sec:InfiniteRandomDNNs}, we found that the outputs of the DNN are Gaussian. Hence, in this limit $P_0$ simplifies to a centered Gaussian distribution with covariance matrix $K(x_{\mu},x_{\nu})$. Denote this $(P+1) \times (P+1)$ matrix simply by $K$, we find the following simple result in the standard-over-parametrized (SOP) limit 
\begin{align}\label{eq:S_SOP}
Z_{\text{SOP}} &= \int df_0..df_P \; e^{-\frac{1}{2T} \sum_{\mu=1}^P (f_{\mu}-y(x_{\mu}))^2} e^{-\frac{1}{2} \sum_{\mu \nu=0}^P f_{\mu} [K^{-1}]_{\mu \nu} f_{\nu}}
\end{align}
Namely a Gaussian partition function. 

Calculating observables under this partition function, using block matrix inversion lemmas, yields the well-known Gaussian Process Regression (GPR) formulas \citep{Rasmussen2005},  namely 
\begin{eqnarray}
\label{Eq:GPR} \\ \nonumber
&\langle f(x_0) \rangle_{Z_{\text{SOP}}} &= \sum_{\mu \nu=1}^P K(x_0,x_{\nu}) [(K_D+TI)^{-1}]_{\nu \mu} y(x_{\mu}) \\ \nonumber 
&\langle f^2(x_0) \rangle_{Z_{\text{SOP}}} - \langle f(x_0) \rangle^2_{Z_{\text{SOP}}} &= K(x_0,x_0) \\ \nonumber
& & - \sum_{\mu \nu=1}^P K(x_0,x_{\nu}) [(K_D+T I)^{-1}]_{\nu \mu} K(x_{\mu},x_0) 
\end{eqnarray}
where $K_D$ is the $P \times P$ matrix with elements $[K_D]_{\mu>0,\nu>0}=K(x_{\mu},x_{\nu})$. 

Despite the difficulty of inverting $K_D$, which may be extremely large in practice, the above formula provides concrete predictions for the average output of a trained DNN and its fluctuation under the ensemble of DNN induced by the Langevin dynamics. Experimentally, the average value can be obtained by training many networks using different realizations of the gradient noise and initial conditions, and averaging the output over this set of DNNs. Such a technique (ensembling) is used in practice for small datasets when training is not too costly. 

The above result also sheds light on why over-parametrization is, to the very least, not strictly detrimental to performance. Indeed in this limit, we have obtained GPR--- a known and respectful way of doing inference. One can still argue that the strong weight decay we required, coming from the fact that  $\sigma^2_{\alpha}$ scales one over fan-in, is sufficiently restrictive to counter the addition of parameters. Here one should note that having weight decay such that weights are limited to a scale of $1/\sqrt{N}$ is not very restrictive on the size of the resulting functions. Indeed if such weights are summed coherently, they can still reach a diverging magnitude of $\sqrt{N}$. However, to counter this most sharply, together with potential concerns about using fully equilibrated DNNs, we next discuss a way of obtaining a similar GPR result without using weight decay or assuming equilibrium, via the celebrated Neural Tangent Kernel result \citep{jacot2018neural}. 

\section{The DNN-NTK mapping for trained networks}
Consider next a DNN trained with Gradient Flow and unnormalized MSE loss and no weight decay. Following Refs. \cite{Jacot2018,lee2019wide} let us study the dynamics on the outputs $f(x_0)..f(x_P)$ induced by the dynamics on the weights via 
\begin{align}
\frac{\dif f_t(x_{\mu})}{\dif t} &= \sum_{\alpha} \partial_{\theta_{\alpha,t}}f(x_{\mu}) \frac{\dif \theta_{\alpha,t}}{\dif t} \\ \nonumber 
\frac{\dif \theta_{\alpha,t}}{\dif t} &= - \gamma \partial_{\theta_{\alpha}} L = -  \gamma  \sum_{\nu=1}^P [f_t(x_{\nu})-y(x_{\nu})] \partial_{\theta_{\alpha}} f_t(x_{\nu})
\end{align}
where we have kept the dependence of $f_t(x_{\nu})$ on the weights implicit. Note that the second equation defines Gradient Flow with the said loss. Joining these two equations and defining the time-dependent Neural Tangent Kernel (NTK) $\Theta_t(x_\mu, x_\nu)$, we obtain 
\begin{align}\label{Eq:NTK}
\frac{\dif f_t(x_{\mu})}{\dif t} &= -  \gamma \sum_{\nu=1}^P \Theta_t(x_{\mu},x_{\nu}) [f_t(x_{\nu})-y(x_{\nu})]  \\ \nonumber 
\Theta_t(x_{\mu},x_{\nu}) &= \sum_{\alpha} \partial_{\theta_{\alpha}}f_t(x_{\mu}) \partial_{\theta_{\alpha}}f_t(x_{\nu})
\end{align}

So far our manipulations have been exact, the important insight is that $\Theta_t(x_{\mu},x_{\nu})$, the NTK, becomes independent of time and independent of initialization at infinite over-parametrization. This surprising result is related to the aforementioned possibility of generating diverging functions using coherent sums of weights. More accurately, minor but coherent $1/N$ (i.e. one over width) changes to the $1/\sqrt{N}$-sized random weights are sufficient to generate $O(1)$ functions. Given such vanishingly small changes throughout training, it is reasonable to expect that a leading order Taylor expansion of $f(x_{\mu};\theta_t)$ around the initial $\theta_0$ value is enough \citep{lee2019wide}. If so we have that throughout training  $f_{t}(x_{\mu})\approx f_{0} (x_{\mu})+\sum_{\alpha}(\theta_{t,\alpha}-\theta_{0,\alpha})\partial_{\theta_{\alpha}}f_{0}(x_{\mu})$. Plugging this into the above definition of $\Theta_t(x_{\mu},x_{\nu})$ and noting that all $\theta_{\alpha}$ dependence of $f_t(x_{\mu})$ now comes from the $\theta_{\alpha,t}$ factor, yields a time-independent NTK evaluated at the initial weights which we denote by $\Theta(x_{\mu},x_{\nu})$. Having $N \rightarrow \infty$ equivalent parameters being summed over, it is also natural to expect this quantity to concentrate/self-average and become independent of the particular draw of initial weights. These are all, of course, heuristic arguments and proofs can be found in Ref. \cite{jacot2018neural}. 

Given fixed $\Theta(x_{\mu},x_{\nu})$, the dynamics is now governed by a linear ODE in $P+1$ variables whose solution is  
\begin{align}
f_{t}(x_{\mu>0}) &= \sum_{\nu=1}^P [e^{-\gamma \Theta_D t}]_{\mu \nu}
 [y(x_{\nu})-f_{0}(x_{\nu})]
\end{align}
where $\Theta_D$ is the matrix obtained by taking the NTK only between training points. Turning to the test point ($x_0$) we can place the above solution within the differential equation to obtain
\begin{align}
\partial_t f_{t}(x_0) &= -\gamma \sum_{\mu=1}^P \Theta(x_0,x_{\mu}) \left[\sum_{\nu=1}^P [e^{-\gamma \Theta_D t}]_{\mu \nu}
 [y(x_{\nu})-f_{0}(x_{\nu})]-y(x_{\mu})\right] 
\end{align}
Integrating both sides from $t=0$ to $\infty$ one obtains the outputs after training infinitely long on a test point $x_0$, namely
\begin{align}
\label{Eq:NTKInfiniteTime}
f_{\infty}(x_0) &= \sum_{\mu \nu=1}^P \Theta(x_0,x_{\mu}) [\Theta_D^{-1}]_{\mu \nu} y(x_{\nu}) + I_0 \\ \nonumber
I_0 &= f_0(0)-\sum_{\mu \nu=1}^P \Theta(x_0,x_{\mu}) [\Theta_D^{-1}]_{\mu \nu} f_{0}(x_{\nu})
\end{align}
Ignoring for the moment $I_0$, the first line appears as GPR with zero ridge ($\kappa^2=0$, ridgeless) using however the NTK kernel ($\Theta$) rather than $K$, the NNGP kernel. The second line then appears as the discrepancy in using such ridgeless GPR in predicting the test point's initial value given the train points' initial values. Notably averaging over an ensemble of initialization seeds, the $I_0$ contribution vanishes and one is left with the first line alone. 

Thus, the NTK approach leads to GPR formula for the predictor/average. We thus obtain a similar result to the DNN-NNGP correspondence described in the previous chapter. Nonetheless, the variance contribution (related to moments of $I_0$) differs from that in the previous section. We note by passing that GPR with the NTK kernel is very well suited to predict functions generated at random by the DNN with that NTK. Hence, for complex target functions requiring large $P$, we generally expect the $I_0$ contribution to be quite small.  

\section{Field theory of Gaussian Process Regression (GPR) - on data}
We next wish to rephrase some of the above GPR results as a field theory. Even putting aside the powerful machinery offered by field theory, a more concrete motivation here is to avoid the spurious dependence on data of the functional prior term in the action in Eq. \eqref{eq:S_SOP}. Such a dependence obfuscates the natural symmetries of the action and makes the process we would soon undertake -- averaging over datasets, cumbersome. 

Accordingly, we wish to write a field theory which would reproduce the statistics for $f(x_0)..f(x_P)$ but allow us to extend it to any sample ($x$). This is important for understanding generalization. To achieve this, we follow Sec. \ref{Sec:PathIntegrals} and take a limit where we gradually extend the set of points $x_0..x_P$ to a much denser set $x_1..x_n$ but make sure $x_1..x_P$ are always a subset of this denser set so that we can still write the loss term. Finally, we take $n\rightarrow \infty$ and replace sums with integrals, matrices with operators, and vectors with functions while making sure we track $n$ factors correctly.  The result which we would soon verify independently is 
\begin{align}
\label{Eq:Z_FieldTheoryOnData}
Z &= \int Df e^{-S_{\GP}} \\ \nonumber
S_{\GP} &= \frac{1}{2} \int d\mu_x d\mu_y f(x) K^{-1}(x,y) f(y) + \frac{1}{2\kappa^2}\sum_{\mu=1}^P (f(x_{\mu})-y(x_{\mu}))^2
\end{align}
where $K^{-1}(x,y)$ is the inverse operator to $K(x,y)$ under the $d\mu_x$ measure as defined in Sec. \ref{Sec:Intro_Measure}.

Let us demonstrate how we reproduce the original discrete posterior  ($Z_{\textrm{SOP}}$ of Eq. \eqref{eq:S_SOP}) from this one. To this end, we introduce the variables $g_0..g_P$, force them to equal to $f(x_0)..f(x_P)$ using delta functions, and integrate out $f$ to obtain the partition function for such $g_0..g_P$. The latter enables us to calculate any statistical property $f(x_0)..f(x_P)$ and would match, up to inconsequential factors, the original discrete $Z$. Specifically, 
\begin{align}
Z &= \int dg_0..dg_P \int Df  e^{-S_{\GP}} \prod_{\nu=0}^P \delta(g_{\nu}-f(x_{\nu})) \\ \nonumber 
&= \int dt_0..dt_P \; dg_0..dg_P \int Df \exp\biggl\{-S_{\GP}+i\sum_{\nu=0}^P t_{\nu}(g_{\nu}-f(x_{\nu})) \biggl\} \\ \nonumber
&= \notag \int (...) \int Df \exp\biggl\{-\frac{1}{2} \int d\mu_x d\mu_y f(x) K^{-1}(x,y) f(y) \\&- \frac{1}{2\kappa^2}\sum_{\mu=1}^P (g_{\mu}-y(x_{\mu}))^2+i\sum_{\nu=0}^P t_{\nu}(g_{\nu}-f(x_{\nu})) \biggl\}
\end{align}
where the ellipsis $(...)$ means copying the appropriate expression from the previous line which here is $dt_0..dt_P \; dg_0..dg_P$. We also rewrote the delta function as an exponential. To obtain our original discrete $Z$, we next integrate over $Df$, using a square completion of the form $f(x) \rightarrow f(x)+i\sum_{\nu} K(x,x_{\nu})t_{\nu}$ which yields (ignoring, as always, constant factors)  
\begin{align}
\notag Z &= \int dt_0..dt_P \; dg_0..dg_P \exp\biggl\{-\frac{1}{2} \sum_{\mu \nu=0}^P t_{\mu} K(x_{\mu},x_{\nu})t_{\nu} \\&- \frac{1}{2\kappa^2}\sum_{\mu=1}^P (g_{\mu}-y(x_{\mu}))^2+i\sum_{\nu=0}^P t_{\nu}g_{\nu}\biggl\}\end{align}
where integrating out the $t$ variables yields $Z$ in the discrete case with $f_{\mu}$ replaced by $g_{\mu}$'s, thereby establishing our claim. For the NTK case, following Eq. \ref{Eq:NTKInfiniteTime}, we simply take $K=\Theta$ and $\kappa^2\to 0$, however obtaining fluctuations ($I_0$) and time dependence requires some more work (see Ch. \ref{Sec:MSRDJ}).

To conclude this section, we discuss some properties of Eq. (\ref{Eq:Z_FieldTheoryOnData}). As we have seen in Sec. \ref{Sec:PathIntegrals}, the first term in $S_{\GP}$ reflects the prior induced on function space by a random DNN. The keen reader may notice an oddity, which is that the prior should be data-measure agnostic. This requirement is indeed obeyed in Eq. \ref{Eq:Z_FieldTheoryOnData}, though implicitly, based on the aforementioned fact that the RKHS norm is measure independent \footnote{Thus measure here plays a somewhat similar role to a Gauge field in physics, at least as far as the prior is concerned, the data-term explicitly breaks this invariance of the theory}. 

Interestingly, the source of the first term is entropic: Indeed it is not difficult to show that, had we replaced the Gaussian iid distribution of each $\theta_{\alpha}$ with a uniform distribution with the same variance, the same kernel would be reproduced at infinite $N$. Considering for simplicity the discrete case, $P_0(f_0..f_P)$ (defined in Eq. (\ref{Eq:Derivation1})) with this Gaussian-to-uniform replacement, one finds that 
\begin{align}
\frac{1}{2} \sum_{\mu \nu}&f(x_{\mu}) [K^{-1}]_{\mu \nu} f(x_{\nu}) = -\log(P_0(f_0..f_P)) + \const \\ \nonumber 
&\stackrel{N \rightarrow \infty}{=} -\log\left(V_{\theta}\int_{\text{intervals}} \dif \theta \prod_{\nu=0}^P \delta(f_{\mu} - z_{\theta}(x_{\nu}))\right) + \const
\end{align}
where $V_{\theta}$ is the total weight space volume encompassed by the uniform measure of the weights. Notably, 
the final line is the entropy/log-volume of weight configurations within intervals set by the uniform measure which yield the outputs $f_0..f_P$. This view carries on the field-theory setting, where it would correspond to the entropy of weights yielding the function $f(x)$. Viewing weights as micro-states and output as the macro-state, the above l.h.s. is proportional to the Boltzmann entropy associated with the macro-state $f(x)$. The second term, in $S_{\GP}$ the data-term, can then be thought of as an energy. Much like in physics, the Gaussian Process (i.e. the partition function $Z$) would seek a $\kappa^2=T$ compromise between entropy and energy, favoring energy at low $T$ and entropy at high $T$. Entropy then acts as a regulating force, pushing the field away from the training data and choosing $f(x)$ which extrapolates between the data points such that it has low entropy. 

This more regular and ``orderly" behavior induced by entropy is analogous to the order-through-disorder effect exhibited by some physical systems \citep{Villain1980}. Note also that this type of entropy is largely a property of the neural network and the relevant volume in weight space and not the training algorithm. Nonetheless, it is likely to affect all training algorithms, as more common-in-weight-space functions are likely to be discovered first by most noisy gradient based optimization methods. 

The fact that entropy acts as a regulator offers a way of rationalizing the original puzzle regarding over-parametrization. Indeed, the absurdity came from the fact that many weight solutions exist, which yield zero train loss. This ambiguity in choosing weights is just what entropy measures. The fact that entropy regulates the solution implies that weight configurations which generate functions with smaller RKHS norms (i.e. are favored by the entropy term) are far more abundant in weight space. Hence during training, dynamics is biased by this abundance to find those smaller-RKHS solutions and goes to high-RKHS/low-entropy solution only given sufficient data. In typical real world setting the eigenvalues of the kernel are very multiscale creating a strong differentiation between functions \footnote{This is not the case for some artificial kernels and datasets, say those with a degenerate spectrum, leading to overfitting issues near the interpolation threshold (e.g. \cite{Canatar2021})}. Given the empirical fact that such GPR works quite well \citep{lee2019wide}, one concludes that this bias toward low RKHS norm, should also be the bias towards more favourable functions, performance-wise. In contrast, highly over-fitting solutions which match the training data but do something erratic outside the training set, have a very low phase space in weight space and hence are much more difficult to find during training. In Sec. \ref{Sec:Symmetries}, we flesh out this bias for the case of an FCN, and show that it favors low degree polynomials. 

\section{Field theory of GPR - data-averaged}
\label{Sec:AveragedGPR}
Looking to make precise analytical statements on the GPR formula, or equivalently on the above field theory, is made difficult by the randomness induced by the specific draw of dataset. This randomness spoils any potential symmetry of the problem and requires us to invert a specific, dataset-dependent, $P\times P$ matrix ($K_D$) to obtain predictions. We thus wish to average over many such different random draws of $P$ points thereby smoothening dataset draw specific effects and restoring potential symmetries.  

Following Sec. \ref{Sec:Replicas}, we focus on the quenched averaged free energy written using the replica identity namely 
\begin{align}
\label{Eq:ReplicatedZGPR}
\langle \log(Z) \rangle_{x_{1..P}} &= \lim_{Q\to 0}\left\langle \frac{Z^Q-1}{Q} \right\rangle_{x_{1..P}} 
\end{align}
where we denote by $\langle A(x_1..x_P) \rangle_{x_{1..P}} \equiv \int d\mu_{x_1}..d\mu_{x_P} A(x_1..x_P)$ for some function of the training points $A(...)$. Focusing on the average of $Z^Q[\alpha]$, one convenient fact is that the prior term in $S_{\GP}$ does not depend on variables being averaged ($x_1..x_P$). This is one of the advantages of working with the $S_{\GP}$ action compared to the discrete one.  Consequently, we find 
\begin{align}
\label{eq:quenched_ZQ}
&\langle Z^Q\rangle_{x_{1..P}} = \int D f_1..Df_Q e^{-\frac{1}{2} \sum_{a=1}^Q \int d\mu_x d\mu_y f_a(x) K^{-1}(x,y) f_a(y)}  \\ \nonumber 
&\times \left\langle e^{-\frac{1}{2\kappa^2}\sum_{a=1}^Q \sum_{\mu=1}^P [f_a(x_{\mu})-y(x_{\mu})]^2}\right\rangle_{x_{1..P}} \\ \nonumber 
&= (...)\times \left(\int d\mu_x e^{-\frac{1}{2\kappa^2}\sum_{a=1}^Q [f_a(x)-y(x)]^2}\right)^P 
\end{align}
at this point, we can already take the log of the data term and get a non-local data term. A slightly more elegant solution which yields a local loss term is to raise the data term to the exponent by noticing the following algebraic identity   
\begin{align}
\label{eq:poisson_averaged_data_term}
&\exp\left(-\tilde{P}+\tilde{P} \int d\mu_x e^{-\frac{1}{2\kappa^2}\sum_{a=1}^Q [f_a(x)-y(x)]^2}\right) \\ \nonumber 
&= e^{-\tilde{P}}\sum_{n=0}^{\infty} \frac{\tilde{P}^n}{n!} \left(\int d\mu_x e^{-\frac{1}{2\kappa^2}\sum_{a=1}^Q [f_a(x)-y(x)]^2}\right)^n
\end{align}
and that the r.h.s. now appears as the data term we need, however, averaged over $P$ (played by $n$ in the above). Thinking about $P$ as a Poisson random variable with mean $\tilde{P}$. At large $\tilde{P}$, fluctuations in the Poisson-averaged data term are of the order $\sqrt{\tilde{P}}$ and so negligible compared to the average. Setting $\tilde{P}=P$ and combining \ref{eq:poisson_averaged_data_term} with \ref{eq:quenched_ZQ}, we obtain, 
\begin{align}
\label{Eq:IntroducingSBar}
&\langle Z^Q[\alpha]\rangle_{x_{1..P}} \underset{P \gg 1}{\approx} e^{-P}\sum_{n=0}^{\infty} \frac{P^n}{n!} \langle Z^Q[\alpha]\rangle_{x_{1..n}} \\ \nonumber 
&e^{-P}\sum_{n=0}^{\infty} \frac{P^n}{n!} \langle Z^Q[\alpha]\rangle_{x_{1..n}} \equiv Z_Q[\alpha] = \int Df_1..Df_Q e^{-\bar{S}_{\GP}[f_1..f_Q]} \\ \nonumber 
&\bar{S}_{\GP} = \frac{1}{2} \sum_{a=1}^Q \int d\mu_x d\mu_y f_a(x) K^{-1}(x,y) f_a(y) - P \int d\mu_x e^{-\frac{1}{2\kappa^2}\sum_{a=1}^Q [f_a(x)-y(x)]^2} \\ \nonumber 
&+ P - \int dx \; \alpha(x) \sum_{a=1}^Q f_a(x)
\end{align}
where we also included a source ($\alpha(x)$) for computing dataset averaged moments of $f(x)$ \footnote{Specifically, we introduce an $\int dx \alpha(x) f(x)$ term in the original, un-replicated action, which then leads to the above term in the replicated action.}.  We thus obtained a field theory of replicas governing the dataset-averaged free energy associated with GPR. This can either be thought of as an approximation to averaging over all datasets with $P$ elements drawn from $p(x)$ or as an exact result associated with averaging over all datasets drawn from $p(x)$ with the number of elements drawn from a Poisson distribution with average $P$ \footnote{We note by passing one can circumvent this Poisson averaging by exponentiating $[\int d\mu_x e^{-\frac{1}{2\kappa^2}\sum_{a=1}^Q [f_a(x)-y(x)]^2}]^P$, the resulting data-term is non-local but still manageable and leads to the same predictions at large $P$}. A similar approach was taken in \cite{malzahn2001variational,cohen2021learning}.  As common when taking quenched averages, the original Gaussian theory became non-Gaussian/interacting following the averaging procedure. To obtain, say, the dataset averaged value of the average GPR predictor at a point $x_0$, namely the dataset average of the first line in Eq. (\ref{Eq:GPR}), one needs to calculate (see Eq. \ref{Eq:ReplicatedZGPR}) 
\begin{align}
&\left\langle \left \langle f(x_0) \right \rangle_{S_{GP},D=[x_1..x_n]}  \right\rangle_{x_1..x_n;n\sim \text{Poisson}(P)} = \lim_{Q\rightarrow 0}\partial_{\alpha(x_0)} \left[Q^{-1}Z_Q[\alpha]-Q^{-1} \right]|_{\alpha=0} \\ \nonumber 
&= \lim_{Q\rightarrow 0} Q^{-1} \partial_{\alpha(x_0)}Z_Q|_{\alpha=0} = ...
\end{align}
Given the relevant setting here, where all replicas of the field would play a similar role (i.e. replica symmetric), the division by $Q$ in the above formula would be cancelled by $\sum_{a=1}^Q$ accompanying $\alpha(x)$. Noting in addition that in the replica limit the action vanishes and hence $Z_{Q\rightarrow 0}=1$ we find 
\begin{align}
... &= \lim_{Q\rightarrow 0} Q^{-1} \partial_{\alpha(x_0)}Z_Q|_{\alpha=0} &= \lim_{Q\rightarrow 0} Q^{-1} Z^{-1}_Q\partial_{\alpha(x_0)}Z_Q|_{\alpha=0} = \lim_{Q\rightarrow 0}\langle f_{a}(x_0) \rangle_{\bar{S}_{GP}},
\end{align}
where $a$ is any specific replica index. 
The above two equations provide a concrete connection between dataset averages of GP predictors and standard averages under $\bar{S}_{GP}$. 

Next, we present several approximation techniques for evaluating the average of $f_a(x)$ under the above field theory. 
\subsection{Equivalent Kernel}
\label{Sec:EK}
The simplest approximation method, yielding the so-called Equivalent Kernel (EK) result \citep{Silverman1984}, is to assume $(f_a(x)-y(x))^2$ is much smaller than $\kappa^2$ on the measure $d\mu_x$. Roughly speaking, this should be correct when the typical MSE loss is smaller than $\kappa^2$ as expected to happen at large enough $P$ and fixed $\kappa^2$. More formally, we can think of EK as the asymptotic limit of taking $P\rightarrow \infty$ with $P/\kappa^2$ kept fixed (e.g. more data points but also more noise to balance the amount of information). Concretely, we approximate 
\begin{align}
\label{Eq:Approx_S_GP}
\bar{S}_{\GP} &\approx \bar{S}_{\GP,\text{EK}} = \frac{1}{2} \sum_{a=1}^Q \int d\mu_x d\mu_y f_a(x) K^{-1}(x,y) f_a(y) \\ \nonumber  &- P \int d\mu_x \left(1- \frac{1}{2\kappa^2}\sum_{a=1}^Q [f_a(x)-y(x)]^2 + O(\kappa^{-4})\right)  
+ P  \\ \nonumber 
&= \frac{1}{2} \sum_{a=1}^Q \int d\mu_x d\mu_y f_a(x) K^{-1}(x,y) f_a(y) +\frac{P}{2\kappa^2} \int d\mu_x [f_a(x)-y(x)]^2  \\ \nonumber 
\end{align}
where we dropped the $\alpha(x)$ source term. 
Notably, the above action is Gaussian and does not contain any coupling between replicas. Following this, we can focus on a single replica and move to the eigenfunction basis of $K(x,y)$ ($\phi_k(x)$) to obtain the following simple ``diagonal'' action per replica
\begin{align}
\label{Eq:EK_Av_Pred}
\bar{S}_{\text{GP,EK}}&= \sum_{k=1}^{\infty} \frac{f_k^2}{2 \lambda_k} + \frac{(f_k-y_k)^2}{2\kappa^2}
\end{align}
where $f_k = \int d\mu_x \phi_k(x) f(x)$ (and similarly with $y_k$). This then yields the following approximation for the dataset averaged GPR predictor 
\begin{align}
\langle f_a(x) \rangle_{\bar{S}_{\GP}} &= \sum_k \frac{\lambda_k}{\lambda_k + \frac{\kappa^2}{P}} y_k \phi_k(x) + O(\frac{Loss^2}{\kappa^4})
\end{align}
whereby $Loss^2$ we mean more formally the typical value of $(f_a(x)-y(x))^2$ which is of the order of the MSE loss unless some outliers appear. We thus estimate that when the mean loss ($Loss$) is much smaller than the ridge parameter, the GPR predictor behaves as a linear high-pass spectral filter on the target. Namely, eigenfunctions with $\lambda_k P \gg \kappa^2$, can be copied into the predictor and eigenfunctions with $\lambda_k P \ll \kappa^2$ get discarded. 

Turning to variances, there are three different types here: {\bf (i)} Taking the variance per dataset according to GPR (i.e. the second line of \ref{Eq:GPR} giving us fluctuations induced by the gradient noise and, more importantly at low $\kappa$, by the remaining randomness in the weights which minimize the loss) and averaging it over datasets {\bf (ii)} Taking the GPR predictor (i.e. first line in \ref{Eq:GPR}) and taking its variance under different dataset draws, {\bf (iii)} The sum of these two. This third variance is the most natural one since it reflects the total variance contribution ($V$) to the dataset averaged loss namely, 
\begin{align}
&\int d\mu_x \langle \langle (f(x)-y(x))^2 \rangle_{S_{\GP}} \rangle_{x_1..x_P} = B + V \\ \nonumber 
B &=  \int d\mu_x (\langle \langle f(x) \rangle_{S_{\GP}} \rangle_{x_1..x_P}-y(x))^2 \\ \nonumber 
V &=\int d\mu_x \langle \langle f^2(x) \rangle_{S_{\GP}} \rangle_{x_1..x_P} - \int d\mu_x \langle\langle  f(x) \rangle_{S_{\GP}} \rangle_{x_1..x_P}^2 
\end{align}
By adding another source term which couples to $\sum_{a=1}^{Q} f_a^2(x)$, one can show that $V$ in our theory is $\langle \int d\mu_x (f(x)-\bar{f}(x))^2 \rangle_{\bar{S}_{\text{GP,EK}}}$ or equivalently $\sum_k \langle (f_k-\bar{f}_k)^2 \rangle_{\bar{S}_{\text{GP,EK}}}$ and therefore given by 
\begin{align}
V &\stackrel{\text{EK}}{\approx} \sum_k \frac{1}{\lambda_k^{-1}+\frac{P}{\kappa^{2}}} = \frac{\kappa^2}{P} \sum_k \frac{\lambda_k}{\lambda_k+\frac{\kappa^{2}}{P}}
\end{align}
which can be roughly interpreted as the number of learnable modes (i.e. modes with $\lambda_k P > \kappa^2$) over the effective number of data points ($P/\kappa^2$). We note that the same result would have been obtained had we just taken a second derivative w.r.t. to the  $\alpha$ source term, thereby obtaining type (i)  contribution. We thus find a qualitative limitation of the EK limit which is that it neglects type (ii), dataset-choice-induced, variances. This could have been anticipated since it is exact in the limit of infinite $P$ (with $P/\kappa^2$ fixed) where all the randomness induced by the choice of dataset disappears. Another issue with this limit is that it breaks down in the ridgeless limit $\kappa^2$ (e.g. for NTK). Notwithstanding, it is accurate in its regime of validity, which is quite a sensible regime (small loss over ridge). Furthermore, as we shall see later, other, more advanced approximation techniques which are valid at vanishing ridge end up looking like EK with an effective ridge parameter. 

The EK approximation thus exposes a central qualitative aspect of the problem, which is that learning is done diagonally in kernel eigenfunctions and there is no ``cross-talk'' between eigenfunctions. Let us delve into this point a bit more. Since the GPR predictor is linear in the target, its dataset average must also be some linear functional of the target. However, while this linear functional could be diagonalized leading to some spectral filtering of the target, this spectrum and its associated eigenfunctions, did {\it not} have to coincide with the eigenbasis of the kernel or be independent of $P$ and $\kappa^2$. The fact that for all $P$, it is the kernel eigenfunctions and spectrum which control the induced filter on the target, is sometimes referred to as eigenlearning \citep{simon2021} or spectral bias.
As we shall shortly see, while at low input dimensions and low ridge, eigenlearning may break down, it appears to be quite robust at large effective input dimension, which is the more common use case of GPR in the context of deep learning.

\subsection{Perturbative treatment}
A straightforward approach to improve upon the previous EK limit is to view it as a perturbative expansion in $1/\kappa^2$ and take the next to leading order, namely the $1/\kappa^4$ contribution. Specifically, this means splitting $\bar{S}_{\GP}$  into its free part (last line in Eq. \ref{Eq:Approx_S_GP}) plus the following interaction term 
\begin{align}
U &= -\frac{P}{4 \kappa^4} \int d\mu_x \left(\sum_{a=1}^Q [f_a(x)-y(x)]^2\right)^2
\end{align}
To compute the first moment of the data averaged GPR predictor under this perturbative correction one needs to integrate over $Q$ path integrals in replica space. Performing this calculation with the limit $Q\to 0$, it emerges that simply picking one replica is sufficient. 

Using leading perturbation theory approach (Sec. \ref{Sec:PT}) one then obtains (\cite{cohen2021learning}, Appendix I)

\begin{align}
&\langle f_a(x_0) \rangle_{\bar{S}_{\GP}}= \langle f(x_0) \rangle_{\bar{S}_{\text{GP,EK}}} \\ \nonumber &
-\frac{P}{\kappa^4} \sum_{ijk}  \frac{\frac{\kappa^2}{P}}{\frac{\kappa^2}{P} + \lambda_i} y_i \phi_j(x_0) V_{j} V_k\left(\int d\mu_x \phi_i(x)\phi_j(x)\phi^2_k(x)\right) + O(\kappa^{-6}) \\ \nonumber 
V_i &= \frac{1}{\lambda_i^{-1} + \frac{P}{\kappa^2}}.
\end{align}

To familiarize ourselves with the above expression we note that: {\bf (i)} The factor $\frac{\frac{\kappa^2}{P}}{\frac{\kappa^2}{P} + \lambda_i} y_i$, is related to the bias component of the loss of $i$'th mode according to EK, hence the correction increases if performance according to EK is poor. {\bf (ii)} $V_k$ is the variance contribution of the $k$'th mode, hence the correction increase is the variance contribution to the loss is high according to EK. {\bf (iii)} Non-linear correlations between eigenfunctions enter this correction and may, together with the $y_i \phi_j(x_0)$ non-diagonal factor, generate cross-talk between features. 

Interestingly, given a high dimensional input measure, one can argue that this cross-talk vanishes. Here, we assume $\phi_k(x)$ to be a sum over many ($O(d)$ or higher) different components of $x$ at different powers. If so, it is reasonable to treat $\phi_k(x)$, via the central limit theorem, as a Gaussian random variable over $p(x)$. Assuming for simplicity $\int d\mu_x \phi_k(x)=0$ (i.e. centered Gaussian), the above non-linear correlation simplifies via Wick's theorem to 
\begin{align}
&\int d\mu_x \phi_i(x)\phi_j(x)\phi^2_k(x) \stackrel{d \gg 1}{\approx} \int d\mu_x \phi_i(x)\phi_j(x) \int d\mu_x \phi_k^2(x) \\ \nonumber 
&+ 2\int d\mu_x \phi_i(x)\phi_k(x) \int d\mu_x \phi_j(x) \phi_k(x) = \delta_{ij}+\delta_{ik} \delta_{jk}
\end{align}
consequently, $i=j$ and cross-talk is eliminated. However, at low dimensions cross talk is possible, see for example \citep{Tomasini2022,howard2024}. Removing the $\int d\mu_x \phi_k(x)=\bar{\phi}_k=0$ restriction, cross-talk re-emerges since even for $i\neq j \neq k$ the integral would be $-\bar{\phi}_i\bar{\phi}_j(1+\bar{\phi}^2_k)$. In practice, however, $\bar{\phi}_j$ decays very quickly with $k$ and practically vanishes for $j\gg 1$. \footnote{Indeed $\bar{\phi}_j$ is also the overlap of the constant function with the $\phi_j(x)$ feature. Typically, the constant function is easy to learn hence spanable by the top kernel eigenfunctions and thus orthogonal to the rest.} At the same time, the terms related to the loss (items (i) and (ii)) above, typically vanish for $j,i=O(1)$ and large $P$, hence, in practice, cross-talk may be ignored. 

\subsection{Effective ridge treatment}
\label{SSec:Canatar}
In the case of small gradient noise (i.e. small $\kappa^2$) relevant, in particular, to NTK one cannot rely on the above perturbative treatment. Instead, several authors \citep{Tsigler2020,cohen2021learning,Canatar2021} noted that, at high input dimensions, the fluctuations of the unlearnable/low kernel eigenvalues simply generate an effective ridge parameter for the learnable kernel eigenvalues. Here we will re-derive the most general of those (\cite{Canatar2021, Sollich2001}) using our field theory approach. 

The main underlying assumption of Ref. \cite{Canatar2021} is that $f_a(x)-y(x)$, appearing within the interaction term in Eq. (\ref{Eq:IntroducingSBar}), is Gaussian when viewed as a random variable over $p(x)$. More specifically, a $Q$-variate Gaussian, as there could be correlations between different replicas. This {\it Gaussian Discrepancy Assumption} follows from our previous Gaussian-eigenfunctions assumption but is generally a weaker requirement. We further assume, for simplicity, that $\int d\mu_x [f_a(x)-y(x)]=0$, which amounts to taking an anti-symmetric DNN or assuming that we quickly learn the constant component in $y(x)$ perfectly without any fluctuations (indeed it is typically easy to learn). Introducing the discrepancy variables $\Delta_a$ via the identity $1=\int d\Delta_a  \delta(\Delta_a-f_a(x)+y(x))$ we find 
\begin{align}
\label{Eq:PoissonInteraction}
&P \int d\mu_x e^{-\frac{1}{2\kappa^2}\sum_{a=1}^Q [f_a(x)-y(x)]^2} \\ \nonumber 
&= P \int d\mu_x \prod_{a=1}^Q \left[\int d\Delta_a \delta(\Delta_a-f_a(x)+y(x))\right]e^{-\frac{1}{2\kappa^2}\sum_{a=1}^Q [f_a(x)-y(x)]^2}\\ \nonumber 
&\approx P\int d\Delta_{1}..d\Delta_Q \; {\cal N}(0,C;\Delta) \; e^{-\frac{\Delta^T \Delta}{2 \kappa^2}} 
= P e^{-\frac{1}{2} \Tr\log(I+ \kappa^{-2} C)} = ...\\ \nonumber 
C_{ab} &= \int d\mu_x \; [f_a(x)-y(x)][f_b(x)-y(x)] = \sum_{k=1}^{\infty} [f_{k,a}-y_k][f_{k,b}-y_k]
\end{align}
where the approximation in the second line is exact for Gaussian features,
$I$ is an identity matrix in replica space ($I_{ab}=\delta_{ab};a,b\in 1..Q$), in the second equality we view $\int d\mu_x \prod \delta(\Delta_a - f_a(x)-y(x))$ as the joint distributions of the $\Delta_a$'s which we assume is Gaussian, in the third equality we used standard Gaussian integration formulas together with the identity $\det B = e^{\text{Tr} \log B}$, and in the last equality we used a spectral decomposition and the orthogonality of features/eigenfunctions. 

Since $C_{ab}$, the replica correlations of discrepancy over the data measure, are made up of a contribution of many independent $k$-modes, it makes sense to try a mean-field approximation. Specifically, following our previous section  \ref{ssec:mean_field}, we define a mean-field kernel $C=C_{\MF}+C-C_{\MF}\equiv C_{\MF}+\Delta C$, plug this in the above term, and expand to leading order in $\Delta C$. Using the Taylor expansion $e^{-\frac{1}{2}\Tr\log(M+X)}=e^{-\frac{1}{2} \Tr\log(M)}(1-\frac{1}{2}\Tr[M^{-1}X])+O(X^2)$ with $X=\Delta C$ and $M=I+\kappa^{-2} C_{\MF}$ yields 
\begin{align}
&... = \frac{Pe^{-\frac{1}{2}\Tr\log(I+\kappa^{-2}C_{\MF})}}{2} \Tr[(\kappa^2 I+C_{\MF})^{-1}\Delta C] \\ \nonumber &= \frac{Pe^{-\frac{1}{2}\Tr\log(I+\kappa^{-2}C_{\MF})}}{2} \sum_k \sum_{ab=1}^Q [f_{k,a}-y_k][(\kappa^2 I+C_{\MF})^{-1}]_{ab}[f_{k,b}-y_k] + \const \\ \nonumber 
&=|_{Q \rightarrow 0}\frac{P}{2} \sum_k \sum_{ab=1}^Q [f_{k,a}-y_k][(\kappa^2 I+C_{\MF})^{-1}]_{ab}[f_{k,b}-y_k] \end{align}
where in the last equality we ignored the irrelevant additive constant term and, anticipating the zero replica limit, took $\Tr\log(I+\kappa^{-2}C_{\MF})=O(Q)$ to zero upfront. 

The resulting data term is conveniently quadratic in the fields and diagonal in the kernel eigenmodes. Consequently, computing averages according to this theory (i.e. Eq. \ref{Eq:IntroducingSBar} with the interaction term replaced by the above) is straightforward apart from resolving the coupling between replicas. Expecting a replica symmetric solution, we may write $C_{\MF}=A I+ D \Gamma$ where $\Gamma_{ab}=1$. Noting that $\Gamma^2=Q\Gamma$ and $(a I + b \Gamma)(a^{-1} I - \frac{b}{a^2} \Gamma)=I-Q\frac{b^2}{a^2} \Gamma$, we obtain the following action which coincides with the original one in the replica limit ($Q\rightarrow 0$) 
\begin{align}
&\bar{S}_{\GP,\MF} = \sum_k \sum_{a=1}^Q \frac{f_{a,k}^2}{2\lambda_k} \\ \nonumber &+ \frac{P}{2} \sum_k \sum_{ab=1}^Q [f_{k,a}-y_k]\left[(\kappa^2+A)^{-1} I+\frac{D}{(\kappa^2+A)^2}\Gamma\right]_{ab}[f_{k,b}-y_k] 
\end{align}
Using similar matrix inversion formula, dropping $Q$ factors, and denoting $\kappa_{\text{eff}}^2=\kappa^2+A$ one finds 
\begin{align}\label{eq:barf}
\langle f_{k,a}\rangle_{\bar{S}_{\GP,\MF}} &= \frac{\lambda_k}{\lambda_k+\kappa^2_{\text{eff}}/P} y_k
\end{align}
which coincides with EK prediction using an effective ridge $k^2_{\text{eff}}$. For the variance, we obtain, 
\begin{align}\label{eq:f_k_f_k_connected}
\langle f_{a,k} f_{b,k} \rangle_{\bar{S}_{\GP,\MF},\con} &= \left( \frac{P} {\kappa^{2}_{\eff}} + \lambda_k^{-1} \right)^{-1} \delta_{ab} + \left( \frac{P} {\kappa^{2}_{\eff}} + \lambda_k^{-1} \right)^{-2}\frac{P}{\kappa^{4}_{\eff}} D 
\end{align}
which contains an EK-like contribution to the variance (the first term) along with a new contribution that reflects the aforementioned type (ii) contributions to the variances missed by EK. 

What remains is to obtain $A,D$ which determine $C_{\MF}$ or namely to solve the self-consistency equation
\begin{align}
&[AI+D\Gamma]_{ab} = [C_{\MF}]_{ab} = \sum_k \langle [f_{k,a}-y_k][f_{k,b}-y_k] \rangle_{\bar{S}_{\GP,\MF}} \\ \nonumber 
&= \sum_k \langle f_{a,k} f_{b,k} \rangle_{\bar{S}_{\GP,\MF},\con} 
+ \sum_k [\langle f_{k,a}\rangle_{\bar{S}_{\GP,\MF}}-y_k][\langle f_{k,b}\rangle_{\bar{S}_{\GP,\MF}}-y_k]. 
\end{align}
Plugging Eqs. \ref{eq:barf},\ref{eq:f_k_f_k_connected} in the above, we note that contribution proportional to $I$ on the r.h.s. comes solely from the variance term and yields 
\begin{align}
\label{Eq:CanatarEffectiveRidge}
A &= \sum_k \left(\frac{P} {\kappa^{2}_{\eff}}+\lambda_k^{-1}\right)^{-1} \Rightarrow \kappa_{\eff}^2 =  \kappa^2 + \sum_k \left(\frac{P} {\kappa^{2}_{\eff}}+\lambda_k^{-1}\right)^{-1}
\end{align}
which has a simple interpretation: The effective ridge is the bare ridge ($\kappa^2$) together with the posterior-variance (i.e. $V_k$) of all modes in the presence of the effective ridge and $P$. Turning to $D$ we find two contributions proportional to $\Gamma$ on the r.h.s. yielding 
\begin{align}
\label{Eq:D}
\left(1-\sum_k \left( \frac{P}{\kappa_{\eff}^2} + \lambda_k^{-1} \right)^{-2}\frac{P}{\kappa^{4}_{\eff}} \right)D &= \sum_k \left(\frac{\lambda_k}{\lambda_k+\kappa^2_{\eff}/P} y_k\right)^2
\end{align}
which, given $\kappa^2_{\eff}$, is a simple linear equation for $D$. It can also be shown that the first factor on the l.h.s. is always larger than zero.

\subsection{Renormalization group treatment}
Last, we review a recent approach \citep{howard2024} which utilizes renormalization group ideas to tackle the same problem. The renormalization group (RG) is a powerful analytical technique used extensively in physics. At its core, it is a greedy-algorithm for performing path integrals: First, we integrate over a small set of field modes. These should be chosen wisely so that their integration/marginalizing is simple. A common choice is modes with small variance, allowing one to treat all interaction terms involving such modes via perturbation theory. The result of this partial integration is a different representation of the partition function, which nonetheless reproduces the correct statistics for all the remaining field modes. This new partition function can be described by a slightly different action (renormalized/effective action) coming from the feedback of those modes we integrated out on the remaining ones. Interestingly, in many cases, we can track the change (flow) of this effective action as we progressively integrate-out more and more field modes. The resulting effective action is often simpler than the original one, thus providing a computational advantage. These flows of actions may also be contracting in the space of effective actions, which is the underlying mechanism of universality in physics. Here we shall use this RG approach to reproduce some of the results of the previous section. 

Our starting point is the following form of $\bar{S}_{\GP}$ obtained from Eq. \ref{Eq:PoissonInteraction} we limit the summation over eigenmodes to some potentially huge but finite cut-off ($\Lambda$)
\begin{align}
\label{Eq:SLambda}
&\bar{S}_{\GP,\Lambda} = \sum_{k=0}^{\Lambda} \sum_{a=1}^Q \frac{f_{a,k}^2}{2\lambda_k} +P e^{-\frac{1}{2} \Tr\log\left(\kappa^2 I+ \sum_{k=1}^{\Lambda} (\bm{f_k}-\bm{y_k})(\bm{f_k}-\bm{y_k})^T\right)} \\ \nonumber 
Z_{\Lambda} &=  \int \left[\prod_{a=1}^Q df_{1,a}..df_{a,\Lambda}\right] e^{-\bar{S}_{\GP,\Lambda}}
\end{align}
where we used the {\it Gaussian Discrepancy Assumption} (Eq. \ref{Eq:PoissonInteraction}), took the replica limit upfront by pulling out a $\Tr\log(\kappa^{-2} I)$ term from the exponent, we further sort $k$ such that $\lambda_k$ come in descending order and also introduce the replica space vector notation $[\bm{f_k}]_a=f_{k,a}$ and $[\bm{y_k}]_a=y_k.$ 
Next, we consider integrating out $f_{k,a}$ modes with $k = \Lambda$, assuming that $\lambda_{\Lambda},y^2_{\Lambda}\ll \kappa^2$. To this end we split the interaction term into its so-called lesser part (i.e. involving only modes we keep i.e. $k < \Lambda$) and its greater part which involves the modes we are about to integrate over
\begin{align}
e^{-\frac{1}{2}\Tr\log\left(\kappa^2 I+ \sum_{k=1}^{\Lambda-1} (\bm{f_k}-\bm{y_k})(\bm{f_k}-\bm{y_k})^T+(\bm{f_{\Lambda}}-\bm{y_{\Lambda}})(\bm{f_{\Lambda}}-\bm{y_{\Lambda}})^T\right)} &= ...
\end{align}
Knowing that the first term in the action limits $f_{a,\Lambda}^2$ to order $\lambda_{\Lambda} \ll \kappa^2$ and assuming also that $y^2_{\Lambda} \ll \kappa^2$ it is reasonable to treat $(\bm{f_{\Lambda}}-\bm{y_{\Lambda}})(\bm{f_{\Lambda}}-\bm{y_{\Lambda}})^T$ to leading order in perturbation theory namely, 
\begin{align}
\label{Eq:RGTaylorExpansion}
... &= e^{-\frac{1}{2} \Tr\log\left(\kappa^2 I+ \sum_{k=1}^{\Lambda-1} (\bm{f_k}-\bm{y_k})(\bm{f_k}-\bm{y_k})^T\right)}[1 + U] \\ \nonumber 
U &= \Tr\left[\left(\kappa^2 I+ C_{\Lambda-1}
\right)^{-1}(\bm{f_{\Lambda}}-\bm{y_{\Lambda}})(\bm{f_{\Lambda}}-\bm{y_{\Lambda}})^T \right]
\\\nonumber 
&C_{\Lambda-1} = \sum_{k=1}^{\Lambda-1} (\bm{f_k}-\bm{y_k})(\bm{f_k}-\bm{y_k})^T
\end{align}
Since $U$ is proportional to the highest, GP modes ($f_{\Lambda}$) which is small in average and variance compared to $\kappa^2$ and the highest target mode which we assume is similarly small compared to $\kappa^2$, we may  
 perform the integration over $d \bm{f_{\Lambda}}$ to order $O(U)$ and re-exponentiate the $O(U)$ contribution to obtain 
\begin{align}
&Z_{\Lambda-1} = \int \left[\prod_{a=1}^Q df_{1,a}..df_{a,\Lambda-1}\right] e^{-\bar{S}_{\GP,\Lambda-1}} \\ \nonumber 
&\bar{S}_{\GP,\Lambda-1} = \sum_{k=1}^{\Lambda-1} \sum_{a=1}^Q \frac{f_{a,k}^2}{2\lambda_k} + Pe^{-\frac{1}{2} \Tr\log\left(\kappa^2 I+ \sum_{k=1}^{\Lambda-1} (\bm{f_k}-\bm{y_k})(\bm{f_k}-\bm{y_k})^T\right)}\times \\ \nonumber 
&\left[1+\Tr\left[\left(\kappa^2 I+ C_{\Lambda-1}\right)^{-1}\langle (\bm{f_{\Lambda}}-\bm{y_{\Lambda}})(\bm{f_{\Lambda}}-\bm{y_{\Lambda}})^T\rangle_{{\bm f_{\Lambda}} \sim {\cal N}(0,\lambda_{\Lambda}I_Q)} \right]\right]
\end{align}
where $\langle ... \rangle_{\bm{f}_{\Lambda} \sim {\cal N}(0,\lambda_{\Lambda}I_Q)}$ comes from integrating over/out the ${\bm f}_{\Lambda}$ using the action in Eq. \ref{Eq:SLambda} with the interaction appearing in \ref{Eq:RGTaylorExpansion} at $U=0$, where it becomes a single Gaussian field. This averaging yields $\lambda_{\Lambda} I + y^2_{\Lambda} \Gamma$ where $\Gamma_{ab}=1$. Next we use the fact that to leading order in $U$ we may undo the Taylor expansion of $e^{-\frac{1}{2}\Tr\log(..)}$ we obtain 
\begin{align}
\bar{S}_{\GP,\Lambda-1} = \sum_{k=1}^{\Lambda-1} \sum_{a=1}^Q \frac{f_{a,k}^2}{2\lambda_k} + Pe^{-\frac{1}{2} \Tr\log\left((\kappa^2+\lambda_{\Lambda}) I+ y^2_{\Lambda}\Gamma + \sum_{k=1}^{\Lambda-1} (\bm{f_k}-\bm{y_k})(\bm{f_k}-\bm{y_k})^T\right)}
\end{align}
where we recognize the appearance of a renormalized ridge ($\kappa^2_{\RG,\Lambda-1}=\kappa^2+\lambda_{\Lambda}$). 
This process can be repeated recursively, reducing $\Lambda$ to some $\Lambda'$, until we reach the point our perturbative assumption breaks down, namely $\lambda_{\Lambda'} \not \ll \kappa^2_{\RG,\Lambda'}$ or $y_{\Lambda'}^2 \not \ll \kappa^2_{\RG,\Lambda'}$. However, provided this does not happen too soon when we reach $\Lambda'$ we find an effective ridge ($\kappa^2_{\RG}$) equal to 
\begin{align}
\label{Eq:Effective_Ridge_RG}
\kappa_{\RG}^2 &= \kappa^2_{\RG,\Lambda'} =   \kappa^2+\sum_{k>\Lambda'}^{\Lambda}\lambda_k 
\end{align}
where it is sensible to choose $\Lambda'$ as the index associated with the minimal unlearnable mode, given $\kappa_{\RG}^2$. Thus, similar to Eq. \ref{Eq:CanatarEffectiveRidge}, determining what is $\Lambda'$ requires solving a non-linear equation. Here it is given by the above equation together with the requirement $P\lambda_{\Lambda'}/\kappa_{\RG}^2 = \epsilon$, where $\epsilon \ll 1$ is some arbitrary threshold value defining what we mean by an unlearnable mode.  

It is interesting to compare the above result with Eq. \ref{Eq:CanatarEffectiveRidge}. In both equations the contribution of unlearnable modes ($\kappa^2_{\eff/\RG}/P \ll \lambda_k$), to, $k_{\eff/\RG}^2$ is identical. Furthermore, stopping our RG when the above condition is violated, then using Eq. \ref{Eq:CanatarEffectiveRidge} with $\kappa^2=\kappa_{\RG}^2$ would lead to the same final result. However, provided $\kappa^2_{\RG}$ becomes large enough, we can simply carry on directly using EK or perhaps EK along with a perturbative correction. The latter approach was used in Ref. \cite{cohen2021learning}, although using a very different heuristic justification which was also limited to infinite fully connected networks, and led to good prediction for ridgeless GPR with the NTK. In Ref. \cite{howard2024}, it was also generalized to non-Gaussian $f_a(x)-y(x)$, where evidence for cross-talk at low dimensions and beyond spectral-bias effects appeared again. Interestingly, this entered as a $x$-dependent effective ridge parameter in the effective action.   

{\bf Related works.} Finally, we review several other works involving RG and neural networks based on several criteria (i) What is the scale or what is being integrated out? (ii) What is the microscopic (UV) theory on which one is applying RG? (iii) Is the induced RG flow tractable, or does it make some quantities more tractable? 

An early line of work sought to augment (or interpret) the action of a layer in deep neural networks so that it coarse-grains the data distribution, thus drawing an analogy with RG \cite{mehta2014exact,Koch_Janusz_2018,Beny:2013pmv}. Similarly, the action of a diffusion model can be interpreted as inverting an RG flow on the data distribution \cite{Cotler_2023}. The UV theories here are thus the data distribution and the IR theory consists of the relevant features in the data for the layer action and white noise data for the diffusion action. Another line of work \cite{Erbin_2022,erbin2023functional} 
views the distribution on functions induced by a random DNN as the UV theory, and uses RG formulations to treat non-Gaussian (finite-width) effects \cite{Halverson:2020trp, Grosvenor:2021eol,Roberts_2022} 
in this distribution. A route for RG where network depth is coarse-grained was put forward in Ref. \cite{Lee2024DynamicNeurons}. 
Another set of works relating RG flows and Bayesian Inference view both as forms of augmenting a probability distribution  \cite{berman2022dynamics,berman2022inverse,Berman_2023,berman2024ncoder,howard2024bayesianrgflowneural}. In this approach however, reducing data generates the RG flow, so the true data-generated model (or target function) is the UV theory, while the  IR theory is a random DNN, so that observables that are consistently reproduced along the flow are the most unlearnable modes.
To the best of our understanding, these works have yet to analytically predict the generalization performance of neural networks. Finally, an earlier work \cite{Bradde_2017} models the data distributions, irrespective of any neural network, as a field theory performs RG by removing large PCA components. Trading PCA with kernel-PCA, both that work and the current work share the same notion of UV and IR modes, however focus on very different ensembles/probability distributions. In the analysis above, the UV theory is the distribution over outputs generated by an ensemble of trained neural networks at a finite amount of data, while the IR is the theory for the learnable modes. To generate an RG flow, we integrate out unlearnable output modes and track their renormalizing effect on the learnable modes.

\subsection{Revisiting the optimization problem}
At this point we can offer a quantitative solution to one of the fundamental questions in deep learning--- given that so many zero train loss minima exist, and that DNNs are generally capable of learning any function \citep{HORNIK1989,zhang2021understanding}, what biases them during optimization to choose the well-generalizing ones? 

Technically, we have so far seen that NTK dynamics leads to an output which is the sum of two terms: (i) The average GPR predictor using NTK kernel with zero ridge and (ii) the $I_0$ fluctuation term which corresponds to the average GPR predictor using the NTK kernel trying to learn the initial output of the trained DNN. The first term will exhibit a good learning curve (drop of error with $P$) if the target is supported on kernel eigenvalues ($\lambda$) obeying $\lambda \gg \int d\mu_x K(x,x)/P > \sigma_{\eff}^2/P$. The second would be small if the outputs at initialization show similar support. 

A more tangible viewpoint of how implicit bias is generated by the dynamics can be obtained as follows. Revisiting the NTK section, one notes that the NTK kernel is proportional to $J^T(x) J(x')$ where $J_{\alpha}(x) = \partial_{\theta_{\alpha}} z_{\theta}(x)$. Thus the NTK $P$-by-$P$ matrix' eigenvalues are the square of the Singular Values of the generally rectangular matrix $[J^T]_{\mu \alpha}=\partial_{\theta_{\alpha}}z_{\theta}(x_{\mu})$. Specifically, we can associate a weight-space singular vector $v$ of $J^T$, with an eigenvector of the NTK matrix ($\phi_{\mu}$) via $\phi = J^T v$, as one can verify algebraically. Next, we note that as $P$ grows large, one generally expects these matrix eigenvalues to be $P$ times the corresponding NTK operator eigenvalues w.r.t. to the data measure (indeed one can show that the discrete eigenvalue equation tends to the operator eigenvalue equation in this limit). Similarly, we expect matrix eigenvectors would be close to NTK-operator eigenfunctions sampled on the data. We thus associate a high singular vector of $J^T$ ($v_{\alpha}$) with a high-eigenvalue eigenfunction of the NTK operator ($\phi(x)$), via the $P \rightarrow \infty$ limit of $\phi =J^T v$. Finally, viewing $J^T$ as the linear Taylor expansion coefficient of network outputs around the initial weights, having $v$ with a large eigenvalue implies that a small motion in weight space along this direction produces a relatively large change in output along this functional direction. Thus we need to travel less in weight space to generate functions associated with high NTK-operator eigenvalues. This biases the optimization procedure towards loss minima which utilize these types of functions.  

\section{Application: Symmetries and spectral bias}
\label{Sec:Symmetries}
One of the main lessons from the previous section is that over-parametrized DNNs in standard scaling \footnote{i.e. pre-activations and outputs being $O(1)$ at initialization or according to the prior as considered so far}, often \footnote{At high input dimension or large ridge.} have an implicit bias (a.k.a. spectral bias) towards learning function associated with high eigenvalues of the kernel. Thus, if we had the diagonalization data for the kernel  $(\lambda_k,\phi_k(x))$ on the true data measure and the true structure of the target $y(x)$ for all $x$, we could predict the training dynamics, the learning curves, and the test loss. Conveniently, all the complex dynamics involved in doing inference resulted in a linear high-pass filter in this $\phi_k(x)$ basis. 

We begin with a discussion putting these results in practical context. The obvious elephant in the room is that the above diagonalization data is analytically intractable and computationally costly. Furthermore, there is a conceptual difficulty which is that, while we can readily obtain concrete expressions for the kernel \citep{lee2019wide}, the data measure is only partially accessible to us, as we only see the finite dataset. We can gather more data points if needed, but this still falls short of a concrete expression for $p(x)$, the true distribution of the data. In practice, however, the high and learnable $\lambda_k$'s are well resolved by the finite data available (i.e. would change negligibly if we add more data). Furthermore, corresponding eigenvectors can be seen as discrete samples of $\phi_k(x)$. \footnote{These can be extended into functions using GPR with $K$ as the kernel and $\phi_k(x)$ as the target.} Thus, we can get numerical access to the dominant kernel eigenmodes at an $O(P^3)$ (matrix diagonalization) computational cost. The more problematic issue is that there is no real computational or interpretation-wise benefit in diagonalizing a kernel versus doing actual GPR. In fact, we mainly replaced the complexity of understanding DNN training with the complexities of characterizing a set of functions  ($\phi_k(x)$) in high dimension. 

However, the picture is also not as stark as the above implies. For instance, ({\bf i}) considering toy target functions and toy datasets \citep{cohen2021learning,Canatar2021,lavie2024towards,yang2019fine}, $\phi_k(x)$'s can be understood analytically, and spectral bias becomes tractable. As such toy data models capture qualitative aspects of real-world DNN behavior (e.g. effects of depth on over-fitting \citep{yang2019fine}, relating power laws in performance to power laws in the data \citep{Bahri2024Explaining}, and relations between architecture choices, symmetries in the data, and performance \citep{novak2018bayesian})--- spectral-bias provides qualitative explanations for phenomena occurring in real world data. As we improve the theoretical modeling of data \citep{Francesco2024}, we may expect this portfolio of applications to increase, with spectral-bias providing the link between data, kernel, and performance.  ({\bf ii}) Turning from predictions to bounds, Ref. \cite{lavie2024symmetric} derived a learnability lower bound on the number of samples ($P_*$) involved in learning target functions which are sums of $O(1)$ different $\phi_k(x)$'s. Importantly, the $\phi_k(x)$ can be derived from any convenient, $p_{\text{ideal}}(x)$ while the learnability bound applies to any $p(x)$. This allows, for instance, to bound the time it would take a transformer in the NTK regime acting on standard Language data to learn ``copying heads'' \citep{olsson2022context}, an important target component which aids in-context learning. 

The above discussion motivates the need to diagonalize kernels on relatively simple data measures, as means of understanding behavior on real-world data. This has worked out for datasets which are uniform on the hypersphere \citep{Azevedo2015EigenvaluesOD}, hypercube \citep{yang2019fine}, or for sequence data that is permutation invariant \citep{lavie2024towards}. Here we explain in some detail the case of the hypersphere using the language of representation theory, which carries through to the other two cases as well. 

Consider a fully connected DNN with $x \in \R^{d_{\text{in}}}$ and $p(x)$ obeying $p(x)=p(Ox)$ where $O$ is any orthogonal transformation from the orthogonal group ${\rm O}(d_{\text{in}})$ represented by a rotation matrix $O$. As discussed briefly in Sec. \ref{Sec:Intro_Measure}, an FCN kernel together with such a measure constitutes a symmetric operator ($\hat{K}$) acting on function space. Next, we promote $O \in {\rm O}(d_{\text{in}})$ to a linear operator on function space via $\hat{O} f(x)\equiv f(Ox)$ as well as $\hat{O} K(x,y)=K(Ox,y)$ and $K(x,y)\hat{O}=K(x,yO)=K(x,O^T y)$. These definitions together with the symmetry of $K$ ($K(Ox,Oy)=K(x,y)$) imply that the commutation relation $[\hat{K},\hat{O}]\equiv \hat{K}\hat{O}-\hat{O}\hat{K}$ vanishes. When two matrices/operators commute, they preserve each other's eigenspaces. When an operator $\hat{K}$ commutes with an entire group (e.g. ${\rm O}(d_{\text{in}})$) it preserves its {\it irreducible representations} (irreps), which we next define. 

An irreducible representation $R_{\alpha}$ of a group (where $\alpha$ enumerates all possible irreps) associated with a subspace of dimension $d_{\alpha}$, is {\bf (1)} A representation: a set of $d_{\alpha} \times d_{\alpha}$ unitary matrices, one for each element $O \in {\rm O}(d_{\text{in}})$, which obey the algebra of ${\rm O}(d_{\text{in}})$ (i.e. the matrix representing $OO'$ equal the matrix representing $O$ times the matrix representing $O'$.) {\bf (2)} Irreducible: There is no unitary transformation that would transform all representing matrices $O$ to a block diagonal form with identical block structure and more than one block. A block diagonal matrix could be decomposed into two matrices, each acting on one of the subblocks; it would be reducible in this sense. Namely, there is no smaller subspace of this ${\R}^{d_{\alpha}}$-space on which these matrices could be defined. 

The benefit of irreducible representations is that {\bf (1)} function space can be split into a 
linear sum of the spaces associated with each irreducible representation. {\bf (2)} If each irreducible representation ($R_{\alpha}$) appears only once in the previous linear sum, $\hat{K}$ is block diagonal and proportional to the identity matrix within each such subspace \footnote{The latter follows from Schur's lemma, using the linear mapping $\hat{K}-\lambda I$, where $\lambda$ is any one of $K$'s eigenvalues.}. 
As would be elaborated upon in the example below, since $\lambda_k \geq 0$ and $\sum_k \lambda_k=\int d\mu_x K(x,x)=O(1)$, we have a "spectral budget". Irreps of size $d_{\alpha}$, having degenerate eigenvalues, must be associated with eigenvalues of the order $O(1)/d_{\alpha}$ to respect this spectral budget. 
Last we note that representation theory is a well-developed mathematical field \citep{fulton_representation_2004} with many concrete results and practical tools which enable one to find concrete data on these irreducible representations. 

Focusing back on our FCNs, let us take $p(x)=V_{d_{\text{in}}}^{-1}\delta(|x|-1)$ where $V_{d_{\text{in}}}$ is the volume of a hypersphere in $d_{\text{in}}$-dimension. In this case, the space of functions on this hypersphere is spanned by irreducible representations known as hyper-spherical harmonics. Representations are labelled by $\alpha \in {\cal N}_+$, and correspond to homogeneous polynomials of degree $\alpha$ which are in the null-space of the Laplacian operator \citep{Frye2012}. For instance, $\alpha=0$ has $d_{\alpha=0}=1$ and correspond to the constant function, $\alpha=1$ has $d_{\alpha=1}=d_{\text{in}}$ and corresponds to all linear functions, $\alpha=2$ has $d_{\alpha=2}=(d+2)(d-1)/2$ and correspond to the polynomials $x_i x_j$ with $i \neq j$ as well as $x^2_i-|x|^2/d_{\text{in}}$ for $i \in [1..d_{\text{in}}-1]$. One can manually verify that these indeed obey the above requirement of an irreducible representation. More generally, $d_{\alpha}=\frac{2\alpha+d_{\text{in}}-2}{\alpha+d_{\text{in}}-2}\binom{\alpha+d_{\text{in}}-2}{\alpha}$ which scales as $d_{\text{in}}^{\alpha}$ for $\alpha \ll d_{\text{in}}$. Since each such irrep appears only once when spanning the space of functions on the hypersphere, we immediately have that $\hat{K}$ is given by $\lambda_{\alpha}$ within each $d_{\alpha}$ block. 

Next, we can relate the degeneracy or dimension of the block associated with an irrep to its kernel eigenvalue $\lambda_{\alpha}$. Indeed, working with a normalized kernel we take $K(x,x)=O(1)$\footnote{Within an EK approach, the meaningful scale for the average predictor is $K(x,x)/\kappa^2$. Within the effective ridge approach, the ridge would scale as the kernel, hence the scale has no impact on the average predictor and can just as well be taken to $1$} and consequently $\Tr[K]=\int \dif \mu_x K(x,x)=\int \dif \mu_x \lambda_k \phi_k(x) \phi_k(x)=\sum_k \lambda_k = O(1)$. Since $K(x,y)$ represents a covariance, it is a positive semi-definite operator and hence $\lambda_k \geq 0$. Consequently, we have that, within each irrep block $d_{\alpha} \lambda_{\alpha}\leq \Tr[K] \leq O(1)$. Thus, $\lambda_{\alpha}$ scale as $1/d_{\alpha}$. 

We thus obtain an important insight regarding inference with FCNs on high dimensional spaces. Let us analyze this from the effective ridge viewpoint. Say we wish to learn a target function $y(x)$ which is a degree $q$ polynomial in the span of the $\alpha=q$ representation (i.e. a degree $q$ polynomial which vanishes under the Laplacian). For $P$ at which $y(x)$ starts becoming learnable, we can use Eq. \ref{Eq:Effective_Ridge_RG} with $\Lambda'$ set to the first index associated with the $\alpha=q$ irrep, to determine $\kappa_{\eff}^2$. For such $P$, it would be of the scale $d_{\alpha}\lambda_{\alpha}=O(1)$ or larger, where $\lambda_{\alpha}$ is the eigenvalue associated with the $\alpha=q$ representation. Since $\lambda_{\alpha}=O(d_{\text{in}}^{-q})$ we need $P_* > O(\kappa_{\eff}^2/d_{\text{in}}^{-q}) \approx O(d_{\text{in}}^q)$ samples to start learning such modes. Thus FCNs a have strong spectral bias towards extrapolating between training points using low-order polynomials.

Turning to real-world data, consider taking some datasets like imageNet and normalizing all image vectors to have a fixed norm. Since this only puts a single constraint on the image (and also just implies an overall change of brightness) we expect this to have a minor effect of performance. Interestingly, the above results for $P_*$ hold for any FCN even if the data measure is not uniform. Following Ref. \cite{lavie2024symmetric}, the main change for an $O(1)$ target is that the above condition on $P_*$ becomes $P_* > d_{\text{in}}^q[\int V_{d_{\text{in}}}^{-1} \delta(x- |x|)  y^2(x)] /[\int \dif \mu_x y(x)]$. Thus, provided that the normalization of $y(x)$ under the uniform/idealized measure does not scale very differently than under the true measure, the same scaling of $P_*$ with the degree of the polynomial holds. Though imageNet is far from being uniform on a hyper-sphere, it is highly plausible that low order polynomials such as, say, the red-color of the middle pixel times the blue-color of two of its adjacent pixels, would have the same norm under both. This, however, implies, that such a simple local filter would require $P_*$ equal to the dimension of ImageNet to the third power, which is astronomical. Indeed, however, FCN do not yield good results on ImageNet, CNN do. Taking CNN with non-overlapping convolutional-kernel windows, one obtains a much smaller symmetry group proportional to rotation within each patch \citep{naveh2021self}. These smaller groups would have $d_{\alpha}$ scaling as the patch dimension (typically 27) to the power of the polynomial rank, making learning such local features feasible with $27^3 \approx 20k$ data points. The first steps towards applying a similar approach to Transformers, taking the idealized dataset measure as permutation invariant in the sequence index, have been taken in Ref. \cite{lavie2024towards}.

\section{Application: Scaling laws}
\label{Sec:ScalingLaws}
A question of central practical importance is how to predict the performance of a large trained model, requiring expensive computing, based on the training of smaller, cheaper, ones. An interesting empirical observation \citep{kaplan2020scalinglawsneurallanguage} found in training large language models is that performance as a function of model size and training set size, is well approximated by a sum of two power-law decays (a scaling law) specifically 
\begin{align}
{Loss}(P,N_{\text{params}}) &= \left[ \left(\frac{N_0}{N_{\text{params}}}\right)^{\frac{\alpha_{N}}{\alpha_P}}+\left(\frac{P_0}{P} \right) \right]^{\alpha_{P}}
\end{align}
where $N_{\text{params}}$ is the number of model parameters, $N_0,P_0$ are fitting constants, and $\alpha_P$ ($\alpha_N$) are the exponents associated with the decay w.r.t $P$ at $N_{\text{params}}
\rightarrow \infty$ (w.r.t. $N_{\text{params}}$ at $P \rightarrow \infty$). 

While making analytical predictions on realistic large language models (LLMs) is challenging, those models too have a GPR limit when the number of attention heads goes to infinity. One can therefore wonder to what extent can scaling laws can be explained using the tools developed in this chapter. This has been studied in Ref. \cite{Bahri2024Explaining}, under the reasonable assumption that the spectrum of the kernel operator is power-law distributed namely 
\begin{align}
\lambda_k &= \lambda_1 k^{-1-\alpha} \,\,\,\ \alpha > 0 
\end{align}
This assumption is loosely related to Zipf's law scaling of word frequencies in natural languages or power-law decay of PCA data on images and can be verified experimentally. We also assume that the target is similarly distributed namely that $y^2_k \approx y_1^2 k^{-1-\alpha}$. 

Next, we show how the power-law $\alpha$ is related to $\alpha_P$ in the large ridge regime (EK limit) and in the vanishing ridge regime.  
To this end we introduce the notion of a threshold mode $k_T$ that distinguishes the learnable modes $k < k_T$ from those that are not learnable $k \ge k_T$. A reasonable choice for this mode is given by the point where $\kappa^2/P = \lambda_{k_T}$ at which the learnability factor in \ref{Eq:EK_Av_Pred} is $1/2$. So $k_T = (P/\kappa^2)^{1/(1+\alpha}$.

Given our assumption on the target, we can estimate the loss that is dominated by the discrepancies of the unlearnable modes ($\sim y_k$) via $\sum_{k > k_T}^{\infty} y^2_k \approx \int_{k_T}^{\infty} \dif k \; k^{-(1+\alpha)} \propto k_T^{-\alpha} \propto P^{-\alpha/(1+\alpha)}$. Therefore, we find that $\alpha_P=\alpha/(1+\alpha)$. 

Next, we consider how the effective noise approach (specifically in the RG approximation) changes this scaling for $\kappa^2=0$. After integrating out all modes down to some $\Lambda'$ cutoff, we obtain $\kappa^2_{\text{RG}}=\sum_{\Lambda'}^{\infty} k^{-(1+\alpha)} \approx \alpha^{-1} (\Lambda')^{-\alpha}$. The RG flow remains tractable, in the sense of maintaining $\lambda_{\Lambda'}/\kappa^2_{\RG,\Lambda'} \ll 1$, as we have $\kappa^2_{\text{RG},\Lambda'} \propto (\Lambda')^{-\alpha} \gg \lambda_{\Lambda'}=(\Lambda')^{-1-\alpha}$. At this point, we undertake a self-consistent assumption that the growth of $\kappa_{\text{RG},\Lambda'}^2$ puts us sufficiently close to the EK limit so that we can evaluate the threshold for learnable modes via $P=\kappa_{\text{eff}}^2(k_T)/\lambda_{k_T}\approx \alpha^{-1} k_T^{-\alpha}/k_T^{-(1+\alpha)}=\alpha^{-1}k_T$. This leads to a reasonable result which shows that the number of learnable modes scales linearly with the dataset size. Evaluating the MSE loss by resumming all unlearnable modes of $y_k$, we now obtain $\sum_{k > k_T}^{\infty} y^2_k \approx \int_{k_T}^{\infty} \dif k \; k^{-(1+\alpha)} \propto k_T^{-\alpha} \propto P^{-\alpha}$, which is the data exponent $\alpha_P$ (there $\alpha_D$) found in \cite{Bahri2024Explaining} using asymptotic properties of the hypergeometric function obtained from the explicit solution of the self-consistent equations of \cite{Canatar2021} (namely Eqs \ref{eq:barf},\ref{Eq:CanatarEffectiveRidge}).
\begin{figure}
    \centering
    \includegraphics[trim=1.8cm 0 5cm 6cm, clip, width=1\textwidth]{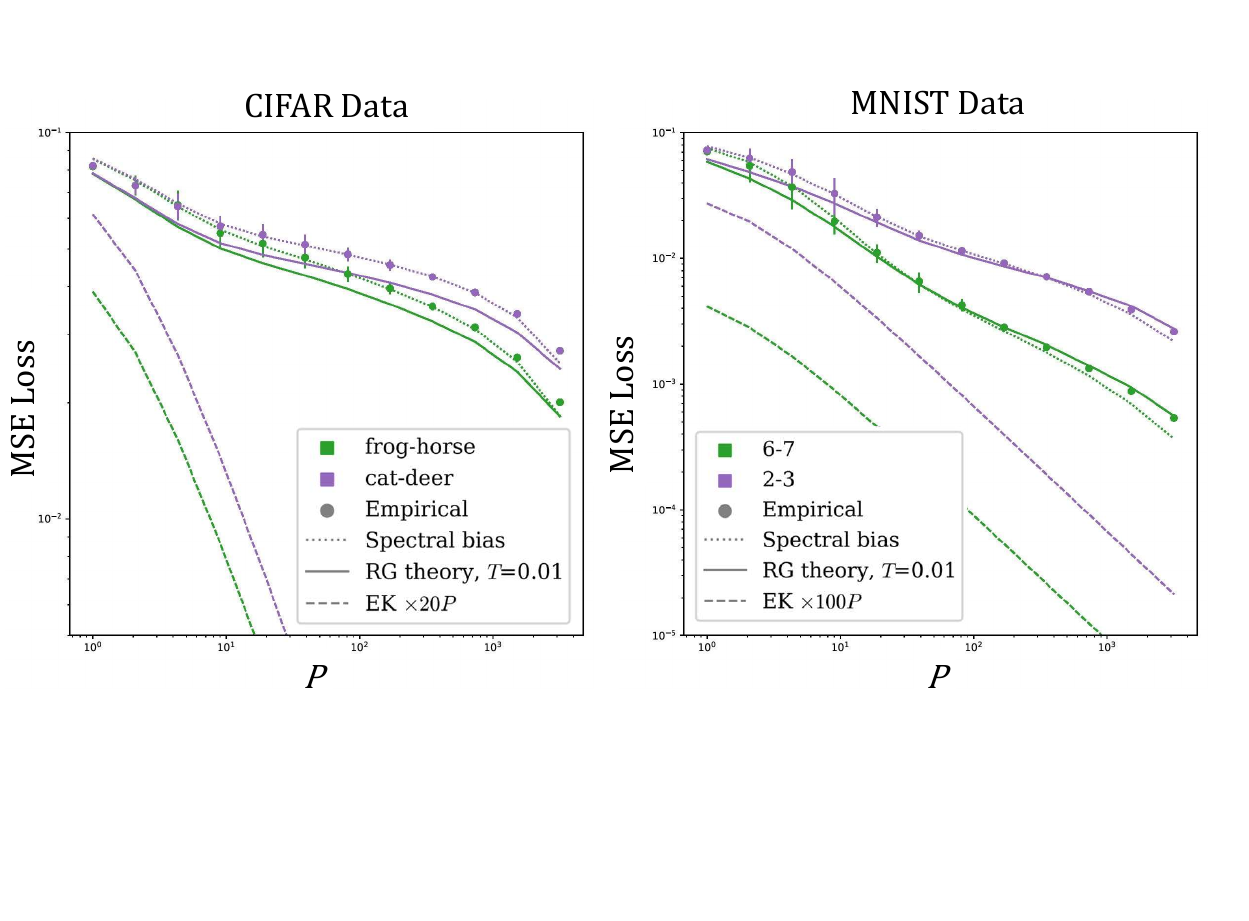}
    \vspace{-2cm}
    \caption{Gaussian Processes Regression on four 10k binary CIFAR and MNIST datasets, at $\kappa^2=1e-8$. Experimental results (dots) match well both the effective ridge theory and the RG theory. In the latter, we took $0.01$-learnability as marking the RG cut-off. We comment that results are similarly accurate for $T=0.001$ and $T=0.1$. The Equivalent Kernel estimator is expected to become accurate when the loss reaches the scale of $\kappa^2$, explaining its poor performance in the shown range of $P$. }
    \label{fig:RG_CIFAR_MNIST}
\end{figure}

Next we return to verify the closeness of the system to the EK limit. We highlight here that the EK approximation comes from the Taylor expansion of $\int d\mu_x e^{-(f(x)-y(x))^2/2\kappa^2}$. Therefore, we need to compare $\kappa_{\text{eff}}^2$ with the mean MSE loss that we have obtained. The relation $\kappa_{\text{eff}}^2=\alpha^{-1} k_T^{-\alpha}$ places $\kappa^2_{\text{eff}}$ at the same scale as the loss thus placing the system at the threshold of the EK approximation, where it is expected to provide a correct order of magnitude estimate. 

This exposes direct relations between power-laws in the data (the exponent $\alpha$) and power-laws in the learning curve via spectral bias. Thus, at least at a qualitative level, reduces the mystery of scaling-laws in DNN performance to that of scaling-laws in data. It further exemplifies a level of universality since, all the information required to model the complex and unknown target function and data measure is one parameter, namely $\alpha$. It also suggests that a finer understanding of scaling laws (e.g., addressing the differences found empirically between $\alpha_N$ and $\alpha_P$ which the GPR approach fails to capture) might be within our analytical grasp.

%% file: Chapters/Bayesian_Neural_Networks_in_the_Feature_Learning_Regime.tex
\chapter{Bayesian Neural Networks in the Feature Learning Regime}
While providing qualitative lessons on deep learning, the infinite over-parametrization or GP limit has several shortcomings. The first, concerns feature learning, a loosely defined term reflecting changes in the internal representations generated by a neural network. There is much evidence for feature learning in realistic networks which, as shown below, is absent in the infinite width limit. Perhaps the most convincing evidence for feature learning is the success of transfer learning, a technique in which one takes several trained layers of DNN, installs new layers on top of these, and trains these new layers on a related but different task. This means that the first, pre-trained, layers generated some useful representation. 

The GP/infinite-width limit is often a less powerful inference model. For instance, finitely over-parametrized CNNs which are the only set of competitive models in which a careful comparison with GP was made, shows that GPs underperform \citep{novak2018bayesian,lee2020finite} by around 5\% depending on whether or not one uses pooling. Furthermore, one can analytically establish the superiority of finite networks over GPs in various shallow network settings \citep{arous2021online,Maillard2020,Luca2024}. Finally, as we will discuss, computing perturbative corrections to the GP limit reveals that these are not controlled by one over the width/channels, but rather by that number times an extensive scale proportional to the amount of data \citep{seroussi2023separation,Hanin2023}. Thus, the GP limit is an uncontrolled approximation for most realistic scenarios. 

In this chapter, we consider the theory of feature learning, focusing on the case of fully-trained/equilibrated DNNs trained using Langevin dynamics. The equilibrium/Bayesian setting contains much of the richness of finite networks but simplifies things substantially, as we do not have to worry about transient non-equilibrium effects. We would first derive an interacting field theory governing the network outputs and preactivations. We would show how the GP limit is obtained from this theory at infinite width/channels, and why there can be no representation learning in this limit. We then consider three ways of dealing with interactions: Perturbation theory, Kernel Scaling, and Kernel Adaptation. The first approach provides information on the scaling required to be in the GP limit, as well as educated guesses for another relevant scaling regime - the mean-field scaling regime. The second provides an accurate quantitative account for the average predictor in a learning regime in which notable feature learning effects occur, however, the DNN is still in the same sample complexity class as GPR. Finally, the kernel-adaptation approach, accurate in mean-field scaling and semi-accurate in standard scaling, further tracks the internal representations directly and allows exploring regimes in which the DNN is in a different complexity class than GPR. 

\section{Multilayer actions}
Here we derive an explicit field theory description of finite DNNs. In the field-theory spirit of this review, our goal is to integrate or marginalize over all weights and obtain an equivalent description of the partition function in terms of fields only. Specifically, the field governing network output ($f(x)$) and those governing pre-activations per layer per neuron ($h^{(l)}_c(x)$). 

A basic tool we would use to derive this action are functional-delta-functions, commonly used in field theory \citep{MoritzBook} along with their Fourier representation. A functional-delta-function, $\delta[f]$, is the path-integral analog of standard delta functions. Formally it obeys the following relation w.r.t. a $Df$ functional integration  
\begin{align}
\int Df \delta[f-g] S[f] &= cS[g] 
\end{align}
where $S[f]$ is any functional. 

Similarly to a usual delta function, $\delta[..]$ can be written in Fourier form as followed
\begin{align}
\delta[f-g] &= \int D\tilde{f} e^{i \int d\mu_x \tilde{f}(x)(f(x)-g(x))}
\end{align}
where it can be verified that for any discretization of the path integral into multidimensional integrals as done in Sec. \ref{Sec:PathIntegrals}, it becomes a multidimensional delta function \footnote{albeit with different $c$ constant which we, as said, justly ignore.}. We refer to the field $\tilde{f}(x)$ as $f(x)$'s helping/auxiliary field.  

Next, we consider the partition function for the Boltzmann distribution (or Bayesian posterior) and apply the above as needed to obtain an action in the weights, which we may then integrate out. The formalism is general; however, let us introduce it for a three-layer FCN network for clarity. Similarly to Sec. \ref{Sec:NNGP_On_Data}, we consider a Langevin-trained network with weight decay and unnormalized-MSE loss which leads to a Boltzmann distribution as $t \rightarrow \infty$. The partition function, or equivalently the normalization of this distribution, is then     
\begin{align}
Z &= \int \dif {a} \;\dif{w^{(1)}} \;\dif{w^{(0)}} \; e^{-\frac{1}{2}\left[N|{a}|^2+N|{w^{(1)}}|^2+d |{w^{(0)}}|^2\right]- \frac{1}{T}\sum_{\mu=1}^P L(z_{{\theta}}(x_{\mu}),y_{\mu})}\\ \nonumber 
z_{\theta}(x) &= \sum_{i,j=1}^N \sum_{k=1}^d a_i \sigma\left(w^{(1)}_{ij} \sigma(w^{(0)}_{jk} x_k)\right)
\end{align}
 where $L(a,b)$ can be any point-wise loss (e.g. $(a-b)^2$). We would also introduce the shorthand notation $|{\theta}|^2$ capturing the first three terms in the above action. We further choose above the variance of individual weights ($a_i,w^{(1)}_i,w^{(0)}_i$) to be one over the fan-in without the data term. Taking a Bayesian perspective, this implies that in the prior all preactivations are order $1$ as they are for a network initialized by the  Kaiming/He initialization.

Evaluating averages under the above action is difficult because of the many weight variables appearing non-linearly in the loss via $z_{ {\theta}}(x)$. To circumvent this we introduce a field governing the network outputs by inserting $\int Df \delta[f-z_{{\theta}}]=1$ and going to the Fourier representation of the delta functional. Namely, 
\begin{align}
Z &= \int \dif {a} \; \dif {w^{(1)}} \; \dif{w^{(0)}} \int Df \; D\tilde{f} \; e^{-\frac{|{\theta}|^2}{2}- \sum_{\mu} L(f(x_{\mu}),y_{\mu})+i \int d\mu_x \tilde{f}(x) (f(x)-z_{{\theta}(x)})}
\end{align}
At this point the readout layer weights (${a}$) can already be integrated out however the others are still non-Gaussian. To remedy this, we proceed by introducing the pre-activations of the pen-ultimate layer by placing the following constant factor in the partition function  $\prod_{i=1}^N \int D h_i^{(1)}\delta[h_i^{(1)}-\sum_{j,k}w^{(1)}_{ij} \sigma(w^{(0)}_{jk}x_k)]$ using again the delta functional identity. This then yields 
\begin{align}
Z&=\int \dif{a} \; \dif {w^{(1)}} \; \dif{w^{(0)}} \; Df \; D\tilde{f} \; D{h}^{(1)}D{\tilde{h}}^{(1)} \; e^{-S} \\ \nonumber 
S &= \frac{|{\theta}|^2}{2}+\frac{1}{T}\sum_{\mu=1}^P L(f(x_{\mu}),y_{\mu})-i \int \dif\mu_x \tilde{f}(x) \Big(f(x)-\sum_i a_i \sigma(h^{(1)}_i(x))\Big) \\ \nonumber 
&- i\sum_i \int \dif \mu_x \tilde{h}^{(1)}_i \Big(h^{(1)}_i(x) - \sum_{jk} w^{(1)}_{ij} \sigma(w^{(0)}_{jk }x_k)\Big)
\end{align}
at this point both the ${a}$ and ${w}^{(1)}$ integrations are Gaussian, but ${w}^{(0)}$ is still non-Gaussian. We may again negotiate this difficulty by placing a factor involving the $h^{(0)}_j(x)=\sum_{k}w^{(0)}_{jk}x_k$ however since, unlike hidden pre-activations, this is just a simple linear transformation on the input weights, it is of little practical use. At this point, we may integrate out only the ${a}$ and ${w}^{(1)}$ weights and obtain our field theory description of a 3-layer FCN (before data averaging) 
\begin{align}
\label{Eq:FCN3}
Z &= \int \int \dif {{w}^{(0)}}Df \; D\tilde{f} \; D{h}^{(1)} \; D{\tilde{h}}^{(1)} \; e^{-S_{\text{FCN3}}} \\ \nonumber 
S_{\text{FCN3}} &= \frac{d |{ w^{(0)}}|^2}{2}+\sum_{\mu} L(f(x_{\mu}),y_{\mu}) \\ \nonumber
&- i\int \dif\mu_x \Big[\tilde{f}(x) f(x)+\sum_i
\tilde{h}^{(1)}_i(x) h^{(1)}_i(x)\Big] \\ \nonumber 
&+\frac{1}{2N^{(1)}} \sum_i \left(\int \dif\mu_x \tilde{f}(x) \sigma(h^{(1)}_i(x))\right)^2\\ \nonumber
&+ \frac{1}{2N^{(0)}}\sum_{ij}\left( \int \dif \mu_x \tilde{h}^{(1)}_i(x) \sigma({w}^{(0)}_{j} \cdot {x})\right)^2
\end{align}
where we also allowed layer-dependent width ($N^{(l)}$). 

Next, we point out several properties of this action, starting from the {\bf fluctuating kernels' interpretation.} Indeed, one can readily integrate all the helping fields that appear in quadratic form by completing the squares. For instance, doing so on $\tilde{f}$, would yield a non-linearity of the form $\frac{1}{2}\int \dif\mu_x \dif\mu_{x'} f(x) f(x') \tilde{K}^{-1}(x,x')$ where $\tilde{K}$, the fluctuating kernel, is given by $N^{-1}\sum_{i=1}^N \sigma(h^{(1)}_i(x))\sigma(h^{(1)}_i(x'))$. The average of this latter quantity, at $N\rightarrow \infty$, was the kernel defined in Sec. \ref{Sec:InfiniteRandomDNNs}. Thus, finite-width effects are like having fluctuating, upstream weight-dependent kernels controlling the outputs. Notwithstanding, for subsequent approximations having the helping fields, and avoiding inversion of fluctuating kernels such as $\tilde{K}$, would prove useful. 

Next, we recover the {\bf GPR limit} at large $N^{(l)}$. The cleanest argument involves taking the limit sequentially downstream: Consider integrating out $w^{(0)}$ at very large $N^{(0)}$ via a cumulant-like expansion ($\log(\int \dif x P(x) e^{O(x)})=\langle O(x) \rangle_{P(x)}+\frac{1}{2} \text{Var}[O(x)]+.. $) with $x$ being all $w^{(0)}$'s, $O(x)$ the $w^{(0)}$-dependent random-variable defined by the last term in $S_{\text{FCN3}}$, $P(x)$ being the probability measure defined by the first term in the action. Discarding constants in partition functions, as usual, one obtains the action $S_{\text{FCN3}}$ with all $w^{(0)}$ dependence removed and additional terms of the form 
\begin{align}
&\left \langle \frac{1}{2N^{(0)}}\sum_{ij}\left( \int \dif \mu_x \tilde{h}^{(1)}_i(x) \sigma({w}^{(0)}_{j} \cdot {x})\right)^2 \right \rangle_{{\cal N}(0,d^{-1}I_{d\times d} \otimes I_{N^{(0)}\times N^{(0)}};{w}^{(0)})}  \\ \nonumber 
&+\frac{1}{2}Var\left[ \frac{1}{2N^{(0)}}\sum_{ij}\left( \int d\mu_x \tilde{h}^{(1)}_i(x) \sigma(\bm{w}^{(0)}_{j} \cdot \bm{x})\right)^2 \right]_{{\cal N}(0,d^{-1}I_{d\times d} \otimes I_{N^{(0)}\times N^{(0)}};\bm{w}^{(0)})}+... \\ \nonumber 
&= \frac{1}{2}\sum_{i=1}^{N^{(1)}} \left \langle \left( \int \dif \mu_x \tilde{h}^{(1)}_i(x) \sigma({w}^{(0)}_{j=1} \cdot {x})\right)^2 \right \rangle_{{\cal N}(0,d^{-1} I_{d\times d} \otimes I_{N^{(0)}\times N^{(0)}};{w}^{(0)})}  \\ \nonumber 
&+\frac{1}{2 N^{(0)}} Var\left[\sum_{i=1}^{N^{(1)}}\left( \int d\mu_x \tilde{h}^{(1)}_i(x) \sigma(\bm{w}^{(0)}_{j=1} \cdot {x})\right)^2 \right]_{{\cal N}(0,d^{-1}I_{d\times d} \otimes I_{N^{(0)}\times N^{(0)}};{w}^{(0)})}+...
\end{align}
where we used the independence of different ${w}^{(0)}_j$ under the Gaussian measure used in the cumulant expansion. At $N^{(0)}\rightarrow \infty$, only the first term above survives, and yields $\frac{1}{2}\sum_i \int \dif\mu_x \dif\mu_{x'} \tilde{h}_i^{(1)}(x) K^{(1)}(x,x') \tilde{h}_i^{(1)}(x')$, where $K^{(1)}(x,x')$ is the latent kernel of the input layer (see  Eq. \ref{Eq:NNGP_K_1}) which also coincides with above average over of $w^{(0)}_{j=1}$. As $\tilde{h}^{(1)}$ now appears only to quadratic order in the action, one can integrate it using square completion, and obtain $S_{\text{FCN3}}$ at $N^{(0)}\rightarrow \infty$ given by  
\begin{align}
S_{\text{FCN3},N^{(0)}\rightarrow \infty} &= \sum_{\mu} L(f(x_{\mu}),y_{\mu}) - i\int \dif\mu_x \Big[\tilde{f}(x) f(x)+\sum_i
\tilde{h}^{(1)}_i(x) h^{(1)}_i(x)\Big] \\ \nonumber 
&+\frac{1}{2N^{(1)}} \sum_i \left(\int \dif\mu_x \tilde{f}(x) \sigma(h^{(1)}_i(x))\right)^2 \\ \nonumber &+ \frac{1}{2}\sum_i \int \dif \mu_x \dif \mu_{x'} h_i^{(1)}(x) [K^{(1)}]^{-1}(x,x') h_i^{(1)}(x')
\end{align}
with all $\tilde{h}^{(1)},w^{(0)}$-dependent terms removed. The process can then be repeated, namely, perform a cumulant expansion in the second to last term of the above action under the Gaussian measure induced by the new kernel term (last term above) and integrating out $\tilde{f}$ to get a kernel term for $f(x)$. This step would recover $S_{\GP}$ (Eq. \ref{Eq:Z_FieldTheoryOnData}) as the action for $f$. We note by passing that, under the additional approximations used below, one can also establish this result using the weaker requirement of $N^{(l)}=N \rightarrow \infty$. 

The above derivation of the GP limit makes precise the notion that it is indeed a {\bf lazy learning} \citep{chizat2018lazy} limit --- one without any useful feature learning. Specifically, considering the first step detailed above, we find that the effect of $w^{(0)}$ on downstream variables ($h^{(1)},f$) is completely dataset agnostic as $N^{(0)}\rightarrow \infty$. Thus, if some dataset-dependent information gets encoded in the input weights, it has no consequence on the downstream processing of the network or predictions. The same then holds for the subsequent layer.   

\subsection{Dataset averaging}
Here we consider a necessary step towards solving any such theory analytically namely {\it dataset averaging}. Since all the more complex terms involving interactions between layers are dataset-agnostic \footnote{As discussed, around Eq. \ref{Eq:RKHS}, despite appearances the action is largely independent of $d\mu_x$.}, this amounts to repeating the manipulations of Sec. \ref{Sec:AveragedGPR} with the RKHS term replaced with these more complex terms. The result is the following replica field theory with $Q$ copies of each field and variable whose action is  
\begin{align}
\label{Eq:BarSFCN3}
&\bar{S}_{\text{FCN3}}= \sum_{a,j}\frac{d |{ w_{a,j}^{(0)}}|^2}{2} - P\int \dif \mu_x e^{-\sum_{a=1}^Q L(f_a(x),y(x))} \\ \nonumber &- \sum_{a=1}^Q i\int d\mu_x [\tilde{f}_a(x) f_a(x)+\sum_i
\tilde{h}^{(1)}_{ai}(x) h^{(1)}_{ai}(x)] \\ \nonumber 
&+\frac{1}{2N} \sum_{i,a} \left(\int \dif\mu_x \tilde{f}_a(x) \sigma(h^{(1)}_{ai}(x))\right)^2 + \frac{1}{2N}\sum_{ij}\left( \int \dif \mu_x \tilde{h}^{(1)}_{ai}(x) \sigma({w}^{(0)}_{a,j} \cdot {x})\right)^2
\end{align}
on which one can readily apply the approximations carried out in the second chapter. In this chapter, our focus is on finite-width effects. We thus consider the simplest approximation for the finite-sample effects, namely the EK approximation, and further, take the first order in Taylor expansion of the MSE loss term $- P\int \dif \mu_x e^{-\sum_{a=1}^Q L(f_a(x),y(x))} \approx - P + P\int \dif \mu_x \sum_{a=1}^Q L(f_a(x),y(x))$, which can, alternatively to the derivation in Sec. \ref{Sec:AveragedGPR} using Poisson averages, be regarded as the lowest order cumulant expansion of fluctuations of the loss. Accordingly, we obtain
\begin{align}
&\bar{S}_{\text{FCN3,EK}}= \sum_{a,j}\frac{d |{ w_{a,j}^{(0)}}|^2}{2}+P\int \dif\mu_x \frac{(f_a(x) - y(x))^2}{2\kappa^2}\\ \nonumber &- \sum_{a=1}^Q i\int \dif \mu_x [\tilde{f}_a(x) f_a(x)+\sum_i
\tilde{h}^{(1)}_{ai}(x) h^{(1)}_{ai}(x)] \\ \nonumber 
&+\frac{1}{2N} \sum_{i,a} \left(\int \dif \mu_x \tilde{f}_a(x) \sigma(h^{(1)}_{ai}(x))\right)^2 + \frac{1}{2N}\sum_{ij}\left( \int \dif \mu_x \tilde{h}^{(1)}_{ai}(x) \sigma({w}^{(0)}_{a,j} \cdot {x})\right)^2
\end{align}
To illustrate the power of our method, in the following, we will do concrete computations for a simpler case of 2-layer FCN action given, following similar manipulation to the 3-layer case, by 
\begin{align}
\label{Eq:S_FCN2_nonEK}
&\bar{S}_{\text{FCN2}}= -P\int \dif\mu_x e^{-\frac{(f_a(x)-y(x))^2}{2\kappa^2}} - \sum_{a=1}^Q i\int \dif \mu_x \tilde{f}_a(x) f_a(x) \\ \nonumber 
&+\frac{1}{2N} \sum_{i,a} \left(\int \dif\mu_x \tilde{f}_a(x) \sigma({w}^{(0)}_{a,i} \cdot {x})\right)^2 + \frac{d |{ w_{a,i}^{(0)}}|^2}{2}
\end{align}
which in the EK limit gets replaced by  
\begin{align}
\label{Eq:S_FCN2}
&\bar{S}_{\text{FCN2,EK}}= P\int \dif\mu_x \frac{(f_a(x)-y(x))^2}{2\kappa^2} - \sum_{a=1}^Q i\int \dif \mu_x \tilde{f}_a(x) f_a(x) \\ \nonumber 
&+\frac{1}{2N} \sum_{i,a} \left(\int \dif\mu_x \tilde{f}_a(x) \sigma({w}^{(0)}_{a,i} \cdot {x})\right)^2 + \frac{d |{ w_{a,i}^{(0)}}|^2}{2}
\end{align}

\section{Approximation Scheme: Perturbation theory}
We next discuss several approximate, yet asymptotically exact, ways of dealing with the interaction terms in the above action.  

Given that our action becomes Gaussian at $N\rightarrow \infty$, it is natural to consider it as the starting point of a $1/N$ perturbative expansion. Such an expansion can shed further light on the required conditions to be close to the GP limit. We would find that these conditions are rarely met in practice since each term in the series involves extensive summations over $P$, making the expansion poorly convergent unless $P \ll  N$ (as typical values for $N$ are $O(10^3)$ for FCN and $O(10^2)$ for CNNs). Interestingly, though, leading order perturbation theory appears to provide several insights about the correct scaling of hyperparameters (e.g. $\mu$P scaling as proposed by \cite{yang2022tensorprogramsvtuning} and also \cite{dinan2023effectivetheorytransformersinitialization}). This effectiveness is yet to be explained theoretically but seems to imply some correlation between $P \ll N$ and $P \gg N$ behavior of realistic neural networks.  

We turn to the technicalities of such an expansion focusing, for simplicity, on the case of $\bar{S}_{\text{FCN2,EK}}$ then sketch the generalization for deeper networks. Following the previous derivation of the GP limit, we may write our action as 
\begin{align}
&\bar{S}_{\text{FCN2,EK}}= \bar{S}_{0} + U \\ \nonumber 
&\bar{S}_{0} =  \frac{d|{w^{(0)}}|^2}{2}+\sum^Q_{a=1} P\int \dif \mu_x \frac{(f_a(x)-y(x))^2}{2\kappa^2} -  i\int \dif\mu_x \tilde{f}_a(x) f_a(x) \\ \nonumber &+\sum_{a=1}^Q \frac{1}{2} \int \dif\mu_x \dif\mu_{x'} \tilde{f}_a(x) K^{(1)}(x,x') \tilde{f}_a(x') \\ \nonumber 
&U = \bar{S}_{\text{FCN2,EK}}-\bar{S}_0 \\ \nonumber 
= & \frac{1}{2N} \sum_{a} \int \dif\mu_x \dif\mu_{x'} \tilde{f}_a(x) \sum_{i=1}^N\left[\sigma({w}^{(0)}_{a,i} \cdot {x})\sigma({w}^{(0)}_{a,i} \cdot {x'})-K^{(1)}(x,x')\right]\tilde{f}_a(x')
\end{align}
 
A first simplification, which is a natural outcome of working in the EK limit, is that all replica modes are decoupled. Hence, the replica limit is trivial here (see Sec. \ref{Sec:Replicas}) and we may remove replica indices and treat the action as a regular field theory. 

As the interaction involves the helping field, we require its average and correlations under the Gaussian theory $\bar{S}_{0}$. A simple computation yields 
\begin{align}
\label{Eq:tilde_f_averages_EK}
&\langle \tilde{f}(x) \rangle_{\bar{S}_0} = i\frac{P}{\kappa^2} \langle f(x)-y(x) \rangle_{\bar{S}_0} \equiv  i\frac{P}{\kappa^2} \langle f(x)-y(x) \rangle_{\GPR,\EK} \\ \nonumber
&\langle \tilde{f}(x)\tilde{f}(x')\rangle_{\bar{S}_0,\con} = [K^{(1)}+(\kappa^2/P) \hat{I}]^{-1}(x,x') \equiv \tilde{K}^{-1}(x,x') 
\end{align}
where $\hat{I}$ is the identity operator on function space. We find that the average $\tilde{f}(x)$ is proportional to the discrepancy in GPR prediction using the kernel $K^{(1)}(x,x')$ within the EK approximation \footnote{We note by passing that had we not done dataset averaging, this average would have been  $i\kappa^{-2} 
\sum_{\mu} \delta(x-x_{\mu}) \langle f(x_{\mu})-y(x)\rangle_{\GPR}$. This allows one to recover the fixed dataset case from our average dataset computations}. Furthermore, by adding a source term that couples $\tilde{f}$ to the action and taking $\log$-derivatives w.r.t to the source to obtain averages, one can show that the average of $\tilde{f}$ is proportional to the discrepancy of the average predictor also under the full interacting theory. 

As would soon be apparent, the leading order correction in $1/N$ is obtained from a 2nd order perturbative expansion in $U$. Following Sec. \ref{Sec:PT}, this is given by 
\begin{align}
\langle f(x_*) - y(x_*) &\rangle_{\bar{S}_{\text{FCN2,EK}}} = \langle f(x_*) - y(x_*)\rangle_{\text{GPR,EK}} \\ \nonumber &-i\frac{\kappa^2}{P} \langle \tilde{f}(x_*) U \rangle_{\bar{S}_0,\con}-i\frac{\kappa^2}{P}\frac{1}{2}\langle \tilde{f}(x_*) U^2 \rangle_{\bar{S}_0,\con} 
\end{align}
The term linear in $U$ vanishes due to the independent average of the Gaussian input weight under which $\langle \sigma({w}^{(0)}_{a,i} \cdot {x})\sigma({w}^{(0)}_{a,i} \cdot {x'})\rangle_{\bar{S}_0}=K^{(1)}(x,x')$. 
The second term involves the variance of $\sigma({w}^{(0)}_{a,i} \cdot {x})\sigma({w}^{(0)}_{a,i} \cdot {x'})$ which is non-vanishing. Taking this average and summing over the width index, one obtains  
\begin{align}
&\frac{1}{2}\langle \tilde{f}(x_*) U^2 \rangle_{\bar{S}_0,\con} = \int \dif\mu_{x_1} \dif\mu_{x_2} \dif\mu_{x_3}\dif\mu_{x_4} \\ \nonumber & \frac{1}{8N}\left\langle \tilde{f}(x_*)   \tilde{f}_a(x_1) \tilde{f}_a(x_2) \tilde{f}_a(x_3) \tilde{f}_a(x_4)\right\rangle_{\bar{S}_0,\con} \\ \nonumber 
& \times\left[\langle \sigma({w} \cdot {x_1}) \sigma({w} \cdot {x_2})\sigma({w} \cdot {x_3}) \sigma({w} \cdot {x_4}) \rangle_{w} -K^{(1)}(x_1,x_2)K^{(1)}(x_3,x_4)  \right]  \\ \nonumber 
=& \int d\mu_{x_1} \dif\mu_{x_2} \dif\mu_{x_3}\dif\mu_{x_4} \frac{1}{24}\left\langle \tilde{f}(x_*)   \tilde{f}_a(x_1) \tilde{f}_a(x_2) \tilde{f}_a(x_3) \tilde{f}_a(x_4)\right\rangle_{\bar{S}_0,\con} \\ \nonumber 
& \times\left[3\langle \sigma({w} \cdot {x_1}) \sigma({w} \cdot {x_2})\sigma({w} \cdot {x_3}) \sigma({w} \cdot {x_4}) \rangle_{w} -K^{(1)}(x_1,x_2)K^{(1)}(x_3,x_4)[3]  \right]/N
\end{align}
where $\langle ... \rangle_w = \langle ... \rangle_{{\cal N}(0,d^{-1}I_{d\times d};w)}$ and we used to interchangeability of $x_1,x_2,x_3,x_4$ in the connected average to symmetrize the variance term. To this end we introduce the pair-symmetrizer notation  $g(x_1,x_2,x_3,x_4)[3]=g(x_1,x_2,x_3,x_4)+g(x_1,x_3,x_3,x_4)+g(x_1,x_4,x_2,x_3)$. 

We next relate the factor appearing in the bottom line above, to the fourth cumulant associated with the output of a random network (For a more general discussion of this structure and its relation with the Edgeworth expansion, see Ref. \cite{naveh2021predicting}). Since the random network output is $z_{\theta}(x)=\sum_{i=1}^N a_i \sigma({w}_i \cdot x)$ and made of $N$ independent variables, the cumulants are just $N$ times those of any specific $i$. The fourth cumulant is therefore  
\begin{align}
&u(x_1\ldots x_4) \equiv \\ \nonumber
&N \langle a \sigma({w} \cdot {x}_1) a \sigma({w} \cdot {x}_2) a \sigma({w} \cdot {x}_3) a \sigma({w} \cdot x_4) \rangle_{P_0(a,{w})} - N^{-1} K(x_1,x_2)K(x_3,x_4)[3] \\ \nonumber 
&= \left[3\langle  \sigma({w} \cdot {x}_1)  \sigma({w} \cdot {x}_2)  \sigma({w} \cdot {x}_3)  \sigma({w} \cdot {x}_4) \rangle_{w} - K(x_1,x_2)K(x_3,x_4)[3]\right]/N
\end{align}
where $P_0(a,{w}) =  {\cal N}(0,N^{-1};a),{\cal N}(0,d^{-1} I_{d\times d};w)$ leading to a $1/N$ scaling of all the above terms. The above expression generalizes to any neural network, even deeper FCNs or CNNs, by simply replacing the last line above with the fourth cumulant of the desired random network.

Next, we resolve the above connected average over the $\tilde{f}$'s to obtain 
\begin{align}
&\nonumber -i\left\langle \tilde{f}(x_*)   \tilde{f}_a(x_1) \tilde{f}_a(x_2) \tilde{f}_a(x_3) \tilde{f}_a(x_4)\right\rangle_{\bar{S}_0,\con}  \\  
&=  -\tilde{K}^{-1}_{x_*,x_1} \frac{P^3\Delta(x_2)\Delta(x_3)\Delta(x_4)}{\kappa^{6}} \otimes \left[\sum_{i=1}^4 \{x_1 \leftrightarrow x_i\} \right] \\ \nonumber
&+\tilde{K}^{-1}_{x_*,x_1}\left[\frac{P\Delta(x_2)}{\kappa^2}\tilde{K}^{-1}_{x_3,x_4}\otimes[1+\{x_2 \leftrightarrow x_3\}+\{x_2 \leftrightarrow x_4\}]\right] \\ \nonumber 
&\otimes \left[\sum_{i=1}^4 \{x_1 \leftrightarrow x_i\}\right]
\end{align}
where $\Delta(x_1)=\langle f(x_1) - y(x_1)\rangle_{\GPR,\EK}$  and we have further introduced the variables swap notation ($\otimes [\{x_i \leftrightarrow x_j\}]$) implying for instance $A(x_1,x_2,x_3) \otimes [1+\{x_1 \leftrightarrow x_2 \}+\{x_1 \leftrightarrow x_3\}]=A(x_1,x_2,x_3)+A(x_2,x_1,x_3)+A(x_3,x_2,x_1)$.

Next, we require a concrete expression for $u(x_1 \ldots x_4)$ to complete the computation. To this end, we first note that for FCNs, due to the rotation symmetry of the kernel, it must obey the ansatz 
\begin{align}
u(x_1 \ldots x_4)\propto u(|x_1|\ldots |x_4|,x_1 \cdot x_2,x_1 \cdot x_3,\ldots ,x_3 \cdot x_4)
\end{align}
Next we make the reasonable assumption that the dimension of $x_i$'s together with the data measure implies that typically $|x_i|^2 = O(1)$ and $x_i \cdot x_{j\neq i}=O(1/\sqrt{d_{\eff}})$, with $d_{\eff}\gg 1$. Further,  taking, for simplicity, $|x_i|^2=1$, and an antisymmetric network ($z_{\theta}(x)=-z_{\theta}(-x)$) we obtain the following series expansion for $u(x_1...x_4)$
\begin{align}
u(x_1 \ldots x_4) = u_1[(x_1 \cdot x_2) (x_3 \cdot x_4)][3]/N+O(\frac{1}{d_{\eff}^2 N}) 
\end{align}
where we kept the leading $O(1/d_{\eff})$ contribution and ignored higher order anti-symmetric (e.g. $(x_1 \cdot x_2)^3(x_3 \cdot x_4)$) terms \footnote{A more careful approximation is to model those quartic and higher terms as white noise \cite{Cui2023} when sampling from $d\mu_x$, however in our context this would not affect the average predictor.}. Underlying this expansion is the assumption that $u(...)$ has $O(1)$ Taylor coefficient, such as $u_1$ above, when viewed as a function of $x_i \cdot x_j$. This is consistent with the fact that, when normalizing properly, $u(x,x,x,x)$ being a 4th cumulant of the network outputs, is order $1$. Hence, it would require delicate cancellations if its Taylor coefficients were $O(d_{eff})$. We further note by passing that the above becomes exact for deep linear networks, which show many of the phenomena associated with generic deep linear networks (e.g. their training dynamics is non-linear \citep{saxe2013exact}, they exhibit feature learning \citep{LiSompolinsky2021,seroussi2023separation}, some linear CNNs can show sample complexity separation from GPR \citep{naveh2021self}).  Indeed the above argument, also related to the Gaussian equivalence principle \citep{Cui2023}, can be seen as the source of this effective linear behaviour.  

Let us focus on the most interesting $1/N$ correction involving three $\Delta$ factors or equivalently three $y$ factors. Whereas the linear in $y$ terms can still be absorbed into some task/target agnostic complex redefinition of the GPR kernel, the cubic term represents some initial form of feature learning or interaction between features. Gathering all relevant contributions, it amounts to the following correction to the average discrepancy (or predictor) 
\begin{align}
\label{Eq:1stOrderCorrection}
&\frac{u_1 P^2}{24 N\kappa^4}\int \dif \mu_{x_1}... \dif\mu_{x_4} 4 \tilde{K}^{-1}_{x_*,x_1} \Delta(x_2)\Delta(x_3)\Delta(x_4)[(x_1 \cdot x_2) (x_3 \cdot x_4)[3]] = \\ \nonumber 
& \frac{3u_1P^2}{6 N\kappa^{4}}\int  \dif\mu_{x_1} \dif\mu_{x_2} \tilde{K}^{-1}_{x_*,x_1} (x_1 \cdot x_2)\Delta(x_2) \int  \dif\mu_{x_3}  \dif\mu_{x_4} \Delta(x_3) (x_3 \cdot x_4) \Delta(x_4)
\end{align}
where the factor of $4$ came from the 4 equivalent exchanges implied by $\otimes [\sum_{i=1}^4 \{ x_1 \leftrightarrow x_i\}]$.

Analyzing the above term, we first note that, despite appearances, it is non-divergent as $\kappa^2 \rightarrow 0$ since it follows from taking Eq. (\ref{Eq:EK_Av_Pred}) back to real space and subtracting $y$ that  $P\kappa^{-2} \Delta(x) =\int  \dif\mu_{x'} \tilde{K}^{-1}(x,x')y(x')$. In fact, it vanishes, which is, however, an artefact of working with the EK approximation beyond its validity regime. 

More troubling, for our perturbative approach, is the appearance of $P^2/N$ and the fact that other $P$ dependencies can only appear through the discrepancy in GPR prediction. Assuming $\Delta(x)$ falls off as $P^{-\alpha_D/2}$, we require $\alpha_D > 4/3$ for this correction to be small compared to $f(x_*)$ (which we assume is $O(1)$). Reference \cite{Bahri2024Explaining}, studied $\alpha_D$ in various settings and found $\alpha_D\approx 1$ in variance-limited regimes and $\alpha_D \in [0.26,0.58]$ in resolution-limited regimes. Both imply that just scaling up $P$, keeping $N$ fixed, would place us outside the perturbative regime. More advanced scaling can be considered. A reasonable choice is to scale instead the number of parameters and $P$ together which implies, for deep networks, $N \propto \sqrt{P}$. For $\alpha_D=1$, the perturbative correction would not scale. However, for $\alpha_D<1$, it would still diverge as we scale up the model.


A more precise analytical statement can be carried out for deep linear networks, in which case, given our normalization here that $|x|=1$ and not $d$ as before, $K(x,x')=x \cdot x'$. Noting in addition that $\frac{P}{\kappa^2}\Delta(x)=[[K+\kappa^2 I/P]^{-1}y](x)$ we find the following simplified expression \footnote{The correct interpretation for operator inverses in this rank-deficient case is to use pseudo-inverses, namely inverses which project out the null-space of the operator.}
\begin{align}
\frac{3u_1}{6 N}\Delta(x_*) \sum_k y_k^2 \frac{\lambda_k}{(\lambda_k+\kappa^2/P)^2}
\end{align}
where $\lambda_k$ are, as always, the eigenvalues of the kernel w.r.t. the data measure, and $y_k$ is the decomposition of the target on these modes. 

Consequently, we find the following $1/N$ correction for a deep linear network 
\begin{align}
\langle f(x_*) - y(x_*) &\rangle_{\bar{S}_{\text{FCN2,EK}}} &= \Delta(x_*)\left[1+\frac{3 u_1}{6N} \sum_k y_k^2 \frac{\lambda_k}{(\lambda_k+\kappa^2/P)^2} \right]
\end{align}
To estimate the magnitude of this correction, we first note that our normalization in the EK limit implies that for a normalized target ($y(x)=O(1)$), $\sum_k y_k^2=O(1)$. To make further progress, we should make some assumptions on the spectrum. The simplest is Gaussian iid data, in which case $\lambda_k=1/d$. Here for $\kappa^2=1$ and $P$ being of the same order of $d$ (intermediate learning), we find that $P^2/(dN)\propto d/N$, in line with the heuristic estimation above. At high performance ($P \gg d$) we obtain $d/N$ as the control parameter for the relative correction in the discrepancy. However, as the discrepancy vanishes for $P \gg d$, this is not easily noticeable in practice. Similar assessments can be carried in the more realistic case where $\lambda_k = k^{-1-\alpha}$ and $y^2_k = O(k^{-1-\alpha})$, which can arise for example by considering power law covariance structure on the data and uniform discrepancies at initialization. In this case, taking again $\kappa^2=1$ for clarity, the unlearnable modes ($k > k_T$) where $k_T$ is defined by $\lambda_{k_T}P=1 \Rightarrow k_T = P^{1/(1+\alpha)}$ would dominate the relative correction leading to a $\frac{P^2}{N} \sum_{k>k_{T}} y_k^2 \lambda_k \approx \frac{P^2}{N} k^{-1-2\alpha}_{T}=\frac{P^{\frac{1}{1+\alpha}}}{N}$-factor which is again extensive in $P$. 

As a comment to field theorists, we stress that our perturbative approach, and others', \citep{roberts2021principles} are strict low-order perturbation theories. As such they do not involve partial resummations (e.g. self-energy correction, one-loop corrections, etc.), commonly used in field theory. Indeed we have no RG-based justification for keeping only a subset of corrections, although it would be interesting to develop such a reasoning \citep{howard2024}.

\section{Application: Hyperparameter transfer}
\label{Sec:Hyper}
Hyper-parameter transfer is a technique by which hyper-parameter tuning (e.g. choice of weight-decays, widths, depth, learning rates, etc.) is done by brute-force grid scans on small models (e.g. fewer parameters, smaller dataset, shorter training times). Subsequently, the resulting optimal hyper-parameters are re-scaled (transferred) appropriately to larger models.    

A recently proposed prescription for the above re-scaling is the consistent-limit prescription \citep{yang2022tensorprogramsvtuning,bordelon2023depthwise}. Here one scales hyperparameters, for example, the width, such that they do not affect the small $P$ behavior, where one may use perturbation theory. This ensures that, within perturbation theory, scaling up the width one remains with an optimal choice of hyperparameters for the small $P$ model. One then makes the working {\it assumption} that this remains a good choice for the model at larger $P$. This assumption is supported by the standard practice of tuning other hyperparameters such as learning rates, momentum, and weight decays based on early training times (and hence effectively small $ P $). However, why this optimal choice for small $P$ remains optimal for larger $P$ is non-trivial from a theoretical perspective. Still, if we adopt this viewpoint, we should scale parameters such that leading order perturbation theory remains invariant under scaling.

To this end, we consider our obtained perturbative correction to the predictor (Eq. \ref{Eq:1stOrderCorrection}) and note that it scales as $1/N$. Following said logic, we wish to change other quantities to cancel out this $1/N$ dependence. Taking cues from the deep linear case, we note that if we scale down both $\lambda_k$ and $\kappa^2$ by $1/N$ we obtain an $N$-independent result. Such a change can be obtained mechanistically, by dividing the network's readout layer weights by $\sqrt{N}$ and, viewing it as mimicking gradient noise, dividing Langevin noise variance by $1/N$, then scaling down all weight-decay terms by $1/\chi$ to maintain the variance despite the reduced noise. In the context of gradient flow and NTK dynamics, where weight-decay and noise are absent, $1/\sqrt{N}$ scaling down of the outputs is known as mean-field scaling which had some practical successes \citep{yang2022tensorprogramsvtuning,dinan2023effectivetheorytransformersinitialization}. In our Langevin training case, the re-scaling prescription is then to optimize the model's $\kappa^2$, weight-decays ($\gamma$), and $N$ at small $P,N$, then take $N,\kappa^2,\gamma \rightarrow SN,\kappa^2/S,\gamma/S$ for the larger $P$ experiments. The choice of scaling factor $S$, can be loosely chosen such that the overparmaterization ratio (number of parameters over the number of data points, which goes roughly as $N^2/P$) remains fixed (i.e. $P\rightarrow S^2 P$).

\section{Approximation Scheme: Kernel Scaling}
A shortcoming of the previous perturbative approach is that it appears uncontrolled in various relevant settings. Furthermore, it seems unlikely that one can describe qualitatively new phenomena, such as feature learning, by perturbing away from a theory with no feature learning. This motivates us to consider non-perturbative approaches. 

One such approximation scheme, yielding surprisingly simple outcomes argues that under various circumstances non-perturbative on the average predictor can be absorbed into a single scalar variable governing the scale of the kernel. This approach is based on two approximations. The first was introduced in \cite{LiSompolinsky2021} in the context of deep linear networks, and the second approximation in \cite{ariosto2022statistical} which extends the former to non-linear networks. Here we derive a variant of this approach, where we also average over the data and work within the replicated field theory formalism. We also focus on the case of two-layer networks (i.e. Eq. \ref{Eq:S_FCN2_nonEK}).

Our starting point is thus Eq. \ref{Eq:S_FCN2_nonEK} where our first aim is to integrate out the read-in layer weights (${w}^{(0)}$). As one can verify in Eq. \ref{Eq:S_FCN2}, for deep linear networks ($\sigma(x)=x$) this amounts to a simple Gaussian integration. However, when $\sigma(..)$ is non-linear, direct integration of ${w}^{(0)}$ becomes intractable. To this end, Ref. \cite{ariosto2022statistical} suggests we treat terms such as $\int \dif \mu_x \tilde{f}(x) \sigma({w}^{(0)} \cdot x)$ as a Gaussian random variable under the probability measure induced by the weight decay part on the action (${\cal N}(0,I/d;w^{(0)})$) \footnote{Focusing for simplicity on anti-symmetric activation functions it also has zero mean.}. 
This Gaussianity assumption is analogous to the central limit theorem if we view the integral over $\mu_x$ as a ``summation'' on sufficiently many weakly correlated variables under the ${\cal N}(0,I/d;w^{(0)})$ ensemble, namely the set of random variables $\tilde{f}(x) \sigma({w}^{(0)} \cdot x)$ one obtains for different $x$ points. We next see what theory this approximation leads to, then flesh out one typical scenario in which this works very well and another, with strong feature learning, where it fails. 

Focusing only on the ${w}^{(0)}$ relevant terms in Eq. \ref{Eq:S_FCN2_nonEK}, we have the following integral to perform \footnote{where we again allowed ourselves the freedom to add or remove constant factors partition functions, specifically here the normalization of the Gaussian distribution of ${w}^{(0)}$.}
\begin{align}
Z_w &\equiv \int \dif {w}^{(0)} e^{-\frac{1}{2N} \sum_{i,a} \left(\int \dif\mu_x \tilde{f}_a(x) \sigma({w}^{(0)}_{a,i} \cdot {x})\right)^2} {\cal N}(0,I/d;{w}^{(0)})
\end{align}
Following Ref. \cite{ariosto2022statistical} we 
introduce the variables $q_{ai}=\int \dif \mu_x \tilde{f}_a(x) \sigma({w}^{(0)}_{a,i} \cdot {x})$ by inserting a product of the following term$$\int \dif q_{ai} \delta(q_{ai}-\int \dif\mu_x \tilde{f}_a(x) \sigma({w}^{(0)}_{a,i} \cdot {x}))$$ such that,
\begin{align}
... &= \int \dif q  e^{-\frac{1}{2N} \sum_{i,a} q^2_{ai}} \prod_{ai} \int \dif{w}_{ai}^{(0)} {\cal N}(0,I/d;{w}_{ai}^{(0)})
\\ \nonumber 
&\cdot \delta\left(q_{ai}-\int \dif\mu_x \tilde{f}_a(x) \sigma({w}^{(0)}_{a,i} \cdot {x})\right) \\ \nonumber 
&\stackrel{\text{Gaussianity}}{\approx} \int \dif q  e^{-\frac{1}{2N} \sum_{i,a} q^2_{ai}} \prod_{ai} {\cal N}(0,C[\tilde{f}_a];q_{ai}) = ... \\ \nonumber 
C[\tilde{f}_a] &= \int  \dif {w}_{ai}^{(0)} {\cal N}(0,I/d;{w}_{ai}^{(0)})\left(\int  \dif\mu_x \tilde{f}_a(x) \sigma({w}^{(0)}_{a,i} \cdot {x})\right)^2 \\ \nonumber 
&= \int  \dif\mu_x  \dif\mu_{x'} \tilde{f}_a(x) K^{(1)}(x,x') \tilde{f}_a(x')
\end{align}
where in the last equality we used Eq. \ref{Eq:NNGP_K_1} which defines the kernel. Performing the remaining Gaussian integration is straightforward and yields 
\begin{align}
Z_w &= \prod_{a=1}^Q \left(\int \dif q  e^{-\frac{1}{2N} \sum_{i=1,a} q^2_{ai}} {\cal N}[0,C[\tilde{f}_a];q_{a,i=1}]\right)^N = e^{-\sum_a\frac{N}{2}\log(1+C[\tilde{f}_a]/N)}
\end{align}

Thus we obtain the following approximation for Eq. \ref{Eq:S_FCN2_nonEK} following this approximate integrating out procedure of the input layer weights  
\begin{align}
&\bar{S}_{\text{FCN2}} \approx -P\sum_a \int \dif \mu_x e^{-\frac{(f_a(x)-y(x))^2}{2\kappa^2}} - \sum_{a=1}^Q i\int \dif\mu_x \tilde{f}_a(x) f_a(x) \\ \nonumber 
&+\sum_a\frac{N}{2}\log(1+C[\tilde{f}_a]/N)
\end{align}

Our second approximation is a mean-field type approximation for the above log term. The underlying assumption is that, $C[\tilde{f}_a]$ which can be re-written in $K^{(1)}$'s eigenfunction basis as $\sum_k \lambda_k \tilde{f}^2_{a,k}$, is a sum of many positive contributions and hence weakly fluctuating. Introducing the mean-field average of $C[\tilde{f}_a]$ under the above action which we denote by $C_{\MF}$), Taylor expanding the log to first order in $\Delta C[\tilde{f}_a]=C[\tilde{f}_a]-C_{\MF}$, and neglecting constant contribution to the action we obtain the following action 
\begin{align}
\label{Eq:S_FCN2_Rotondo}
&\bar{S}_{\text{FCN2,Scaling}}= -P\int \dif \mu_x e^{-\sum_a \frac{(f_a(x)-y(x))^2}{2\kappa^2}} - \sum_{a=1}^Q i\int \dif\mu_x \tilde{f}_a(x) f_a(x) \\ \nonumber 
&+\sum_a \frac{ \int \dif\mu_{x} \dif \mu_{x'}\tilde{f}_a(x)K^{(1)}(x,x')\tilde{f}_a(x')}{2 (1+ C_{\MF}/N)}
\end{align}
several comments are in order: {\bf (i)} Upon integrating out $\tilde{f}$ using simple square completion formulas, the above action coincides with that of data-average GPR (Eq. \ref{Eq:IntroducingSBar}) with a scaled kernel $K(x,x')=K^{(1)}(x,x')/(1+C_{\MF}/N)$. This is why it is often understood as a kernel scaling approximation \citep{ariosto2022statistical}, although the output variance does not simply scale \citep{rubin2024a} \footnote{This may perplex the keen reader given the above action, however, mean-field actions may not reliably predict variances. Specifically, adding source term ($\alpha$) which couples to $f$, one finds that $C_{\MF}$ contains an additional $O(\alpha^2)$ contribution which comes into play only when taking a 2nd derivative w.r.t. $\alpha$ to obtain the variance.} 
{\bf (ii)} at $N \rightarrow \infty$, we retrieve the standard GP-limit. {\bf (iii)} As far as the average predictor goes, scaling down the kernel is equivalent to scaling up the ridge ($\kappa^2$). Hence, we again find a simple renormalization of the ridge, as in Sec. \ref{SSec:Canatar}. Moreover, at vanishing ridge, the simple GP-limit ($N \rightarrow \infty$ with $d,P$ kept fixed) yields the same average predictor.

\begin{figure*}[t]
\vskip -0.2in
\begin{centering}
\includegraphics[width=1\textwidth]{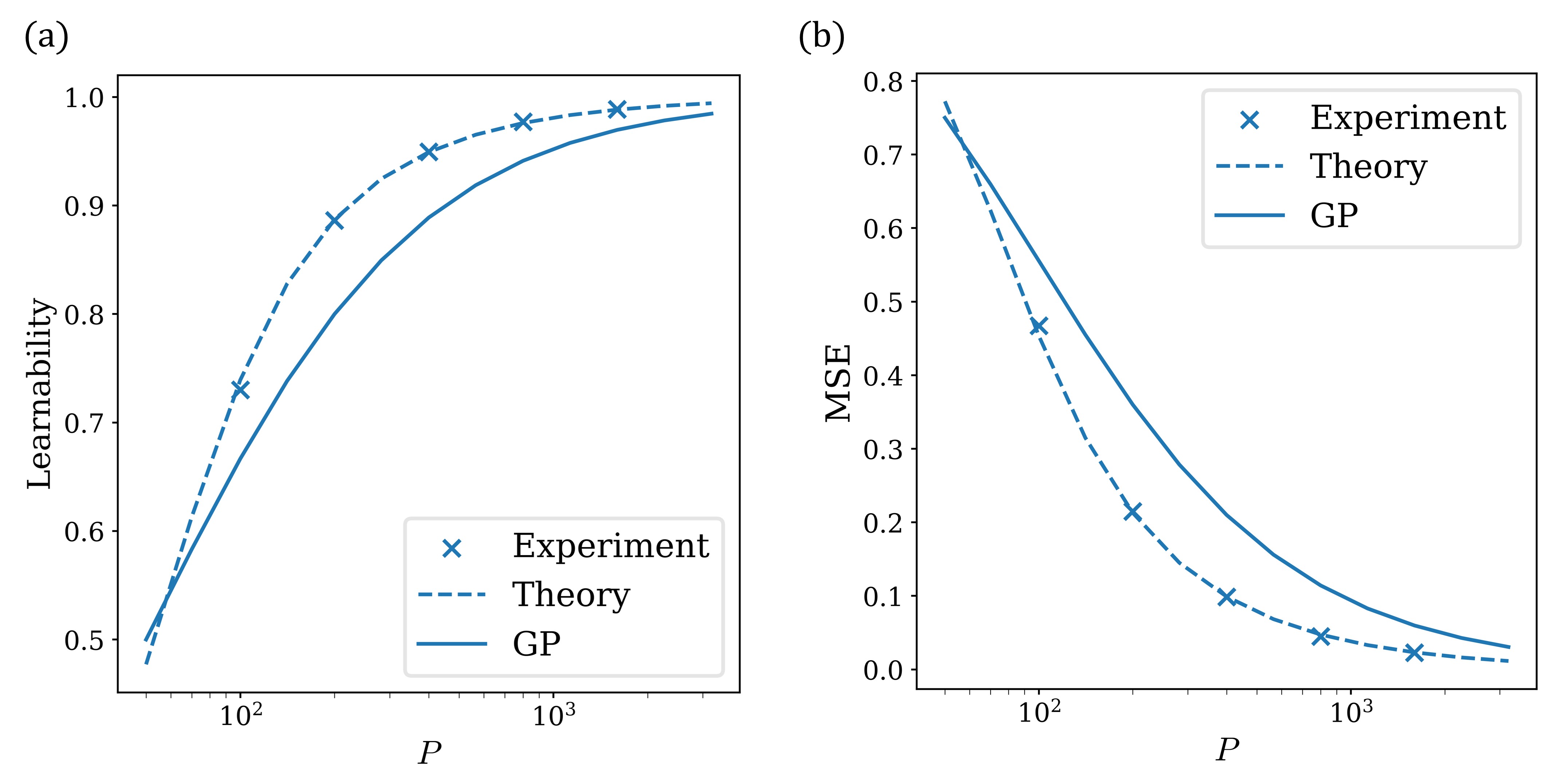}\vskip -0.1in\caption{Here we consider a single hidden layer linear network trained on a linear single index target, and compare the theoretical predictions of the kernel scaling approximation for the network output, as well those of the NNGP. We study two measures for the network output: (a) Learnability which we define as $\frac{f\cdot  y}{ y \cdot y},$ which corresponds to the proportion of the target learned by the network, as well as (b) mean squared test error. Network Parameters:
$d=50,N=1000$, each experimental point corresponds to an ensemble of $\sim$30 networks trained on different data seeds. Each network was trained until there was no visible change to the learnability, loss or hidden layer weight variance.}
\label{fig:scaling_output}
\par\end{centering}
\vskip -0.2in
\end{figure*}

For non-vanishing ridge, one cannot avoid computing $C_{\MF}$. This requires the average and mean of $\tilde{f}_a$ under the above mean-field theory. To this end, we introduce a source term $-i\sum_a \int \dif x \; \alpha(x) \tilde{f}_a(x)$ in the action such that $\delta_{\alpha(x)}\log(Z_{\text{FCN2,Scaling}})|_{\alpha=0}$ and $\delta_{\alpha(x)}\delta_{\alpha(x')}\log(Z_{\text{FCN2,Scaling}})|_{\alpha=0}$ give the average and variance ($Cov$) of $\tilde{f}$ under $\bar{S}_{\text{FCN2,Scaling}}$. To obtain concrete expressions for these averages, we go back to $\bar{S}_{\text{FCN2}}$ (Eq. \ref{Eq:S_FCN2_nonEK}) and note that integrating $\tilde{f}$, in the presence of such source terms, yields a modified RKHS term ($\frac{1}{2} \int ({f}(x)+i\frac{\alpha(x)}{p_{\data}(x)}) K^{-1}(x,x') ({f}(x')+i\frac{\alpha(x')}{p_{\data}(x')})$ where $\dif\mu_x = p_{\data}(x)\dif x$. Taking the above functional derivatives at $\alpha=0$ we obtain 
\begin{align}
&\langle \tilde{f}_a(x) \rangle_{\bar{S}_{\text{FCN2}}} = i \int \dif\mu_x K^{-1}(x,x') \langle f_a(x') \rangle_{\bar{S}_{\text{FCN2}}} \\ \nonumber 
&Cov(\tilde{f}_a(x), \tilde{f}_a(x')) = K^{-1}(x,x')\\ \nonumber &-\int \dif \mu_z \dif \mu_{z'} K^{-1}(x,z) Cov_{\bar{S}_{\text{FCN2}}}(f_a(x),f_a(y)) K^{-1}(z',x') 
\end{align}
The above can be verified to agree with Eq. \ref{Eq:tilde_f_averages_EK} in the EK limit. By replacing the above averages with respect to the exact action, with those with respect to our mean-field approximation ($S_{\text{FCN2,Scaling}}$) we obtain the following self-consistency equation for $C_{\MF}$ namely 
\begin{align}
C_{\MF} &= \int \dif \mu_x \dif\mu_{x'} \langle \tilde{f}_a(x) \tilde{f}_a(x') \rangle_{\bar{S}_{\text{FCN2,Scaling}}} K^{(1)}(x,x') 
\end{align}
where $C_{\MF}$ appears on the r.h.s. via $\bar{S}_{\text{FCN2,Scaling}}$ which is itself $C_{\MF}$ dependent. Using the approximation scheme of Ref. \cite{Canatar2021} for GPR together with the spectral decomposition of $K(x,x')$ ($\lambda_k,\phi_k(x)$) and using Eqs. \ref{eq:barf},\ref{eq:f_k_f_k_connected} we obtain 
\begin{align}
\label{Eq:CMF_Consistency}
\frac{C_{\MF}}{(1+\frac{C_{\MF}}{N})} &= \sum_k \left( \frac{\lambda_k}{\lambda_k + \kappa^2_{\eff}/P} - \frac{P}{\kappa^4_{\eff}} \frac{\lambda_k}{(P/\kappa_{\eff}^2 \lambda_k+1)^2}D - y_k^2 \frac{\lambda_k}{(\lambda_k+\kappa_{\eff}^2/P)^2} \right)
\end{align}
where the first two terms on the r.h.s. come from the covariance of $\tilde{f}$ and the last one from the squared average. Also, $\kappa_{\eff},D$ are those obtained from Eqs. \ref{Eq:CanatarEffectiveRidge},\ref{Eq:D} with $K(x,x')=K^{(1)}(x,x')/(1+C_{\MF}/N)$ where we identify the kernel (down) scaling factor 
\begin{align}
{\cal Q}\equiv (1+C_{\MF}/N)
\end{align}
This self-consistent equation yields highly accurate predictions regarding the network test output, as can be seen in Fig. \ref{fig:scaling_output} for a linear network trained on a single index linear teacher. Accurate predictions have been obtained for nonlinear networks and on real datasets in \cite{RotondoFCN2}.  The above r.h.s. contains three distinct contributions. The first positive term accounts for GPR fluctuations, and roughly counts the number of learnable modes. The two other terms are associated with the second moment of the average predictor, across data-set draws. All these different contributions are extensive in the input dimension or $P$, consistent with our previous result that $N \gg P$ is required for the GP-limit (see also \cite{Hanin2023} for a proof of this and extension to infinite depth). 

A simple limit of the above formula is when the original/bare $\kappa^2$ goes to zero, as in this case the kernel scaling factor (${\cal Q}$) affects both $\lambda_k$ and $\kappa_{eff}$ the same. Specifically, both $\lambda_k$ and $\kappa_{\eff}^2$ get scaled down by ${\cal Q}\equiv (1+C_{\MF}/N)$, from their GP value since the latter by being a sum of the former. Using the notation $\lambda_{k}^{(1)} = \lambda_k {\cal Q},\ \kappa^{2}_{(1),\eff} = \kappa_{\eff}^{2}{\cal Q}$, where $\lambda_{k}^{(1)}$ are the eigenvalues of the GP kernel $K^{(1)}$ and $\kappa^{2}_{(1),\eff}$ is the GP solution to Eq. \ref{Eq:CanatarEffectiveRidge}, we can take out this ${\cal Q}$ scale, resolving all $C_{\MF}$ and $N$ dependencies of the r.h.s. we obtain a simple quadratic equation for ${\cal Q}$
\begin{align}
\frac{N({\cal Q}-1)}{{\cal Q}} &\stackrel{{\scriptscriptstyle \kappa\rightarrow0}}{=} \sum_k \frac{\lambda_{k}^{(1)}}{\lambda_{k}^{(1)} + \kappa^{2}_{(1),\eff}/P} - \frac{{\cal Q}\lambda_{k}^{(1)}D }{P(\lambda_{k}^{(1)}+\kappa^{2}_{(1),\eff}/P)^2}- \frac{{\cal Q} y_k^2  \lambda_{k}^{(1)}}{(\lambda_{k}^{(1)}+\kappa^{2}_{(1),\eff}/P)^2}
\end{align}
where we also note that $D$ is independent of ${\cal Q}$. 

Let us flesh out the usefulness of such ``scale learning" at different bias (average-predictor) variance (GPR fluctuations) trade-offs. When the target is large \footnote{which can also be seen as a form of mean-field scaling, namely taking the variance of $a_i$'s to be   $O(1/N)$ time some small number.}, the loss in the strict GP-limit will mainly depend on average predictor (i.e. resolution-limited regime). In the above, the two negative, target-dependent, terms would dominate the r.h.s. We denote the sum (taken as positive) by $Y$ which is $O(y^2)$ and ignore the first term on the r.h.s. Using this notation the above equation can be written as $\frac{Y}{N}{\cal Q}^2+{\cal Q}-1=0$, yielding, in our limit of large target (specifically $\frac{Y}{N}\gg 1$), ${\cal Q} \approx \sqrt{N/Y} \ll 1$. Thus, the kernel scales up to partially approach the scale of the target, still however leaving the bias term dominant due to the rather slow $O(\sqrt{Y})$ kernel-scaling. In the opposite regime, of a small target, the first positive term on the r.h.s. dominates. Based on \ref{Sec:ScalingLaws}, this term, which counts the number of learnable modes, would scale as $P$. Making an order of magnitude analysis, we thus have ${\cal Q}-1={\cal Q} \frac{O(P)}{N}-{\cal Q}^2 \frac{Y}{N}$, implying at $Y \ll |O(P)/N-1|$, ${\cal Q} = O(P)/Y$. Thus, the kernel is scaled down, to meet the magnitude of the target, thereby making the variance contribution to the loss comparable to the bias. Notably, this happens in a setting where, GP-wise, we expect the variance to be strongly dominant. 

{\bf Revisiting the Gaussianity assumption.} We turn to verify the Gaussianity assumption underlying this analysis. Notably this assumption is equivalent to saying that under a typical draw of $\tilde{f}(x)$, $\int \dif \mu_x \tilde{f}_a(x) \sigma({w}^{(0)}_{a,i} \cdot {x})$ is, to a good approximation, a linear function of ${w}^{(0)}$ as this is the most general Gaussian variable under ${\cal N}(0,I/d;w^{(0)})$. Focusing again on the case of a large target (or mean-field scaling) $\tilde{f}(x)$ would be weakly fluctuating around its mean. We thus aim to  assess how close the next expression is to a linear function of $w^{(0)}$
\begin{align}
\int \dif\mu_x \langle \tilde{f}_a(x)\rangle_{S_{\text{FCN2}}} \sigma({w}^{(0)}_{a,i} \cdot {x}) &\propto \int \dif\mu_x [\langle f(x) \rangle_{S_{\text{FCN2}}} - y(x)] \sigma({w}^{(0)}_{a,i} \cdot {x}) =... 
\end{align}

For concreteness, let us next consider an FCN network with data taken from a uniform measure on the hypersphere. In this case, we may decompose $y(x)$ as $\sum_l \phi_l(x) y_l$ where $\phi_l(x)$ is the sum of all spherical Harmonics of order $l$ in $y(x)$ and $y_l$ is the norm of their coefficients. We thus obtain  
\begin{align}
... &= \sum_{l=0}^{\infty} \frac{\kappa^2_{\eff}/P}{\lambda_l + \kappa^2_{\eff}/P} y_l \int \dif\mu_x \phi_l(x) \sigma({w}^{(0)}_{a,i} \cdot {x})
\end{align}
Taking $\sigma(..)$ to be linear, the r.h.s can only be a linear function of ${w}^{(0)}$ and the contribution of $l>1$ is exactly zero. Taking $\sigma(..)$ to be $\text{Erf}$, and focusing on $\phi_{l=1}(x)$ which we may write as ${w}_* \cdot x$, the above integral is $\sqrt{\pi/2} ({w}_* \cdot {w}_{a,i}^{(0)})/\sqrt{1+2|{w}_{a,i}^{(0)}|^2}$ \citep{rubin2024grokking}. Thus at large input dimension, where the fluctuations of $|{w}_{a,i}^{(0)}|$ become negligible, we indeed obtain a contribution linear in ${w}^{(0)}$, in support of the Gaussianity assumption. However, for a non-linear target and at $P$ large enough the $l=0,l=1$ contribution to the discrepancy would be highly learnable and thus strongly suppressed compared to $l=2$ and above. Focusing on $l=2$, we then formally Taylor expand the activation function to infinite order, and note that the zeroth and first-order terms in this expansion are fully spanned by the $l=0,l=1$ spherical Harmonics. Consequently, these $O(1),O(w^{(0)})$ contributions disappear from the integral. The remaining contributions, say $l=2$ one, would give rise to $(w^* \cdot w^{(0)}_{a,i})^2$ or higher terms \citep{rubin2024grokking} which are clearly non-Gaussian random variables under ${\cal N}(0,I/d;w^{(0)})$. Examining the original argument for Gaussianity, in the light of this result, the culprit is that $\tilde{f}_a(x)\sigma({w}^{(0)} \cdot x)$ becomes correlated across different $x$ values, thereby undermining the central limit theorem which relies on having no or weak correlations between the summed random variables.  

We conclude that the assumption underlying kernel scaling breaks down when the correlation between linear functions and the discrepancy over the dataset measure becomes negligible compared to those of higher-order functions. This is likely to be the case when the linear components of the target have been almost fully learned, as one expects for small $\kappa^2$ and $P \gg d$. Taking Gaussian iid data, mean-field scaling, and $P \propto d$ one indeed finds that kernel-scaling fails namely, actual finite neural networks can learn the higher (quadratic, cubic, etc...) components of the target 
\citep{abbe2021staircasepropertyhierarchicalstructure,naveh2021self,  Cui2023,arous2021online,rubin2024a}. In contrast, GPR with the scaled kernel requires $P \propto d^l$ to learn the $l$'th target component. At finite ridge, this poorer scaling stems from the fact that the learnability of an $l$'th spherical Harmonics goes as  $\kappa^{-2}_{\eff}P\lambda_l < \kappa^{-2} P\lambda_l \propto P d^{-l}$. At zero-ridge, this can be argued for in more detail based on our formulas. Instead, we suffice ourselves with the following heuristic argument: GPR with the scaled kernel is still rotationally invariant as is the Gaussian iid data. As such its Gaussian prior favors all $l=2$ spherical harmonics the same. Furthermore, due to spectral bias, information learned from the $l=1$ piece of the target cannot affect the $l=2$ piece. As there are $O(d^2)$ $l=2$ functions, all equally likely, $n=O(d^2)$ data points are needed to determine their coefficients. 

We note that an actual DNN can work differently. Specifically, it can use the knowledge obtained when learning the $l=1$ component of the target, and use that to emphasize certain $l=2$ Spherical Harmonics. This form of assisted (or staircase \cite{abbe2021staircasepropertyhierarchicalstructure}) learning can be tracked via the adaptive kernel approach which we next discuss.   

\section{Approximation Scheme: Kernel Adaptation}
\label{Sec:KernelAda}
An essential strategy of statistical physics is to describe complex systems in terms of a few order parameters. In the previous analysis this order parameter was ${\cal Q}$, the scaling factor of the kernel which, for deeper networks, would become a set of scaling factors (see Refs. \cite{LiSompolinsky2021,ariosto2022statistical} where this is made explicit). Here we consider a richer set of order parameters meant to capture a richer set of phenomena-- the layer-wise kernels themselves. Instead of obtaining a non-linear equation for the scaling factor (i.e. Eq. \eqref{Eq:CMF_Consistency}) we would obtain an equation for these kernels. This would have the clear disadvantage of being more complex but the advantage of describing how internal representation develops and captures beyond-GPR effects in DNNs, leading to a new sample complexity class, as explained below. 

{\bf Related approaches.} This line of thought has been explored concurrently by several groups with various similarities and differences. The equilibrium statistical mechanics approach was used in \citep{naveh2021self,seroussi2023separation} and can also be viewed as a first-principles statistical mechanics formalization of the ideas of kernel flexibility of Refs. \cite{aitchison2019bigger,aitchison2021deepProcess} and a generalization of the latter to non-linear networks. Analogous dynamical approaches appear in Refs. \citep{yang2022tensorprogramsvtuning,bordelon2022self}, but focused on on-data results and avoided some of the approximation we take below (specifically, Gaussian or Gaussian mixture approximation for pre-activations) and traded those with numerics, otherwise they essentially do the same approximation (layer-wise decoupling using average kernel). Unique to Refs. \citep{naveh2021self,seroussi2023separation} is the observation that in many scenarios (but not all \citep{seroussi2023separation,rubin2024grokking}), pre-activation statistics are Gaussian or mixtures of Gaussians \cite{rubin2024grokking}. It should be noted that unlike \citep{seroussi2023separation}, where the transition from on-data to data-average was carried in an ad-hoc manner \footnote{Specifically the ``$q$-factors" of that work which coincide with the more careful treatment carried here.}-- this review takes a full data-averaged approach and avoids using any data-sized matrices.  

The kernel adaptation approach relies on one main approximation (mean-field layer decoupling) and an additional Gaussianity approximation, which often applies yet needs case-by-case analysis. We first provide an overview of these approximations and then demonstrate them on a simple model. 

{\bf Mean-field decoupling between layers.} In multi-layer actions, e.g. Eq. \eqref{Eq:BarSFCN3}, we will treat terms such as $\frac{1}{N}\sum_{i=1}^N \sigma(h_{ia}^{(l)}(x))\sigma(h_{ia}^{(l)}(y))$ as weakly fluctuating and replace them by their mean-field average and keep only leading order dependence in their fluctuations. By the same token, we would treat terms such as $\frac{1}{N}\sum_{i=1}^N\tilde{h}^{(l)}_{ia}(x)\tilde{h}^{(l)}_{ia}(y)$ as weakly fluctuating (see Eq. \ref{Eq:FCN3}). In addition, we will treat $\tilde{f}_a(x) \tilde{f}_a(y)$ as weakly fluctuating even though it is not a sum over many variables. 

As we would show below, this approximation is justified under mean-field scaling. We introduce such scaling in our action by dividing $\sigma_a^2$ by an additional $\chi \gg 1$ factor which suppresses $\tilde{f}$ fluctuations \footnote{An alternative approach which avoids averaging over the $\tilde{f}$-fields is to introduce kernel-fields using delta-functions, and apply the Gärtner-Ellis theorem \cite{fischer2024critical} which yields the same result.} Taking $\chi=N$ this scaling of parameters is consistent with the mean-field scaling as introduced in Sec. \ref{Sec:Hyper}. When working on-data rather than in the data-averaged/field-theory scenario, as we have done so far, $\chi \gg 1$ (let alone $\chi=N$) is alone controls this approximation. However, as in this review we also take into account the dataset ensemble, we further require that dataset-induced fluctuations do not dominate the mean of $\tilde{f} \tilde{f}$. As we will see below, this is typically the case when overall performance is good and feature learning in consistent between different draws of the same dataset. 

As shown in Ref. \cite{seroussi2023separation}, the above mean-field approximation leads to a set of decoupled layer-wise actions that ``see'' each other through averages of the aforementioned term/operators. In the presence of nonlinear activations function these layer-wise  actions are non-Gaussian requiring additional approximation such as VGA (variational Gaussian approximation) or saddle point or in some cases involving phase transitions, an approximation via a mixture of Gaussians \cite{rubin2024grokking}. 

To demonstrate the various aspects of this approximation scheme, we will consider both 2-layer FCNs and CNNs with non-overlapping convolution windows via the following network 
\begin{align} \label{Eq:2_layer_FCN}
    f({x}) &= \sum_{i=1}^{N_w} \sum_{c=1}^{C} a_{ic} \sigma\left({w}_c \cdot {x}_{i}\right) 
\end{align}
where ${x} \in {\mathbb{R}}^{d}$ with $d = N_w S$ and ${w}_c, {x}_{i} \in {\R}^{S}$. The vector ${x}_{i}$ is given by the $iS,..,(i+1)S-1$ coordinates of ${x}$. 
The dataset consists of $\{ {x}_{\mu} \}_{\mu=1}^{n}$ i.i.d. samples, each sample ${x}_\mu$ is a centered Gaussian vector with covariance $I_d$. The FCN case is obtained by setting $N_w=1$. Notably $C$ now takes the role of $N$ of the previous sections as controlling the overparametrization.   

As before, we train this DNN using GD+noise/Langevin training. However, we also introduce the $\chi$, mean-field scaling factor, to the readout layer weights in the following manner. First, we define $\kappa^2/\chi$ as the noise rather than $\kappa^2$. Next, we tune the weight-decay such that without any data, $a_{ic} \sim {\mathcal N}(0,\sigma_\text{a}^{2}(N_w C)^{-1} \chi^{-1})$ and $[{w}_c]_s \sim {\mathcal N}(0,\sigma_\text{w}^2 S^{-1})$. This is set up such that at the GP limit ($C \rightarrow \infty$), the mean predictor is unaffected by $\chi$ whereas fluctuations are reduced by a factor of $\chi$. 

Before delving into a detailed analysis, let us flesh out three main questions one can address using such a model: {\bf Weight-sharing and laziness.} Focusing for simplicity on $\sigma(x)=x$, one can show, by direct computation, that the GP kernel associated with this network coincides with that of a linear FCN. Clearly, this limit does not reflect the presence of convolutional patches ($i$ index) and misses out on weight-sharing. Similar mismatches between CNNs and their lazy limits were first noted in Ref. \cite{novak2019neural} and used as a qualitative explanation for the poorer performance of the lazy regime. Thus, feature learning corrections play a qualitative role here in restoring the weight-sharing and local connectivity structure of CNN. {\bf Sample complexity.} The next question concerns sample complexity, a concept used to describe the scaling of $n$ with $d_{\inn}$ required for good learning. For $C=O(1)$, and a linear target of the form $y_{\mu} = \sum_{i} a^*_i ({w}^* \cdot {x}_{\mu,i})$ where $a^*_i \sim {\mathcal N}(0,1/N_w)$ and $w^*_s \sim {\mathcal N}(0,1/S)$, $n$ of the order of the number of parameters ($O((N_w+S)C)=O(\sqrt{d_{\inn}})$) is enough to fit the target. We expect the sample complexity (the dependence of $n$ on $d_{\inn}$) of a CNN to be better than that of an FCN ($O(d_{\inn})$) or similarly the lazy limit of that same CNN. {\bf Assisted learning and staircase function.} Another qualitative mechanism \citep{abbe2021staircasepropertyhierarchicalstructure,dandi2023twolayerneuralnetworkslearn} present even in the FCN ($N_w=1$) case, is that simple target components (e.g. linear ones such as $w^* \cdot x$) can focus the network's attention along particular directions in input space (e.g. ${w}^*$) thus enabling one to learn non-linear components at $n=O(d_{\inn})$ even though the lazy limit puts the associated eigenvalues at $1/d^{2}_{\inn}$ or less. Thus, the linear component may assist (or act as a ``staircase'') in learning the more complex parts of the target, something that a GP cannot do, at least not within the eigenlearning framework. 

Our starting point is the equivalent of \ref{Eq:BarSFCN3} for our CNN 
\begin{align}
\label{Eq:S_CNN2}
&\bar{S}_{\text{CNN2}}= -P\int \dif\mu_x e^{-\sum_{a=1}^Q \frac{(f_a(x)-y(x))^2}{2\kappa^2/\chi}} - \sum_{a=1}^Q i\int \dif\mu_x \tilde{f}_a(x) f_a(x) \\ \nonumber 
&+\frac{1}{2CN_w \chi} \sum_{c,a,i}^{C,Q,N_w} \left(\int \dif\mu_x \tilde{f}_a(x) \sigma({w}_{c,a} \cdot {x}_i)\right)\left(\int \dif\mu_{x'} \tilde{f}_a(x') \sigma({w}_{c,a} \cdot {x}'_i)\right) \\ \nonumber &+ \sum_{c,a}^{C,Q}\frac{S |{ w_{c,a}}|^2}{2}
\end{align}
where we replaced ${w}^{(0)}$ by ${w}$ for compactness. 

Our first approximation here is a mean-field approximation wherein we first rewrite $\tilde{f}_a(x) \tilde{f}_a(x')$ in terms of its average value ($A(x,x')$ assuming replica symmetry) and fluctuation namely $\tilde{f}_a(x) \tilde{f}_a(x')=A(x,x')+\Delta A_a(x,x')$ where $\Delta A_a(x,x') = \tilde{f}_a(x) \tilde{f}_a(x')-A(x,x')$ and $(CN_w)^{-1}\sum_{c,i=1}^{C,N_w}\sigma({w}_{a,i} \cdot {x}_i)\sigma({w}_{a,i} \cdot {x}'_i)=K({x},{x}')+\Delta K({x},{x}')$ where $K({x},{x}')+\Delta K({x},{x}')=(CN_w)^{-1}\sum_{c,i=1}^{C,N_w}\sigma({w}_{a,i} \cdot {x}_i)\sigma({w}_{a,i} \cdot {x}'_i)-K({x},{x}')$ and then expand the action up to linear order in $\Delta K,\Delta A$. Ignoring irrelevant constant additive terms in the action, we obtain 
\begin{align}
\bar{S}_{\text{CNN2}} &= S_{\text{readout}}+S_{w}+O(\Delta^2) \\ \nonumber 
S_{\text{readout}} &= -P\int \dif\mu_x e^{-\sum_{a=1}^Q \frac{(f_a(x)-y(x))^2}{2\kappa^2/\chi}} - \frac{1}{\chi}\sum_{a=1}^Q i\int \dif\mu_x \tilde{f}_a(x) f_a(x) \\ \nonumber &+ \frac{1}{2}\sum_{a=1}^Q \int \dif\mu_x \dif\mu_{x'} K(x,x') \tilde{f}_a(x) \tilde{f}_a(x') \\ \nonumber 
S_{w} &= \frac{1}{2CN_w \chi} \sum_{c,a,i}^{C,Q,N_w} \int \dif\mu_x \dif\mu_{x'} A(x,x') \sigma({w}_{a,i} \cdot {x}_i)\sigma({w}_{a,i} \cdot {x}'_i) + \frac{S |{ w_{c,a,i}}|^2}{2}
\end{align}
where $A(x,x')$ and $K(x,x')$ should be determined self-consistently under the above action via their definitions as averages. As advertised, as a result of the mean-field decoupling, two decouple actions have been obtained, one for the read-out layer and one for the input layer. These "see" each other only through the average/mean-field quantities $K(x,x')$ and $A(x,x')$.

Examining $S_{\text{readout}}$, one notes that integrating out the Gaussian $\tilde{f}$ fields, we find that it coincides with the dataset averaged GP action having a ridge given by $\kappa^2/\chi$ and a kernel given by $K(x,x')/\chi$. Thus, we can readily apply the approximations for GP inference of Chapter \ref{Sec:AveragedGPR} to this action. In particular, as far as the average predictor goes, we would find that it behaves effectively as a quadratic theory with $\kappa^2/\chi$ replaced by $\kappa^2_{\eff}/\chi$. Moreover, the fields can be rescaled such that $\chi$ multiplies $y$ thus leading to GPR with $\kappa^2_{\eff}$ ridge, $K(x,x')$ as the kernel, and $y(x)\chi$ as the target. 

Turning to $S_{\text{readout}}$, it clearly describes $C$ times $N_w$ times $Q$ iid weight distributions associated with each ${w}_{c,a,i}$. We may then focus on just one representative weight which we would simply denote by ${w}$. 

Solving the model now amounts to computing the values of $K(x,x')$ (which would depend on $A(x,x')$ appearing in $S_w$) and $A(x,x')$ (which would depend on $K(x,x')$) and requiring self-consistency leading, in general, to a closed operator equation relating $A(x,x')$ and $K(x,x')$. Similar results follow for deeper networks, where $K(x,x')$ would generalize to layer-wise kernels ($K^{(l)}(x,x')$) and $A^{(l)}(x,x')$ to layer-wise auxiliary field with $l$ being the layer number. For simple datasets, such as those considered below, these operators change in only a few relevant directions, simplifying the analysis considerably.  

Let us first consider the case of linear activation ($\sigma(x)=x$) where the action for ${w}$, following our mean-field decoupling above, is Gaussian and appears in quadratic form as  
\begin{align}
S_w &= \frac{1}{2} \sum_{a,i}^{Q,N_w} {w}_{a,i}^T \left[\frac{1}{{C} N_w \chi} \int \dif\mu_x \dif\mu_{x'} A(x,x') {x}'_i {x}^{T}_i+I_{S \times S}S\right] {w}_{a,i} 
\end{align}
where, notably, all different channel indices ($i$) and replica indices ($a$) came out decoupled (i.e. going from $S_w$ to probabilities via $e^{-S_w}\propto P(w)$, the latter is a product of iid distributions across these indices). As we shall soon see, $A(x,x')$ goes as {\it minus} the square discrepancy and hence acts to stretch the distribution of $w$'s along directions which overlap with the discrepancy. In other words, the weight covariance adapts to align with the error in predictions. Another observation is that $A(x,x') \propto \chi^2$, hence taking mean-field scaling in the sense of $\chi = N$, one loses the $N$ dependence, explaining why in this parameterization feature learning persists regardless of $N$. 

Next we note that given our target ($y_{\mu} = \sum_{i} a^*_i \sigma({w}^* \cdot {x}_{\mu,i})$) and choice of data measure (${x} \sim {\cal N}(0,I_{N_w S \times N_w S})$) our original action (Eq. \ref{Eq:S_CNN2}), is symmetric to any rotation of the $w_i$'s along the $O(S-1)$ directions orthogonal to $w^*$. Consequently, the $S \times S$ covariance matrix of $w_i$, must be of the form $c_{\perp} I + (c_*-c_{\perp}) {w}^* [{w}^*]^T$ for some two scalar $c_{\perp},c_*$, as this is the most general (real, symmetric) matrix which is invariant under the above $O(S-1)$ symmetry. As a result 
\begin{align}
\label{Eq:AdaptedKernel1}
K(x,x')&=N_w^{-1} \sum_{i=1}^{N_{w}} \langle ({w} \cdot x_i) ({w} \cdot x'_i) \rangle_{S(w)} \\ \nonumber &= N_{w}^{-1} \sum_{i=1}^{N_{w}} (x'_i)^T [ c_{\perp} I + (c_*-c_{\perp}) {w}^* [{w}^*]^T] x_i 
\end{align}
at which point we obtain {\bf an intuitive explanation} for the extra learning abilities of DNNs: The discrepancy in predictions (embodied by the above $A(x,x')$) can generate a new non-NNGP like kernel which can distinguish between target relevant and target irrelevant directions whenever $c_* > c_{\perp}$. Next, we proceed by computing this $c_*$ and seeing its quantitative effect.

Given our simple Gaussian measure for $x$, this kernel can be diagonalized as follows: $N$-fold degenerate eigenfunctions associated with function which depend only on the teacher direction on each patch namely,  
\begin{align}
\int \dif\mu_{x'} K(x,x') ({w}^* \cdot x_i) &= \frac{c_* |{w}^*|^2}{N_w} ({w}^* \cdot x_i) \equiv \lambda_* ({w}^* \cdot x_i)
\end{align}
whereas all other linear functions are degenerate with an eigenvalue $c_{\perp}/N_w$. 

Having expressed the action of the weights and $K(x,x')$ in terms of $c_{\perp},c_*$ we turn to compute the GPR implied by $S_{\text{readout}}$ with the above $K(x,x')$ and deduce $A(x,x')$ thereby closing our equations for $c_{\perp},c_*$. Since all eigenfunctions are Gaussian in the sense of Sec. \ref{SSec:Canatar} we may use the approximation therein to obtain 
\begin{align}
\bar{f}_{a}(x) &= \frac{\frac{c_* |{w}^*|^2}{N_w}}{\frac{c_* |{w}^*|^2}{N_w}+\kappa^2_{\eff}/P} y(x) 
\end{align}
where we also used the fact that $y(x)$ is a superposition of the above $N$-fold degenerate manifold. Focusing on the statistics of $\tilde{f}_a(x)$, relevant for $A(x,x')$, one can show by introducing a source field which couples to $\tilde{f}_a(x)$, integrating over $\tilde{f}_a$, and taking derivatives w.r.t. to that source that  
\begin{align}
\bar{\tilde{f}}_a(x) &= i \chi \int \dif \mu_{x'} K^{-1}(x,x') \bar{f}_a(x') = \frac{i\chi}{\frac{c_* |{w}^*|^2}{N_w}+\kappa^2_{\eff}/P} y(x) \\ \nonumber 
\langle \tilde{f}_a(x)\tilde{f}_a(x')\rangle_{\con} &= \chi^2 \int \dif \mu_{z} \dif\mu_{z'} K^{-1}(x,z) K^{-1}(x,z') \langle f_a(z)f_a(z') \rangle_{\con}\\ \nonumber &-\chi K^{-1}(x,x')
\end{align}
where $ \langle f_a(z)f_a(z') \rangle_{\con}$ is given, in the eigenfunctions basis, by Eq. \eqref{eq:f_k_f_k_connected}. Working at large $\chi$, we focus on the dominant, $O(\chi^2)$ contributions to $A(x,x')$ given by 
\begin{align}
A(x,x') &= \bar{\tilde{f}}_a(x) \bar{\tilde{f}}_a(x')+\chi^2 \sum_k \frac{\phi_k(x) \phi_k(x')}{\lambda_k^2} \left(\frac{P}{\kappa^2_{\eff}}+\lambda_k^{-1}\right)^{-2} \frac{P}{\kappa^4_{\eff}}D 
\end{align}
where $D$ is defined in Eq. (\ref{Eq:D}). 
Anticipating that the network would learn the target well (i.e. $\lambda_k$ associated with $y(x)$ would be much larger than $\kappa_{\eff}^2/P$), one can show that the second term which reflects dataset fluctuations should be suppressed by $\kappa^2_{\eff}/P$ compared to the first and hence negligible. Conveniently $A(x,x')$ can now be written as $y(x)y(x')(\lambda_* + \kappa_{\eff}^2/P)^{-2}$ which, placed in the quadratic form in $S(w)$ yields the mean-field self-consistency equation 
\begin{align}
[c_{\perp} I + (c_*-c_{\perp}) {w}^* [{w}^*]^T]^{-1} &= c_{\perp}^{-1} I + (c^{-1}_*-c_{\perp}^{-1}) {w}^* [{w}^*]^T \\ \nonumber 
&= -\frac{\chi}{CN_w} \sum_i |a^*_i|^2 {w}^* [{w}^*]^T (\lambda_* + \kappa_{\eff}^2/P)^{-2} + S I. 
\end{align}
By projecting both sides of this equation on any direction orthogonal to $w^*$, we obtain $c_{\perp}=S^{-1}$, i.e. no feature learning along orthogonal directions to the target \footnote{Had we kept the dataset fluctuation terms, proportional to $D$, we would find an insignificant increase in variance in those directions when the target is well learnable.}. Focusing for clarity on the case where $|{w}^*|=|{a}^*|=1$, we obtain the following equation for $c_*$ 
\begin{align}
c^{-1}_* &= S - \frac{\chi}{CN_w}\frac{1}{(\frac{c_*}{N_w}+ \kappa_{\eff}^2/P)^2}  
\end{align}
Assuming like before that $\lambda_* \gg \kappa^2_{\eff}/P$, this simplifies to $c^{-1}_* = S - \frac{\chi}{C} \frac{N_w}{c^2_*}$ implying $c^2_*- S^{-1}c_* - \frac{\chi N_w}{C S}=0$. 

\begin{figure*}[t]
\vskip 0.2in
\begin{centering}
\includegraphics[width=0.8\textwidth]{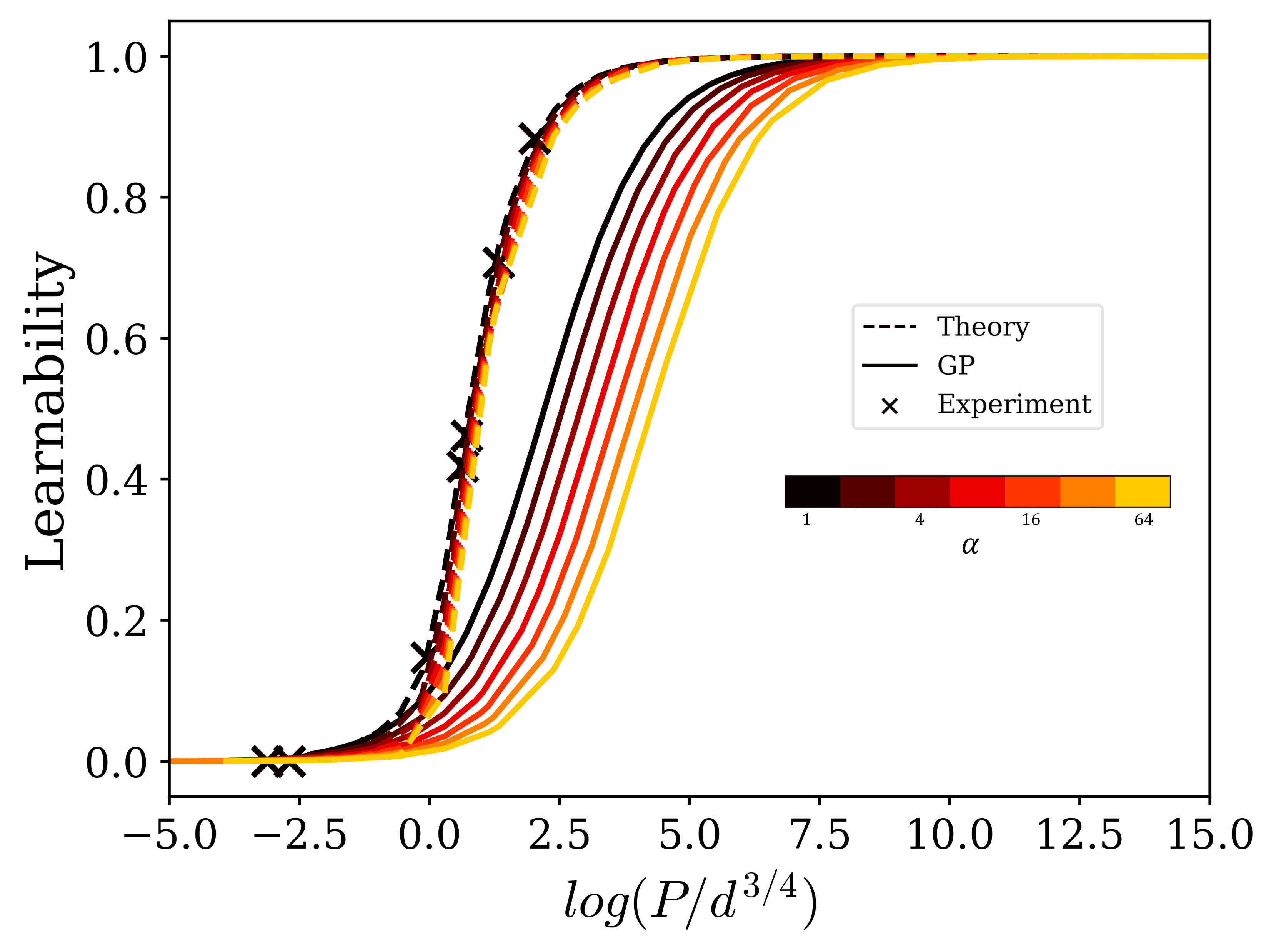}\caption{Learnability of linear CNNs as a function of $P$. We take $S,N,C \propto \alpha$, and consider different $\alpha$ scales of these parameters. Here the network is observed to learn the target at $P\propto d^{3/4}$, regardless of the parameter scale, as opposed to the GP predictions which predict learning at $P\propto d$. Parameters: $\chi=100$, $N=10\alpha,S=50\alpha,C=1000\alpha$.}
\label{fig:sample_complexity}
\par\end{centering}
\vskip -0.2in
\end{figure*}

Let us analyze the above result. Taking first a mean-field scaling of the type $\chi=N_w$ and $N_w=S$, the above equation implies $c_* = 1 + O(1/S)$ and therefore $\lambda_* = 1/N_w + O(1/d_{\inn})$ whereas $\lambda_{\perp}=1/d_{\inn}$. This $O(S)$ inflation of the eigenvalue along the teacher direction implies that $P=O(N_w)$ samples are sufficient to learn the target well instead of $O(N_w S)$ as in the lazy regime or for an FCN \footnote{The fact that an FCN cannot do better than their lazy limit in this linear setting is implied by the kernel-scaling result}. Taking instead $\chi$ to be larger but finite and $N_w \propto S \propto C$, one still obtains a good approximation (though not asymptotically exact). In this milder form of mean-field scaling, we find, $c_* = \sqrt{ \frac{\chi N_w}{SC}}$ hence $\lambda_* = \sqrt{\chi/(SN_w C})$ implying $P=O(\sqrt{SN_w C})=O(d_{\inn}^{3/4})$ samples are enough. This is again a better sample complexity class compared to the lazy limit or an FCN though not as good as that of mean-field scaling. The change in sample complexity relative to the lazy limit can be observed in Fig. \ref{fig:sample_complexity}, where this approach predicts that the network will indeed learn the target at $P=O(d_{\inn}^{3/4})$ in agreement with experiment and in contrast to the GP predictions. We define learnability as the components of the target learned by the network, given by- $\frac{ f \cdot y}{ y \cdot y}$, so that learnability=1 implies perfectly learning the target.


Finally, we comment on $\kappa^2_{\eff}$. Since the readout layer coincides with the data-averaged GP action, $\kappa^2_{\eff}$ obeys Eq. \ref{Eq:CanatarEffectiveRidge} w.r.t. to the adapted kernel $K(x,x')$. In general, this adds an additional equation which must be solved (typically numerically) together with that for $c_*$. For large $\kappa^2$ (compared to the MSE), this can be avoided using the EK approximation for $S_{\text{readout}}$. For small or intermediate values of $\kappa^2$ one notes that $K(x,x')$ contains a single outlier eigenvalue which is well-learnable and typically well-learnable eigenvalues do not contribute to $\kappa^2_{\eff}$ . Hence, $\kappa^2_{\eff}$ can be approximated by its value w.r.t. to the non-adapted kernel, thereby decoupling it from the equation for $c_*$. Moreover, in our simple setting explicit expressions for $\kappa^2_{\eff}$ are given in Ref. \cite{Canatar2021}.   

{\bf Non-linear networks.} Let us next generalize this result to non-linear activations, such as $\sigma(x)=\text{Erf}(x)$. Repeating the previous analytical workflow, symmetry would still play the same role, implying that $c_*,c_{\perp}$ are the only relevant parameters to track. The main complication here compared to the linear case is that the action $S_w$, is non-Gaussian and hence it is unclear how to calculate $K(x,x')$, a quantity which includes averages w.r.t. $S_w$. Another, more technical complication, is that the spectrum of $K(x,x')$ becomes harder to diagonalize. Both these issues turn out to be subleading for large enough $S$, where due to concentration phenomena in the mean-field regime, we may replace $|{x}_i|^2$ and $|{w}|^2$ by their mean. 

Turning to the first issue, let us assume and later verify the assumption self-consistently, that $A(x,x')$ maintains the previous form ($A(x,x') \propto y(x) y(x')$). Using the fact that, 
\begin{align}
\label{Eq:Useful_Integral}
\int \dif x {\cal N_w}(0,I_{S\times S};x) \Erf({w} \cdot x) {v}\cdot x= \frac{2}{\sqrt{\pi}} \frac{{w} \cdot {v}}{\sqrt{1+2|{w}|^2}}
\end{align}
we obtain the following expression for $S_w$ given the above ansatz for $A(x,x')$
\begin{align}
\frac{1}{2CN_w \chi} \frac{1}{(c_*/N_w+\kappa^2_{\eff}/P)^2} \sum_i \frac{4}{\pi} \frac{({w}_i \cdot {w}^*)^2}{1+2|{w}_i|^2}.
\end{align}
A general approximate tool for handling such non-quadratic action is the Variational Gaussian Approximation (VGA), directly analogous to the Hartree-Fock approximation in physics. This was used in Ref. \cite{seroussi2023separation}, based on the general heuristic argument that actions which involve many similar degrees of freedom with all-to-all interactions often admit a good Gaussian approximation. Here we can show this in more concrete terms, as the source of all non-quadratic behavior here is $|{w}_i|^2=\sum_{k=1}^S [{w}_i]_k^2$ appearing in the denominator. Being a sum over many ($S$) degrees of freedom it is reasonable to replace it by its mean ($|{w}_i|^2=1$) and neglect fluctuations entirely (we will revisit this approximation when we take $\chi = O(N_w)$). Following this we obtain a quadratic action and the only effect of the non-linearity on $S_w$ is the extra factor of $\frac{4}{3\pi} \approx 0.42$ accompanying the ${w}^*$ dependent term in the action compared to the linear case. We may thus readily compute $K(x,x')$ within this VGA/mean-field approximation. Specifically, using a direct application of the computation leading to Eq. \ref{Eq:ErfKernel} within each patch we obtain 
\begin{align}
K(x,x') &= \frac{2}{\pi N_w} \sum_i \sin^{-1} \left[\frac{{x}^T_i \Sigma {x}'_i}{\sqrt{1+2 {x}_i^T \Sigma {x}_i}\sqrt{1+2 [{x}'_i]^T \Sigma {x}'_i}} \right] 
\end{align}
where, again due to $O(S-1)$ symmetry, $\Sigma = c_{\perp}I + (c_*-c_{\perp}) {w}^* [{w}^*]^T$. The above is a sum of 2-layer FCN Erf network NNGP kernel acting on each patch. 

\begin{figure*}[t]
\vskip -0.8in
\begin{centering}
\includegraphics[width=1\textwidth]{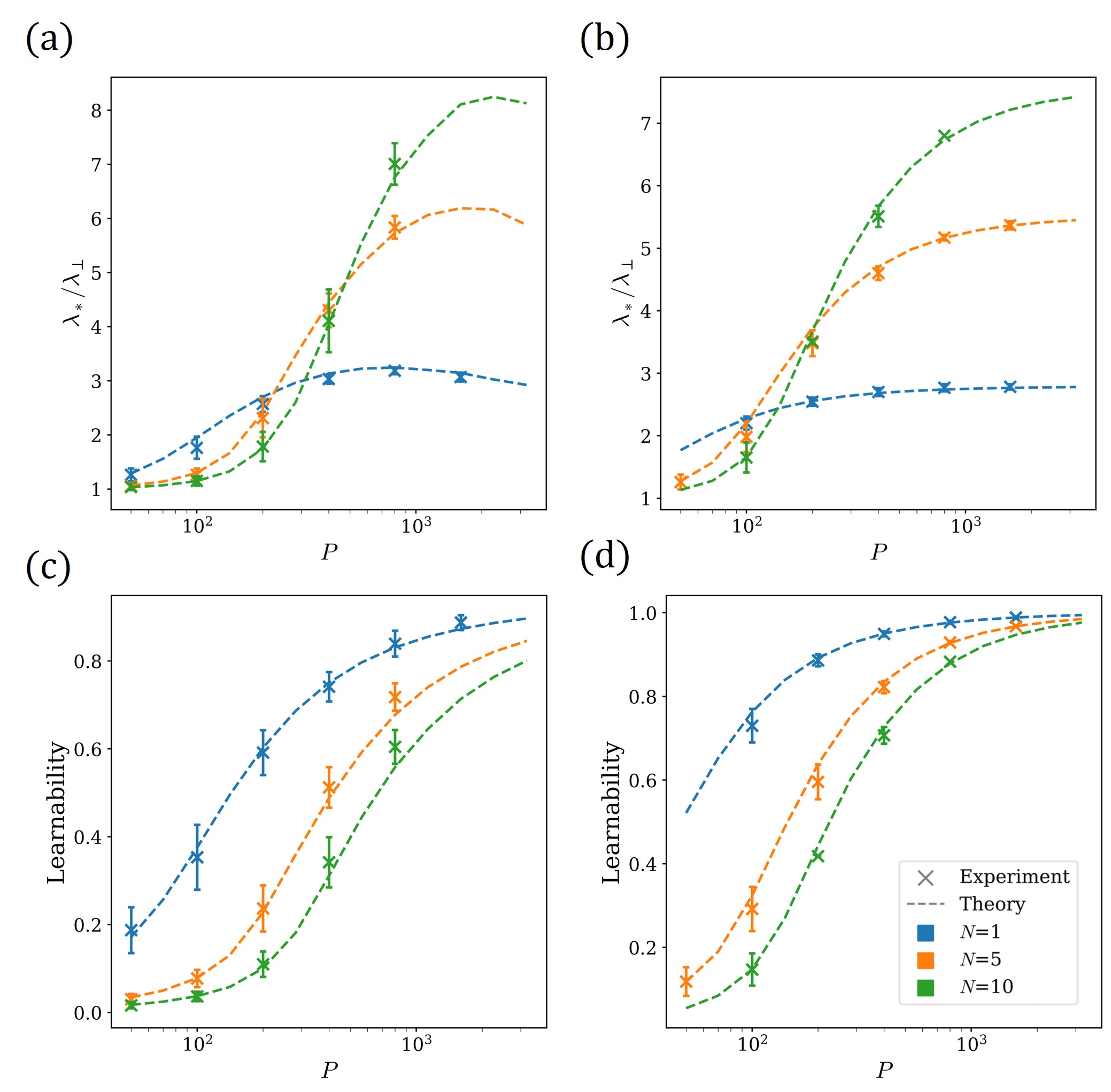}
\caption{In this figure we compare a linear network trained on a single index linear teacher, with an Erf network trained on a cubic single index teacher ($y(x)=w_* \cdot x +0.1 H_3(w_*\cdot x)$, where $H_3$ is the third Hermite polynomial). The ratio between the teacher direction eigenvalue of the kernel to the eigenvalues corresponding to orthogonal directions for the Erf and linear networks is shown in panels (a) and (b) respectively. In panels (c), (d) the learnability ($f\cdot y/y\cdot y$) is shown for the Erf and linear network respectively.  Network parameters:
$\chi=100$, $N_w=1,5,10,S=50,C=1000$.}
\label{fig:amp_fact_cnn}
\par\end{centering}
\vskip -0.2in
\end{figure*}

To obtain an equation for $c_{\perp},c_*$, what remains is to diagonalize the above kernel in order to obtain the discrepancy ($\bar{\tilde{f}}_a(x)$) given $\Sigma$, which, as assumed earlier, would be proportional to $y(x)$. Since these patch-wise kernels all commute under our Gaussian data measure, we may diagonalize each one separately and reconstruct the full spectrum. Focusing on a single patch or equivalently on the $N=1$ case, we note that we may undo the Gaussian average over ${w}$ and combine it with the previous integral to obtain 
\begin{align}
&\int \dif x'_1 {\cal N}(0,I_{S \times S};x'_1) \frac{2}{\pi} \sin^{-1} \left[\frac{{x}^T_1 \Sigma {x}'_1}{\sqrt{1+2 {x}_1^T \Sigma {x}_1}\sqrt{1+2 [{x}'_1]^T \Sigma {x}'_1}} \right] {v} \cdot x'_1 \\ \nonumber 
&= \int \dif x'_1 {\cal N}(0,I_{S \times S};x'_1)\int \dif {w} {\cal N}(0,\Sigma;w) \Erf({w} \cdot x_1)\Erf({w} \cdot x'_1) { v}\cdot x'_1 \\ \nonumber 
&= \int \dif {w} {\cal N}(0,\Sigma;w) \Erf({w} \cdot x_1)\frac{2}{\sqrt{\pi}} \frac{{w} \cdot { v}}{\sqrt{1+2|{w}|^2}} = ... 
\end{align}
which apart from having $\Sigma \neq I_{S \times S}$ is of the same form as that in our Erf integral. Redefining ${w}'=\sqrt{\Sigma}^{-1} w$, where $\sqrt{\Sigma}$ is the matrix having the same eigenvectors as $\Sigma$ but with square root eigenvalues, we have that ${w}' \sim {\cal N}(0,I_{S \times S})$, consequently,  
\begin{align}
... &= \int \dif {w}' {\cal N}(0,I_{S\times S};{w}') \Erf({w}' \cdot [\sqrt{\Sigma} x_1]) \frac{2}{\sqrt{\pi}} \frac{{w}' \cdot [\sqrt{\Sigma} { v}]}{\sqrt{1+2{w'}^T \Sigma {w}'}} \\ \nonumber 
&\approx \int \dif{w}' {\cal N}(0,I_{S\times S};{w}') \Erf({w}' \cdot [\sqrt{\Sigma} x_1]) \frac{2}{\sqrt{\pi}} \frac{{w}' \cdot [\sqrt{\Sigma}{ v}]}{\sqrt{1+2\Tr[\Sigma]}}=...
\end{align}
where we used again the mean-field approximation on $|{w}|^2=[{w}']^T \Sigma {w'}\approx \Tr[\Sigma]$ which is justified provided $\Sigma$ does not have a dominating (larger by an $O(S)$ factor) eigenvalue. We may now readily carry out our Erf integral and use again our mean-field treatment of norms to obtain 
\begin{align}
... &= \frac{4}{3\pi \sqrt{1+2\Tr[\Sigma]}\sqrt{1+2x_1^T \Sigma x_1}} {v}^T \Sigma x_1 \approx \frac{4}{3\pi(1+2\Tr[\Sigma])} {v}^T \Sigma x_1  
\end{align}
We thus arrive at the simple conclusion that at large $S$, ${v} \cdot x_1$ with ${v}$ chosen along eigenvectors of $\Sigma$ are approximate eigenfunctions of the kernel. Thus we again find, like in the linear case, that $\bar{\tilde{f}}_a(x) \propto y(x)$ and hence $A(x,x')\propto y(x)y(x')$, only we a slightly different proportionality factor compared to the linear case.  Solving the equations for $c_*$ numerically, we indeed observe that the amplification factor (namely $\lambda_*/\lambda_{\perp}$) as well as the learnability only minorly differs between the linear and non-linear networks, as evident in Fig. \ref{fig:amp_fact_cnn}.  

Thus, apart from some $O(1)$ factors, we obtain the same result as in the linear case for $\chi \gg 1$ and expect an $P=O(N\sqrt{S})$ complexity. Notably, since the outlier in $\Sigma$ has eigenvalue $c_* \propto 1/\sqrt{S}$ our mean-field replacement of $|w|^2$ and $x_1^T \Sigma x_1$ are justified at large $S$ and in fact $\Tr[\Sigma]\approx 1$. However, taking $\chi = O(N)$, we previously found $c_* \propto 1$ which implies that the contribution of $w$ fluctuations along $w^*$ is as dominant as those along all other directions combined. Failing this consistency check means we may only assume the norm is self-averaging along directions orthogonal to the target. If so one may try the approximation $|w|^2 \approx (S-1) c_{\perp}+(w \cdot {w}^*)^2=\frac{S-1}{S}+(w \cdot {w}^*)^2$ and $x_1^T \Sigma x_1 = \frac{S-1}{S}+c_* (x_1 \cdot {w}^*)^2$, complicating the diagonalization process in general due to creation of ``harmonics'' or powers of $y(x)$ in $A(x,x')$. We can control this extra complication by softening the mean-field scaling by taking, say, $\chi = N/100$, leading to a similar sample complexity class as in the linear case.

\subsubsection{Some relations between Kernel-Scaling and Kernel-Adaptation}
Finally, we establish some similarities and distinctions between the kernel-scaling and kernel-adaptation approaches. The technique we refer to as kernel-scaling was first derived in Ref. \cite{LiSompolinsky2021}, for deep linear networks and was called ``The Backpropagating Kernel Renormalization''. The latter is exact in the scaling limit $P\propto N_w \propto d \rightarrow \infty$, for fixed depth \citep{Hanin2023}. The kernel-adaptation approach is only accurate up to $1/\chi$ corrections, due to the mean-field decoupling of the last layer (though often quite accurate even for moderate $\chi$, see \cite{rubin2024a}). Still, for mean-field scaling ($\chi=O(N)$) it also becomes exact. In fact, using an on-data formulation of the kernel-adaptation approach \cite{rubin2025kernels}, one can show that the equations for the predictor coming from both these approaches coincide. 

Turning to non-linear FCNs, we recall that Ref. \cite{ariosto2022statistical} extended ``The Backpropagating Kernel Renormalization'' by approximating $\int \dif \mu_x \tilde{f}(x) \sigma(w^{(0)} \cdot x)$ 
in Eq. \ref{Eq:S_FCN2} as a Gaussian variable. This is exact for linear networks \footnote{Technically, they did so in an on-data formulation whereas we work here with a data-averaged formulation, but it is the same approximation.}, and provides a good approximation when the summand is weakly correlated in the $\mu$ index. In standard scaling and iid data, this approach appears asymptotically exact in harmony with the fact that FCNs in standard scaling and such data behave effectively as linear networks \citep{Cui2023}. However, going to mean-field scaling, finite FCNs are in a different sample complexity class than their GP counterparts, via, for instance, the aforementioned staircase mechanisms \citep{abbe2021staircasepropertyhierarchicalstructure,Paccolat_2021}. The kernel-scaling approximation, which leaves us in the infinite-width/GP sample complexity class cannot accommodate such effects. 

At least for FCNs with weakly non-linear target functions and for CNNs (as shown above, see also Ref. \cite{naveh2021self}), kernel-adaptation can accurately predict this change in sample complexity class, limited however, to large $\chi$. 

Kernel-scaling has been adapted to CNNs \citep{RotondoCNN1,bassetti2024featurelearningfinitewidthbayesian}, at the price of having a vector order parameter associated with the scaling of each convolutional patch, however, it remains to be seen whether this extension is capable of predicting changes to sample complexity, as shown in Sec. \ref{Sec:KernelAda}.

\subsubsection{Equivalence of kernel scaling and kernel adaptation in limited settings}
Consider a two-layer linear neural network ($f(x) = \sum_c a_c (w^T_c x)$), trained to equilibrium using Langevin dynamics with mean-field scaling. Specifically, we choose the posterior covariance with no data (``the prior'') to have weight variances $\sigma_{\rm a}^2=N^{-1} \chi^{-1},\sigma_{\rm w}^2=d^{-1}$ where $\chi \gg 1$. 

As shown in Ref. \cite{seroussi2023separation} following same approximations (mean-field decoupling between layers, and variational Gaussian approximation on the resulting decoupled layer-wise actions) \footnote{Specifically, Eq. (8) and noting that for a linear network, in that notation, $G(..)=..$, $\frac{\partial [Q_f]_{\mu \nu}}{\partial \Sigma_{ss'}}=[x_{\mu}]_s [x_{\nu}]_{s'}$} the equation for the average discrepancy vector on the training set ($t \in {\rm R}^P$, analogous to $\bar{\tilde{f}}$ in our data-averaged formalism)  is 
\begin{align}
t = \left[[K^{-1} - \frac{\chi}{N} t t^T]^{-1}+(\kappa^2/P) I\right]^{-1} y
\end{align}
This can be understood as the standard GPR equation for the train discrepancy with $[K^{-1} - \frac{\chi}{N} t t^T]^{-1}$ playing the role of the (adapted) kernel. Noting that $w^{*} \cdot x_{\mu}=y_{\mu}$ and that, when we average over data, due to symmetry, $ t \propto y$ and $K^{-1} \propto (x \cdot x')$ on the space of linear functions, we find that this discrete kernel is consistent with Eq. \ref{Eq:AdaptedKernel1}.

Performing a Woodbury identity on the EoS yielding the equivalent form 
\begin{align}
\label{Eq:EoS}
t = \left[K+\frac{\chi}{N}[1-(\chi/N)t^T K t]^{-1} Ktt^TK +(\kappa^2/P) I\right]^{-1} y
\end{align}

Multiplying both sides by $\left[K+\frac{\chi}{N}[1-(\chi/N)t^T K t]^{-1} Ktt^TK +(\kappa^2/P) I\right]$ one finds 
\begin{align}
\left[K+\frac{\chi}{N}[1-(\chi/N)t^T K t]^{-1} Ktt^TK +(\kappa^2/P) I\right] t &= y \\ \nonumber 
K t +\frac{\chi}{N}[1-(\chi/N)t^T K t]^{-1} Ktt^TK t +(\kappa^2/P) t &= y \\ \nonumber 
K t\left[1 +\frac{\chi}{N}[1-(\chi/N)t^T K t]^{-1} t^TK t\right] + (\kappa^2/P) t &= y \\ \nonumber 
K t\left[\frac{1-(\chi/N)t^T K t+\frac{\chi}{N}t^TK t}{1-(\chi/N)t^T Kt}\right] + (\kappa^2/P) t &= y \\ \nonumber
K t\left[\frac{1}{1-(\chi/N)t^T Kt}\right] + (\kappa^2/P) t &= y \\ \nonumber
[\bar{Q} K+(\kappa^2/P)]t &= y \\ \nonumber
t = [\bar{Q} K+(\kappa^2/P)&]^{-1}y \\ \nonumber
\end{align}
where $\bar{Q}=[1-(\chi/N)t^T K t]^{-1}$, hence we obtain the on-data version of the kernel scaling interpretation only without the fluctuation contribution to $\bar{Q}$. Indeed taking a saddle on Eq. (7) of Ref. \citep{ariosto2022statistical} (which agrees with \citep{LiSompolinsky2021} for linear networks) and omitting the $Tr\log$ term associated with GP fluctuations, as it is negligible in the mean-field limit, one arrives at the same result. 

Turning to standard scaling, at least for FCN, this fluctuation correction is non-negligible. Interestingly, including leading saddle point corrections to kernel adaptation, as done in Ref. \cite{rubin2025kernels} accounts for much of its effect. However, while kernel adaptation can be adapted to this FCN + standard scaling setting, it is much more natural to use kernel scaling. Turning to regimes where FCN and CNNs perform better than their GP counterparts, kernel adaptation and its variants which replace the variational Gaussian approximation with a variational Gaussian mixture approximation \citep{rubin2024grokking}-- become more adequate. Still, there is much to be understood in the phase diagram of feature learning. For instance, regimes at which strong channel/channel or width-index/width-index correlations appear as one approaches the underparametrized regime \citep{bricken2023monosemanticity}.

%% file: Chapters/Field_theory_approach_to_dynamics.tex
\chapter{Field theory approach to dynamics}
\label{Sec:MSRDJ}
So far, our focus has been on equilibrium/ infinite-training-time/Bayesian aspects of deep learning. In physics, the equilibrium state is typically more universal and tractable than out-of-equilibrium ones. It also serves as a starting point for various near-equilibrium approaches, making it a sensible scientific starting point. Notwithstanding, sometimes one is interested in dynamical properties. Several relevant examples are Grokking \citep{power2022grokking,rubin2024grokking} where the network goes from a memorization phase to a generalization phase after a long, nearly steady-state behaviour or the issue of early stopping, where some networks show better performance at finite training time rather than in their equilibrium state. 

Conveniently, via field-theoretical approaches to dynamics, the transition from equilibrium to dynamics does not come at a big formal cost. As shown below, much of our formalism carries through to dynamics at the cost of introducing additional time “indices” to equilibrium quantities. One may find field-theoretical approaches to dynamics in the works of \citep{bordelon2022self, Bordelon2023Dynamics, mignacco2020dynamical}. Bordelon et. al. use the MSRDJ formalism to identify, for given preactivation and gradients, an effective time-dependent Gaussian process which characterize the noise due to initialization and the Langevin dynamics noise. They then use this noise distribution in an integral equation governing the preactivations and outputs. This leads to a set of self-consistency equations ensuring that the noise distribution results in the same preactivations it was predicated upon. 
The resulting set of self-consistency equations differs from those derived later in this chapter and characterizes the dynamics in terms of a different and larger set of order parameters. Mignacco et al. focus on networks with a single trainable layer and Gaussian input data, analyzing dynamics under actual stochastic gradient descent (SGD). Our approach diverges by focusing on the stochastic updating process driven by Langevin dynamics, and focusing on the dynamics of more than one trainable layer. 

Below we take the modest task of demonstrating this technique on a two layer network and tracking the transition from NTK-type dynamics to NNGP equilibrium behaviour \citep{avidan2023connectingntknngpunified}. As far as kernel adaptation and feature learning goes, we would only point to similarities with previous Bayesian actions. We further note by passing that the MSRDJ formalism fits perfectly with data averaging as the MSRDJ partition function is constant in the data and hence does not require any introduction of replicas (e.g. \cite{lindner2023theory}).

\section[MSRDJ for one hidden layer]{MSRDJ field theory for a one hidden layer non-linear neural network}
\label{chap:MSR main}
In the following section, we describe the dynamics of a single hidden layer non-linear fully connected network (FCN) using the MSRDJ formalism. Extensions to deeper networks are straightforward. Besides deriving the MSRDJ partition function, we further obtain an integral equation for the dynamics of the mean discrepancy of the network's output in the infinite width limit. We further show how this integral equation recovers as a limiting case both the dynamics governed by the NN-NTK mapping \citep{lee2019wide} and the equilibrium distribution governed by the NNGP mapping \citep{lee2017deep}.

\section{Problem setup}
We consider a two-layer FCN similar to the one defined in Equation \ref{Eq:2_layer_FCN} while keeping track of the training-time dependence of the fields.

\begin{equation}
f(t;x_\mu) =  \sum_{c=1}^{C}\frac{1}{\sqrt{C}} a_{c}(t) \sigma\left(\frac{1}{\sqrt{S}}{w}_c(t) \cdot {x}_{\mu}\right) 
\end{equation}

 In the spirit of kernel adaptation (see Section \ref{Sec:KernelAda}) we consider the FCN to be of width $C$ and the dataset to consists of $\{ {x}_{\mu} \}_{\mu=1}^{P}$ where ${w}_c, {x}_{\mu} \in {\R}^{S}$. In section \ref{Sec:KernelAda}, we scale $a_{ic}$ and $w_c$ such that their variance is of the order of $\frac{1}{NC}$ and $\frac{1}{S}$ respectively. However, unlike section \ref{Sec:KernelAda}, it is preferable here to explicitly scale the network output while taking the variance of $a_c(t)$ and $w_c(t)$ at initialization to be of order of $\frac{1}{\chi}$ and $O(1)$ respectively. This approach facilitates tracking terms that vanish at infinite width. 
We take the loss function
\begin{gather}
    \cL(t) =\cL_{\MSE}(t)+\frac{1}{2}\frac{\kappa^2 }{\sigma_{\rm a}^2 \eta}  a^2(t) +\frac{1}{2}\frac{\kappa^2 }{\sigma_{\rm w}^2\chi\eta} w^2(t)\\
    \cL_{\MSE}(t)=\frac{1}{2} \sum\limits_{\mu=1}^P (f(t;{x}_\mu)-y_\mu)^2
\end{gather}
and consider the time-continuous Langevin dynamics where the network weights evolve under noisy gradient descent with respect to the aforementioned loss:
\begin{gather} \label{dynamics_TB}
    \frac{\dif a_c }{\dif t}=-\eta \partial_{a_c}\cL(t)+{\xi}^{\rm a}_{c}(t) \\  \nonumber
    \frac{\dif w_{ci} }{\dif t}=-\eta \partial_{w_{ci}}\cL(t)+{\xi}^{\rm w}_{ci}(t)
\end{gather}
where $\xi^{\rm a}\in \R^C, \xi^{\rm w}\in \R^{C\times S}$ are some additive Gaussian noise with the following statistics,
\begin{equation}
    \langle \xi^{\rm a}_c \rangle=0, \langle \xi^{\rm a}_{c}(t) \xi^{\rm a}_{c'}(t')\rangle=\frac{\eta\kappa^2}{\chi}\delta_{cc'}\delta(t-t')
\end{equation}
and
\begin{equation}
    \langle \xi^{\rm w}_{ci}\rangle=0,\langle \xi^{\rm w}_{ci}(t) \xi^{\rm w}_{c'i'}(t')\rangle=\frac{\eta\kappa^2}{\chi} \delta_{cc'}\delta_{ii'}\delta(t-t').
\end{equation}
We initialize $a_c(0)\sim\cN(0,\frac{\sigma_{\rm a}^2}{\chi})$ and $w_{ci}(0)\sim\cN(0,\sigma_{\rm w}^2)$ with both $\sigma_{\rm a},\sigma_{\rm w}=O(1)$. 

\section{MSRDJ - User interface}
Field theory and stochastic differential equations (SDEs) both examine distributions of functions and, thus, are naturally connected. The MSRDJ formalism links them by interpreting an SDE like $\bm{L}[\phi]=\chi(t)$, where $\bm{L}[\phi]$ is a possibly non-linear differential operator and $\chi(t)$ is white noise, as a functional delta-function. Given the comprehensive reviews available (e.g. \cite{MoritzBook}), we concentrate on offering a minimal ``user interface'', notably on crafting the field theory related to neural network training. To this end, consider an SDE of the form:
\begin{gather} \label{consider this sde}
\frac{\dif \phi(t)}{\dif t} = f(\phi(t)) +\chi(t), \\  \nonumber
\phi(0^+)=\phi_0 \quad \quad \phi_0\sim P_0 \\  \nonumber
\langle\chi(t)\rangle=0\quad\quad\langle\chi(t)\chi(t')\rangle=\delta(t-t')\sigma^2
\end{gather}
where $f$ is a deterministic non-linear function typically related to the loss function, and $P_0$ is the initial distribution of $\phi(t)$. This SDE is framed as the limit $h\rightarrow 0$ of the dynamics on a discrete time lattice $\phi_n=x(t_n)$,  $t_n=n h$ where $n\in\{1,\dots,M\}$ and $t_0 = 0$.
Under the Itô convention, the drift term depends only on the previous state $\phi_{n-1}$, granting the discrete update rule:
\begin{equation}\label{Ito difference equation}
\phi_{n} = \phi_{n-1} + f(\phi_{n-1}) h +\chi_n.
\end{equation}
We can express the joint probability distribution of the set $\{\phi_n\}$, i.e., a discrete path, by taking the average of the product of delta functions, which enforce the update rule at each discrete time step. This average can be written as:
\begin{align}
P[\{ \phi_n \}_{n=1}^M|\phi_0] = &\prod\limits_{n=1}^M  \int_{-\infty}^{\infty} \frac{\dif \chi_n}{\sqrt{2\pi \sigma^2 h}} \exp\left( -\frac{(\chi_n)^2}{2 \sigma^2 h} \right)\int_{-i\infty}^{i\infty} \frac{\dif \tilde{\phi}_n}{2\pi i}\notag\\&\exp\left( \tilde{\phi}_n\left[ \phi_n-\phi_{n-1} - f(\phi_{n-1} )h - \chi_n \right] \right)
\end{align}
then, by integrating out the noise and in the continuum limit (i.e. $h\rightarrow 0$) the probability for a path $\phi(t)$ is
\begin{gather}
P[\phi(t)|\phi(0^+)] = \int \cD_{2\pi i}{\tilde{\phi}}\ \  e^{-S} \notag\\
S=\int\limits_0^t  -\frac{\sigma^2}{2}\tilde{\phi}(t')^2-\tilde{\phi}(t')\left[\partial_{t'}\phi(t')-f(\phi(t'))\right]\dif t'
\end{gather}
where we define $\int\cD_{2\pi i}\tilde{\phi} = \lim\limits_{h\rightarrow 0}\prod\limits_{n=1}^M\int\limits_{-i\infty}^{i\infty}\frac{d\tilde{\phi}_n}{2\pi i}$ and the partition function under MSRDJ becomes
\begin{gather}
    Z=\int \cD{\phi\ \cD_{2\pi i}\tilde{\phi}}\ d\phi(0^+)\ e^{-S+\log(P(\phi(0^+))} \nonumber\\
    S=\int\limits_0^t - \frac{\sigma^2}{2}\tilde{\phi}(t')^2-\tilde{\phi}(t')\left[\partial_{t'}\phi(t')-f(\phi(t'))\right]\dif t'.
\end{gather}
Thus, to apply MSRDJ to an SDE, one transforms $\frac{d\phi(t)}{\dif t} = f(\phi(t)) +\chi(t)$ into $S=\int\limits_0^t - \frac{\sigma^2}{2}\tilde{\phi}(t')^2-\tilde{\phi}(t')\left[\partial_{t'}\phi(t')-f(\phi(t'))\right]\dif t'$.

\section{Condensed Notation}
It is useful to introduce the following notation. Consider $x,y$ as the time vectors, whose entries are $x(t),y(t)$ then we define their dot product as $x^T y=\int_0^t x(t')y(t') \dif t'$. In this convention, the transpose acts on both the time index and the vector index for vectors and matrices (vector index being the index in the normal sense), i.e. $x^Ty = \int_{0}^t \sum\limits_{i}  x_i(t') y_i(t') \dif t'$.
As another shorthand we shall use the Hadamard product $\odot$ to act time-wise, meaning $M(t,s)N(t,s)=\bm{M}\odot\bm{N}$.

\section{MSRDJ partition function}
In condensed notation the MSRDJ partition function corresponding to equations \ref{dynamics_TB} reads as
 
\begin{align}
Z=
\int_{\tilde{w},w,\tilde{a},a} 
e^{-S-S_{\text{initial}}}\notag
\end{align}
Where
\begin{align}
&S=
\sum\limits_c\left(-\frac{\eta\kappa^2}{2\chi}\tilde{a}_{c}^T\tilde{a}_{c} 
- \tilde{a}_{c}^T[\bm{L}_{\rm a} a_{c} 
+ \eta \partial_{a_c} \cL_{MSE}]\right) \\ \nonumber
&+\sum\limits_{ci}\left(- \frac{\eta\kappa^2}{2\chi} \tilde{w}_{ci}^T \tilde{w}_{ci} 
- \tilde{w}_{ci}^T [\bm{L}_{\rm w} w_{ci}+\eta \partial_{w_{ci}} \cL_{MSE}] \right)
\end{align}
and
\begin{equation}
S_{\text{initial}} = \sum\limits_c\frac{\chi}{2\sigma_{\rm a}^2} a_c(0)^2+\sum\limits_{ci}\frac{1}{2\sigma_{\rm w}^2}w_{ci}(0)^2
\end{equation}

We define $\bm{L}_{\rm a}\equiv\partial_t+\frac{\eta \kappa^2}{\sigma_{\rm a}^2}$ and $\bm{L}_{\rm w}\equiv\partial_t+\frac{\eta\kappa^2}{\sigma_{\rm w}^2\chi}$ .

It should be noted that the action resulting from the MSRDJ formalism is highly non-linear as it contains terms such as $\tilde{a}^T\sigma(w\cdot x_\mu)$ with $\sigma$ being a general non-linear function. To elevate some of this difficulty, we draw upon the strategy used in works such as  \cite{Jacot2018}, where significant progress was made by shifting the description of the network from its weights to its output or output discrepancy. The success of this transition suggests that the network output provides a more natural and fundamental representation of the system's dynamics, similar to the use of generalized coordinates in classical mechanics.
Following their example, we adopt a description based on network output discrepancy.

We define $f_\mu(t)=\sum\limits_{c=1}^{C}a_{c}(t) \frac{1}{\sqrt{C}}\sigma( \frac{1}{\sqrt{S}}w_c(t)\cdot x_{\mu})$. We incorporate $f$ into the action, by introducing the identity $1=\prod\limits_{\mu=1}^{P}\int \cD f_\mu \delta(f_\mu(t)-\sum\limits_{c=1}^{C}a_{c}(t) \frac{1}{\sqrt{C}}\sigma( \frac{1}{\sqrt{S}}w_c(t)\cdot x_{\mu}))$, which enforces the definition of $f_\mu(t)$. This has the effect of introducing $\tilde f_\mu(t)$ the auxiliary output field. 
Finally, using $\frac{\partial \cL(t)}{\partial f(t;x_\mu)}=f_\mu(t)-y_\mu$ and applying the chain rule we obtain:

\begin{align}    
&Z=
\int_{\tilde{w},w,\tilde{a},a,\tilde{f},f} 
e^{-S-S_{\text{initial}}}  \\ \nonumber
&S=
-\frac{\eta\kappa^2}{2\chi}\sum\limits_{c}\tilde{a}_{c}^T\tilde{a}_{c} 
-\sum\limits_{c}\tilde{a}_{c}^T\bm{L}_{\rm a} a_{c}
-\frac{\eta\kappa^2}{2\chi} \sum\limits_{c,i}\tilde{w}_{ci}^T \tilde{w}_{ci} -\sum\limits_{c,i} \tilde{w}_{ci}^T \bm{L}_{\rm w} w_{ci}\\ \nonumber
&-\sum\limits_{c,\mu}\tilde{a}_{c}^T[\eta f_{\mu} \frac{1}{\sqrt{C}}\sigma(\frac{1}{\sqrt{S}}w_{c}\cdot x_{\mu})-\eta y_{\mu}\frac{1}{\sqrt{C}}\sigma(\frac{1}{\sqrt{S}}w_{c}\cdot x_{\mu})] \\ \nonumber
&-\sum\limits_{c,i,\mu}\tilde{w}_{ci}^T [\eta f_{\mu} a_{c} \frac{1}{\sqrt{CS}}\sigma'(\frac{1}{\sqrt{S}}  w_{ci} x_{\mu;i}) x_{\mu;i} -\eta y_{\mu} a_{c} \frac{1}{\sqrt{CS}}\sigma'(\frac{1}{\sqrt{S}} w_{ci} x_{\mu;i}) x_{\mu;i}]\\ \nonumber
&-\sum\limits_{\mu}\tilde{f}_{\mu}^T[f_{\mu}- \sum\limits_{c}a_{c} \frac{1}{\sqrt{C}}\sigma( \frac{1}{\sqrt{S}}w_c\cdot x_{\mu})]
\end{align}
This action is quadratic in the fields $\tilde{a},a$, which allows us to integrate out these fields. The term $\exp(- \sum\limits_c\frac{\chi}{2\sigma_{\rm a}^2} a_c(0)^2)$ induces an initial condition on the time correlation (propagator) of $a_c$, thus determining it uniquely:
$a_{c}\sim \cN(\bar{a}_{c},\bm{\Sigma})$ where $\bar{a}_{c}(t)=-\eta \int\limits_0^t   e^{-\frac{\eta\kappa^2}{\sigma_{\rm a}^2}(t-t')} \sum\limits_{\nu=1}^{P} \frac{1}{\sqrt{CS}}\sigma(\frac{1}{\sqrt{S}}w_{c}(t') \cdot x_{\nu}) (f_{\nu}(t')-y_\nu) \dif t'$ and $\bm{\Sigma}(t,t')=\frac{\sigma_{\rm a}^2}{2\chi} e^{-\frac{\eta\kappa^2}{\sigma_{\rm a}^2}|t-t'|}$.
Leaving us with the following expression for the MSRDJ partition function

\begin{align}\label{Eq:S_f_w_MSRDJ}
    &Z=
    \int_{\tilde{w},w,\tilde{f},f}e^{-S-\sum\limits_{ci}\frac{1}{2\sigma_{\rm w}^2}w_{ci}(0)^2} \\ \nonumber
    &S=-\frac{\eta\kappa^2}{2\chi}\sum\limits_{c,i} \tilde{w}_{ci}^T \tilde{w}_{ci} -\sum\limits_{c,i} \tilde{w}_{ci}^T \bm{L}_{\rm w} w_{ci} \\ \nonumber
   & - \sum\limits_{c,i,\mu}\tilde{w}_{ci}^T [\eta (f_{\mu}-y_\mu) \bar{a}_{c} \frac{1}{\sqrt{CS}}\sigma'(\frac{1}{\sqrt{S}} w_{ci} x_{\mu;i}) x_{\mu;i}] \\ \nonumber
    &-\sum\limits_{c,\mu}\tilde{f}_{\mu}^T[f_{\mu}-\bar{a}_c \frac{1}{\sqrt{C}}{\sigma}(\frac{1}{\sqrt{S}} w_c\cdot x_{\mu})] \\ \nonumber
    &-\sum\limits_{c,\mu,\nu}\Bigg[-\tilde{f}_{\mu}^T
  \frac{1}{\sqrt{C}}\sigma^T(\frac{1}{\sqrt{S}} w_c\cdot x_{\mu})
+\\ \nonumber&
\sum\limits_{i}\tilde{w}_{ci}^T
\eta (f_{\mu}-y_\mu)^T \frac{1}{\sqrt{CS}}\sigma'^T(\frac{1}{\sqrt{S}} w_{ci} x_{\mu;i}) x_{\mu;i}\Bigg] 
\bm{\Sigma}\\ \nonumber
&\Bigg[-  \frac{1}{\sqrt{C}}\sigma(\frac{1}{\sqrt{S}} w_c\cdot x_{\nu}) \tilde{f}_{\nu}
+
\sum\limits_{j}\tilde{w}_{cj}
\eta  \frac{1}{\sqrt{CS}} \sigma'( \frac{1}{\sqrt{S}} w_{cj} x_{\nu;j}) (f_{\nu}-y_\nu) x_{\nu;j}\Bigg] 
\end{align}
At this juncture, we draw the reader's attention to the resemblances between Equation \ref{Eq:S_FCN2_nonEK} and Equation \ref{Eq:S_f_w_MSRDJ}. Both equations encompass terms like $\sum_\mu(\tilde{f}_\mu\sigma(w_c\cdot x_\mu))^2$ or, alternatively, in the limit where sampling the discrete dataset transitions to integrating over the data measure, expressed as $\int d\mu_x(\tilde{f}(x)\sigma(w_c\cdot x))^2$. However, the key distinction lies in the presence of differential operators in Equation \ref{Eq:S_f_w_MSRDJ}, which act on the temporal indices of $f$ and $w$.
\section{Infinite width limit}
Our next goal is to study the infinite width limit of the above action and retrieve the NTK and NNGP behavior. At this point, we turn to two key approximations, enabling the integration of the $\tilde{w},w$ fields.
The first comes by when focusing on the following part of the action.
$S_{\rm w}=-\frac{\eta\kappa^2}{2\chi} \tilde{w}_{ci}^T \tilde{w}_{ci} 
    - \tilde{w}_{ci}^T [\bm{L}_{\rm w} w_{ci}+\eta (f_{\mu}-y_\mu) \bar{a}_{c} \frac{1}{\sqrt{CS}}\sigma'(\frac{1}{\sqrt{S}} w_{ci} x_{\mu;i}) x_{\mu;i}] -\sum\limits_{ci}\frac{1}{2\sigma_{\rm w}^2}w_{ci}(0)^2$
which is composed of two contributions to the evolution of the $w$ field, one $\tilde{w}_{ci}^T\bm{L}_{\rm w} w_{ci}$ and the other $\tilde{w}_{ci}^T\eta (f_{\mu}-y_\mu) \bar{a}_{c} \frac{1}{\sqrt{CS}} \sigma'(\frac{1}{\sqrt{S}} w_{ci} x_{\mu;i}) x_{\mu;i}$. In this second contribution, the field $\tilde{w}_{ci}$ experiences interactions that are suppressed by $\frac{1}{C}$ which is apparent by $\bar{a}_c\sim\frac{1}{\sqrt{C}}$, which in the infinite width limit become negligible.
This simplification leaves us with:
\begin{equation}
    S_{\rm w}=\sum\limits_{c,i}\left(\frac{\eta\kappa^2}{2\chi} \tilde{w}_{ci}^T \tilde{w}_{ci} + \tilde{w}_{ci}^T \bm{L}_{\rm w} w_{ci}-\frac{1}{2\sigma_{\rm w}^2}w_{ci}(0)^2\right)
\end{equation}

And the partition function
\begin{align}
    &Z=
\left\langle
    \int_{\tilde{f},f}
    e^{S_f} 
    \right\rangle_{S_{\rm w}}
   \\ \nonumber
   &S_f=\sum\limits_{\mu}\tilde{f}_{\mu}^T[f_{\mu}-\frac{1}{\sqrt{C}} \bar{a}_c\cdot \sigma(\frac{1}{\sqrt{S}} w_c\cdot x_{\mu})] 
   \\ \nonumber&+\!\sum\limits_{c,\mu,\nu}\left[\!-\tilde{f}_{\mu}^T
\frac{1}{\sqrt{C}}  \sigma^T(\frac{1}{\sqrt{S}} w_c\cdot x_{\mu})
+
\sum\limits_{i}\tilde{w}_{ci}^T
\eta (f_{\mu}-y_\mu)^T\frac{1}{\sqrt{CS}} \sigma'^T(\frac{1}{\sqrt{S}} w_{ci} x_{\mu;i}) x_{\mu;i}\!\right] 
\\ \nonumber &\bm{\Sigma}
\left[-\frac{1}{\sqrt{C}}  \sigma(\frac{1}{\sqrt{S}} w_c\cdot x_{\nu}) \tilde{f}_{\nu}
+
\sum\limits_{j}\tilde{w}_{cj}
\eta  \frac{1}{\sqrt{CS}} \sigma'(\frac{1}{\sqrt{S}} w_{cj} x_{\nu;j}) (f_{\nu}-y_\nu) x_{\nu}\right]
\end{align}
Here, the second approximation arises. We apply cumulant expansion for $S_f$ under $e^{-S_{\rm w}}$, keeping order zero terms in $\frac{1}{C}$, we obtain the following:
\begin{gather}
Z=\int_{\tilde{f},f}
    e^{-S_{f}}
    \\ \nonumber
    S_{f}=\sum\limits_{\mu,\nu}\bigg(\!-\tilde{f}_{\mu}^T[f_{\mu}+
    \eta(\bm{L}^{-1}_{\rm a}\odot\bm{\Phi}(x_\mu,x_\nu))f_{\nu}-\eta(\bm{L}^{-1}_{\rm a}\odot\bm{\Phi}(x_\mu,x_\nu))y_{\nu}]\bigg.\\ \nonumber
     -\eta (f_{\mu}-y_\mu)^T(\bm{\Sigma}_{\rm a}\odot\bm{L}^{-1}_{\rm w}\odot\bm{\Phi}'(x_\mu,x_\nu) K^x({x_\mu,x_\nu})) \tilde{f}_{\nu}
     \\ \nonumber\bigg.\!- \tilde{f}_{\mu}^T (\bm{\Sigma}_{\rm a}\odot\bm{\Phi}(x_\mu,x_\nu)) \tilde{f}_{\nu}\bigg)
\end{gather}
where
\begin{equation}
    \bm{\Phi}(t,t';x_\mu,x_\nu)=\langle\sum\limits_{c,i,j}\frac{1}{C}\sigma(\frac{1}{\sqrt{S}}w_{ci}(t) x_{i\mu})\sigma(\frac{1}{\sqrt{S}}w_{cj}(t)x_{j\nu})\rangle_{S_{\rm w}}
\end{equation}
and
\begin{equation}
\bm{\Phi}'(t,t';x_\mu,x_\nu)=\langle\sum\limits_{c,i,j}\frac{1}{CS}\sigma'(\frac{1}{\sqrt{S}}w_{ci}(t) x_{i\mu})\sigma'(\frac{1}{\sqrt{S}}w_{cj}(t)x_{j\nu})\rangle_{S_{\rm w}}    
\end{equation}

The above action is quadratic in $\tilde{f},f$ , allowing us to take the average of $f_\mu(t)$, resulting in the following integral equation for the average of the network's output (mean predictor)
\begin{align} \label{delta=-y-int}
  &f_{\mu}(t)=-\int\limits_0^t \sum\limits_{\nu=1}^{P} \eta\left(  e^{-\frac{\eta\kappa^2}{\sigma_{\rm a}^2}(t-t')}\bm{\Phi}(t,t';x_\mu,x_\nu)\right. \\ \nonumber
  &\left.+ \frac{\sigma_{\rm a}^2}{2\chi}e^{-\frac{\eta\kappa^2}{\sigma_{\rm a}^2}|t-t'|}e^{-\frac{\eta\kappa^2}{\sigma_{\rm w}^2\chi}(t-t')} (\bm{\Phi}'(t,t';x_\mu,x_\nu)  K^x({x_\mu,x_\nu}))\right)
  (f_{\nu}(t')-y_\nu) \dif t' 
\end{align}
Where $K^x(x_\mu, x_\nu)=x_\mu\cdot x_\nu$.
In our upcoming discussion, we will demonstrate how Equation \ref{delta=-y-int} serves as an interpolation between the Neural Tangent Kernel (NTK) behavior during initial time periods and the Neural Network Gaussian Process (NNGP) behavior as it approaches equilibrium. At this juncture, it is pertinent to mention that the work by \cite{avidan2023connectingntknngpunified} in which they derive an integral equation for the average predictor by employing a Markov proximal approach.
They show that in the vanishing learning rate limit, their approach converges to gradient descent + noise. Their result for the mean predictor converges to the expression above at the vanishing learning rate limit.
\clearpage
\section{Recovering NTK dynamics}
To recover the NTK description for the neural network's output, we take the $t$ derivative of Equation \ref{delta=-y-int}.
 \begin{align}
       &\dot{f}_\mu(t)=-\sum\limits_{\nu}\eta\left( \bm{\Phi}(t,t;x_\mu,x_\nu)+\frac{\sigma_{\rm a}^2}{2\chi}(\bm{\Phi}'(t,t;x_\mu,x_\nu) K^x(x_\mu,x_\nu))\right) (f_\nu(t)-y_\nu)\notag\\ 
       & -\sum\limits_{\nu}\!\int\limits_0^t\left(
       {\eta^2}\frac{\eta\kappa^2}{\sigma_{\rm a}^2} e^{-\frac{\kappa^2}{\sigma_{\rm a}^2}(t-t')}\bm{\Phi}(t,t';x_\mu,x_\nu)+ \right.\notag\\
       &\left. (\frac{\kappa^2}{\sigma_{\rm a}^2}+\frac{\kappa^2}{\sigma_{\rm w}^2\chi}){\eta^2} \frac{\sigma_{\rm a}^2}{2\chi} e^{-\frac{\eta\kappa^2}{\sigma_{\rm a}^2}|t-t'|} e^{-\frac{\eta\kappa^2}{\sigma_{\rm w}^2\chi}(t-t')}(\bm{\Phi}'(t,t';x_\mu,x_\nu)  K^x(x_\mu,x_\nu))
       \right) \notag\\
       &\phantom{\left. (\frac{\kappa^2}{\sigma_{\rm a}^2}+\frac{\kappa^2}{\sigma_{\rm w}^2\chi}){\eta^2} \frac{\sigma_{\rm a}^2}{2\chi} e^{-\frac{\eta\kappa^2}{\sigma_{\rm a}^2}|t-t'|} e^{-\frac{\kappa^2}{\sigma_{\rm w}^2\chi}(t-t')}dsgoirhs\right)}(f_\nu(t')-y_\nu) \dif t'\notag\\
       &-\sum\limits_{\nu}\!\int\limits_0^t\left(
       {\eta^2} e^{-\frac{\eta\kappa^2}{\sigma_{\rm a}^2}(t-t')}(\bm{\Phi}'(t,t';x_\mu,x_\nu) K^x(x_\mu,x_\nu))\frac{\kappa^2}{2\chi} e^{-\frac{\eta\kappa^2}{\sigma_{\rm w}^2\chi}|t-t'|}
       \right.\notag\\
       &+{\eta^2}\frac{\sigma_{\rm a}^2}{2\chi}e^{-\frac{\eta\kappa^2}{\sigma_{\rm a}^2}|t-t'|}e^{-\frac{\eta\kappa^2}{\sigma_{\rm w}^2\chi}(t-t')}(\bm{\Phi}''(t,t';x_\mu,x_\nu) K^x(x_\mu,x_\nu) K^x(x_\mu,x_\nu))\notag\\
       &\phantom{+{\eta}\frac{\sigma_{\rm a}^2}{2\chi}e^{-\frac{\eta\kappa^2}{\sigma_{\rm a}^2}|t-t'|}e^{-\frac{\kappa^2}{\sigma_{\rm w}^2\chi}(t-t')}asga}\left.\frac{\kappa^2}{2\chi} e^{-\frac{\eta\kappa^2}{\sigma_{\rm w}^2\chi}|t-t'|}
       \right)(f_\nu(t')-y_\nu) \dif t' 
    \end{align}
it is now apparent that either by using $\int\limits_0^t g(t') \dif t'=t\ g(0)+\mathcal{O}(t)$ under $t\ll\frac{\eta\kappa^2}{\sigma_m^2}$ where we take $\sigma_m=\min(\sigma_{\rm a},\sigma_{\rm w})$ or under the limit $T\rightarrow0$ the expression above reduces to:
\begin{equation}
    \dot{f}_\mu(t)=-\sum\limits_{\nu}\eta\left(\bm{\Phi}(t,t;x_\mu,x_\nu)+\bm{\Phi}'(t,t;x_\mu,x_\nu) K^x(x_\mu,x_\nu) \right) (f_\nu(t)-y_\nu)
\end{equation}
however, since the time dependence of $\bm{\Phi}(t,t'),\bm{\Phi}'(t,t')$ comes through as a function of the time correlate $\bm{\Sigma}(t,t')$ which has the properties $\bm{\Sigma}(t,t)=\bm{\Sigma}(0,0)$ both $\bm{\Phi}(t,t)=\bm{\Phi}(0,0)$ and $\bm{\Phi}'(t,t)=\bm{\Phi}'(0,0)$
finally
\begin{gather}
    \bm{\Phi}(0,0;x_\mu,x_\nu)+ \bm{\Phi}'(0,0;x_\mu,x_\nu) K^x(x_\mu,x_\nu)=\notag\\\left\langle\bm{\nabla}_a f(0,x_\mu)\cdot\bm{\nabla}_af(0,x_\nu)+\bm{\nabla}_w f(0,x_\mu)\cdot\bm{\nabla}_w f(0,x_\nu)\right\rangle_{S_{\rm w}}=\Theta_0
\end{gather}
This is $\langle\sum_{\alpha} \partial_{\theta_{\alpha}}f_0(x_{\mu}) \partial_{\theta_{\alpha}}f_0(x_{\nu})\rangle_{S_{\rm w}}$ namely the Neural Tangent Kernel (NTK) in the infinite width limit averaged under random initializations, as it was defined in Equation \ref{Eq:NTK}. 
These, together with the initial condition $f_{\mu}(0)=-y_{\mu}$ gotten by setting $t=0$ in Equation \ref{delta=-y-int}, result in the same dynamical description given by the NN-NTK mapping.

\section{Equilibrium and the NNGP limit}\label{Result 3}
This section recovers the mean of the network output at equilibrium as predicted by the NN-NNGP mapping.
Consider Equation \ref{delta=-y-int}, and rewrite  $e^{-\frac{\eta\kappa^2}{\sigma_{\rm a}^2}(t-t')}$ in the first term under the integral as $\frac{\sigma_{\rm a}^2}{\eta\kappa^2}\partial_{t'}[ e^{-\frac{\eta\kappa^2}{\sigma_{\rm a}^2}(t-t')}]$ then integrating by parts. Finally setting $\sigma=\sigma_{\rm a}=\sigma_{\rm w}$ leaves us with:
\begin{align}
  f_{\mu}(t)=&-\left.\frac{\sigma^2}{\kappa^2} e^{-\frac{\eta\kappa^2}{\sigma^2}(t-t')}\sum\limits_{\nu}\! \bm{\Phi}(t,t';x_\mu,x_\nu) (f_{\nu}(t')-y_{\nu}) \right|_{t'=0}^{t}\notag\\&+\sum\limits_{\nu}\!\int\limits_0^t\!  \left(\frac{\sigma^2}{\kappa^2}  e^{-\frac{\eta\kappa^2}{\sigma^2}(t-t')} \bm{\Phi}(t,t';x_\mu,x_\nu)\dot{f}_{\nu}(t') \right) \dif t' \label{the discrepancy equation}
\end{align}
at the limit $t\rightarrow\infty$ the term
\begin{equation} \label{eq:integral term}
    I=\int\limits_0^t  \left( \frac{\sigma^2}{\kappa^2}  e^{-\frac{\eta\kappa^2}{\sigma^2}(t-t')} \sum\limits_{\nu}\!\bm{\Phi}(t,t';x_\mu,x_\nu)\dot{f}_{\nu}(t') \right) \dif t'
\end{equation}
vanishes.
Leaving us with
\begin{equation}
    f_{\GP}(x_\sigma)=\sum\limits_{\mu,\nu}K^{\GP}_{\sigma,\mu}
    [(K^{\GP}+\kappa^2 I)^{-1}]_{\mu,\nu}
    y_\nu
\end{equation}
where $K^{\GP}_{\mu,\nu}=\lim\limits_{t\rightarrow\infty}\sigma^2\bm{\Phi}(t,t;x_\mu,x_\nu)$ and $ f_{\GP}(x_\sigma)=\lim\limits_{t\rightarrow\infty} f_\sigma(t)$ this is Equation \ref{Eq:GPR} for the mean predictor.

%% file: Chapters/Summary_and_Outlook.tex
\chapter{Summary and Outlook}
Besides being the formalism underlying the standard model, field theory serves a major role in physics as a unifying ground to discuss theories of large scale interacting systems. This has enabled fruitful exchanges between condensed matter, statistical mechanics, and high energy communities \citep{Fradkin_2024}, and allows approximations, and other computational techniques, to be compared and shared via one common language. The goal of this review was to portray the possibility and potential benefits of extending such field-theory language to the deep learning domain. For instance, by showing how the approximation of Ref. \cite{Canatar2021}, derived for GPR, integrates rather seamlessly into feature learning formalisms or how Wilsonian RG can be applied to deep learning. 

Considering the above aim, and due to limitations of space and knowledge, various interesting and impactful approaches to deep learning have been, unfortunately, sidelined. Next, we partially remedy this state of affairs by reviewing and establishing links with some of those. 

\section{Other perspectives}

{\bf NTK and its hierarchy.} Following the conceptual simplicity and rigorous nature of the NTK result \cite{Jacot2018}, several authors have developed $1/N$ expansions \cite{roberts2021principles,huang2019dynamicsdeepneuralnetworks,dyer2019asymptotics} around the NTK result. These track the evolution of the NTK kernel in terms of higher-order tensors associated with neural network output and gradients. They can be considered analogues to the perturbative expansion introduced in Sec. \ref{Sec:PT}, having the capacity (and complexity) to track dynamical effects, conditioned on where one may truncate this expansion. In addition, \cite{roberts2021principles} provide a different application of RG in deep learning, by viewing the recursion relations defining network kernels as forms of a discrete RG transformation.In this context, we also mention several works that study Bayesian and random infinitely wide neural networks of large depth in particular depth which is of the order of the width and found an interesting regime of feature learning associated with the depth \cite{Hanin2023,hanin2024bayesian}.

{\bf Neural network field theory.} The interpretation we followed here, viewing neural networks as field theories, was also taken in the context of ``neural network field theories'' \citep{Halverson2021,Demirtas2023} (see also \cite{naveh2021predictingoutputsfinitedeepV1}) though with an emphasis on the ability of neural network priors (or random DNNs) to act as simulators or regulators of actual field theories one encounters in physics. Specifically, Ref. \cite{Demirtas2023}, provides techniques to engineer random weights or neural network priors to mimic physical actions such as a $\phi^4$ theory.

{\bf One step SGD.} Several authors studied feature learning effects in neural networks after one or several $O(1)$/giant steps of SGD \citep{ba2022highdimensional,damian2022neuralnetworkslearnrepresentations,dandi2023twolayerneuralnetworkslearn}, and found that these can change sample complexity classes, as we also discussed in the context of kernel adaptation. Others used such an approach to deduce the scaling of dataset-sampling-induced noise in representation learning with $P$ \citep{Paccolat_2021}, and use this scaling to rationalize the behavior of learning curves. As discussed in Sec. \ref{Sec:Hyper} the success of these approaches seems to rely on strong correlations between network behavior near initialization and close to equilibrium.

{\bf Narrow Neural Networks with Gaussian data.} Several authors study networks of the type $g(Wx)$ where $g$ is a non-linear function of an $N$ dimensional vector $Wx$ (also known as a multi-index model for $N>1$), $W$ is a $N \times d$ trainable weight matrix, and $x$ is Gaussian iid data and a similar type of teacher network with weight matrix $W_* \in R^{N_* \times d}$. This approach can capture aspects of deeper neural networks by taking $g$ to be the function generated by a neural network with fixed second layer weights (typically set to 1). Following \citet{saad1995dynamics,biehl1994line, biehl1995learning, saad1995exact} considering online SGD at a suitably defined vanishing learning rate, one finds a deterministic dynamics for the order parameters $Q = W W^T$ and $R=W W_*^T$ which can be calculated explicitly for activation function such as $g=Erf$. This provides an efficient low-dimensional description of the dynamics at large $d$ and $N,N_*=O(1)$ \citep{saad1995dynamics, goldt2019dynamics,wang2019solvable, arnaboldi2023highdimensional,ben2022high,damian2023smoothing,Bruno,dandi2024benefits, collinswoodfin2023hitting}. The same $Q,R$ order parameters can be used in Bayesian inference context leading, via saddle-point on the action of the Bayesian partition function, to a set of non-linear matrix equations on $Q,R$ (see recent review by \citet{cui2024high} and Refs. therein) which can be solved efficiently using Approximate Message Passing algorithms. 

{\bf Distributional Dynamics.} Considering a two-layer neural network of infinite width and finite input dimension, $d$, at mean-field scaling learning can be recast into a PDE governing the probability flow \citep{MeiMeanField2018,sirignano2020mean,rotskoff2022trainability} on the distribution of $a_i,w_i$ (read-out and input weights per neuron) which enjoys some elegant properties. This approach was also seminal in promoting the notion of mean-field scaling which is related to maximum adaptive scaling  \citep{yang2019scaling,yang2022tensorprogramsvtuning}.

\backmatter  

\printbibliography